\documentclass[review,10pt]{elsarticle}      % TLDH: 10pt body text      % TLDH: 10pt body text
\usepackage{float}
\usepackage{amsmath}
\usepackage{amssymb}
\usepackage{graphicx}
\usepackage[utf8]{inputenc}                  % UTF-8, replaces latin9
\usepackage{times}                            % TLDH: Times New Roman
\usepackage[verbose,tmargin=1in,bmargin=1in,lmargin=1in,rmargin=1.1in]{geometry}
\usepackage{setspace}
\usepackage{enumitem}
\usepackage{bibunits}                            % for cross-reference to supplementary
\usepackage{tabularray}
\usepackage{subcaption}
\usepackage{longtable}
\usepackage{array}
\usepackage{ragged2e}
\usepackage{booktabs}
\usepackage{xurl}
\usepackage{xr-hyper}
\usepackage[unicode=true, bookmarks=false, breaklinks=true]{hyperref}

\usepackage{enumitem}
\usepackage{array}
\newcolumntype{L}[1]{>{\raggedright\arraybackslash}p{#1}}
\usepackage{setspace}
\usepackage{booktabs}
\usepackage{array}
\usepackage{enumitem}
\usepackage{xr-hyper}   
\usepackage{ragged2e}% for cross-reference to supplementary

\usepackage{xcolor}
\usepackage{titlesec}
\titleformat*{\section}{\large\bfseries\sffamily}
\titleformat*{\subsection}{\normalsize\bfseries\sffamily}
\titleformat*{\subsubsection}{\small\bfseries\sffamily}

\renewenvironment{abstract}{\global\setbox\absbox=\vbox\bgroup
  \hsize=\textwidth%
  \noindent\unskip\textbf{\large Summary}
  \par\medskip\noindent\unskip\ignorespaces}{\egroup}

\makeatletter
\renewcommand\@biblabel[1]{#1}
\makeatother
\makeatletter
\def\ps@pprintTitle{%
  \let\@oddhead\@empty
  \let\@evenhead\@empty
  \let\@oddfoot\@empty
  \let\@evenfoot\@empty
}
\let\ps@pprintTitle\ps@empty
\makeatother
\journal{}

\begin{document}
\biboptions{super,sort&compress}
\begin{bibunit}[vancouver]

%The methods reviewed here provide the technical building blocks; the cultural, methodological, and regulatory shifts proposed here are needed to turn them into a clinical verification infrastructure.
%The methods reviewed here provide the technical building blocks for explanation, but their clinical value will depend on standardised evaluation, concept-level reference annotations, scalable WSI implementation, risk-stratified reporting expectations, and integration into pathology workflows.

\begin{frontmatter}

\title{Explainable Artificial Intelligence (XAI) in Computational Pathology: Definitions, Taxonomy, and Recommendations}

%% Authors (group per affiliation; replace with full author list)
\author[1,2]{Shubham Innani}
\author[1,2]{Suhang You}
\author[10]{Adam Shephard}
\author[4]{Bhakti Baheti}
\author[7]{Francesco Ciompi}
\author[11]{Joe Yeong}
\author[10]{Nasir Rajpoot}
\author[1,2]{Michael Feldman}
\author[7]{Solene Florence Kammerer-Jacquet}
\author[3]{Dimitrios Makris}
\author[7]{Geert Litjens}
\author[8,9,14]{Anne L. Martel}
\author[12]{Jana Lipkova}
\author[13,14]{April Khademi}
\author[1,2,3,5,6]{Spyridon Bakas\corref{corrauth}}
\author[]{for MICCAI SIG-CompPath}

\address[1]{Division of Computational Pathology, Department of Pathology and Laboratory Medicine, Indiana University School of Medicine, Indianapolis, IN, USA}
\address[2]{Indiana University Melvin and Bren Simon Comprehensive Cancer Center, Indianapolis, IN, USA}
\address[3]{Department of Computer Science, School of Computer Science and Mathematics, Kingston University London, London, UK}
\address[4]{Department of Biomedical Engineering, Emory University, Atlanta, GA, USA}
\address[5]{Departments of Biostatistics and Health Data Science; Radiology and Imaging Sciences; Neurological Surgery, Indiana University School of Medicine, Indianapolis, IN, USA}
\address[6]{Department of Computer Science, Luddy School of Informatics, Computing, and Engineering, Indiana University, Indianapolis, IN, USA}
\address[7]{Department of Pathology, Radboud University Medical Center, Nijmegen, The Netherlands}
\address[8]{Department of Medical Biophysics, University of Toronto, Toronto, ON, Canada}
\address[9]{Physical Sciences Platform, Sunnybrook Research Institute, Toronto, ON, Canada}
\address[10]{Tissue Image Analytics Centre, Department of Computer Science, University of Warwick, Coventry, UK}
\address[11]{Singapore General Hospital, Singapore}
\address[12]{University of California, Irvine, School of Medicine, CA, USA}
\address[13]{Toronto Metropolitan University, Toronto, ON, Canada}
\address[14]{Vector Institute, Toronto, ON, Canada}
\cortext[corrauth]{Corresponding author.}
\ead{spbakas@iu.edu}

%% =======================================================================
\begin{abstract}
Computational pathology (CompPath) is transforming medicine by leveraging artificial intelligence (AI) algorithms to support diagnosis, prognosis, and treatment prediction from gigapixel whole-slide images. Clinical adoption is progressing, but is constrained by concerns about safety, accountability, and regulatory oversight in high-stakes clinical environments. Explainable AI (XAI) systems hold promise for building trust and enabling verification, yet the literature remains fragmented due to inconsistent terminology, overlapping methodological families, ad hoc validation, and current reviews. This review aims to formalize XAI methods in CompPath through the: i) introduction of a pathology-centric vocabulary comprising seven core terms; ii) development of a taxonomy across methodological families and three orthogonal axes (stage, type, scope); and iii) establishment of a task-driven framework that maps five clinical questions to recommended methods, method evaluation, and deployment context. Five key gaps between current XAI capabilities and clinical deployment are identified, and actionable steps are proposed to advance XAI for CompPath.
\end{abstract}


\end{frontmatter}

% =======================================================================
%% INTRODUCTION
%% =======================================================================
\section{Introduction}
 
% [Your existing Introduction goes here, condensed to ~700 words]
% The condensed Introduction should:
% (1) frame the diagnostic-AI/CompPath landscape;
% (2) name the central challenge (opacity vs. clinical adoption);
% (3) preview the five contributions of this review;
% (4) outline the paper's structure and the narrative-review methodology.

% histopath and ai
    Histopathology has been the cornerstone of disease diagnosis, traditionally based on the visual inspection of tissue samples under a microscope \cite{gurcan2009histopathological}. Recent advances in digital scanners and high-throughput infrastructure have transformed this field into digital pathology, in which glass slides of tissue sections are digitized as whole-slide images (WSIs). This transition has enabled large-scale, data-driven computational analysis for diagnosis, prognosis, treatment response prediction, and biomarker discovery \cite{compayl,collins2025artificial,innani2025ai,innani2025artificial,baheti_eano,baheti_eano_clustering,baheti2025multimodal,baheti2024prognostic,isbi,sno_idh,sno24,innani2025interpretable,innani2025path,graham2023screening,innani2025path67,thakur2025img,thakur2024tmic}, defining the domain of Computational Pathology (CompPath). These advances can reshape diagnostic workflows, improving efficiency, accuracy, and patient care \cite{matthews2024public,liu2024regulatory,schmidt2024mapping,poalelungi2024revolutionizing,mcgenity2024artificial}. 

\begin{figure}[!b]
        \centering
        \includegraphics[width=\textwidth]{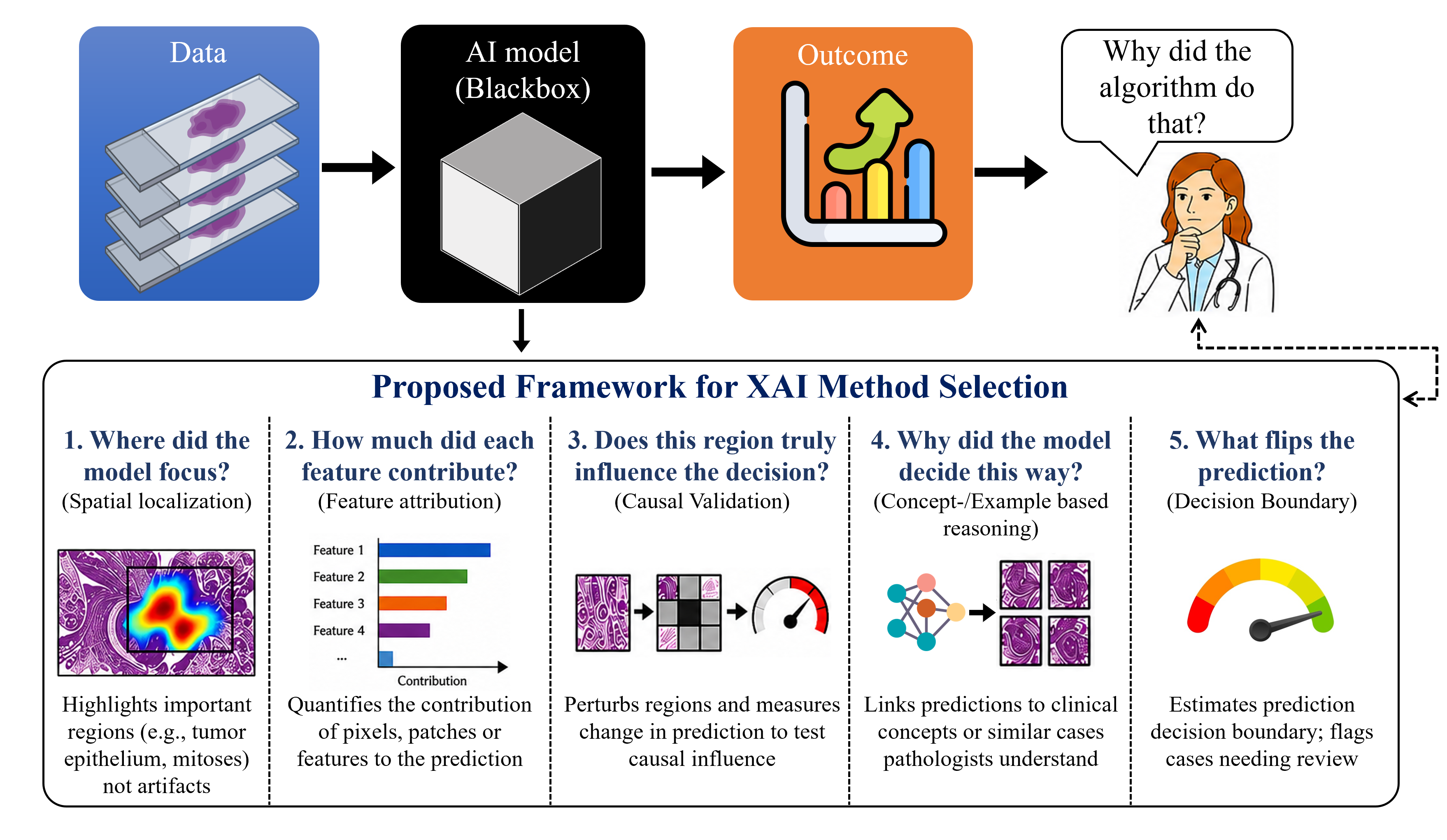}
        \caption{CompPath pipelines comprise data acquisition, model training, and outcome generation. Inputs span WSIs and tissue microarrays, as well as pathology reports, structured clinical and laboratory data, molecular profiles, and prior radiology findings. Outcomes include classification, segmentation and biomarker scores to text outputs generated by language and vision-language models, including structured-report drafts, concept-grounded retrieval results, and natural-language rationales. XAI can be integrated at any part of such pipeline to assess whether predictions rely on clinically relevant evidence.}
        \label{fig:block_diagram}
    \end{figure}
    
% regulatory approved tools with XAI limitations
    In recent years, clinical-grade AI tools for pathology images, such as Paige Prostate \cite{scans2024paige} and Ibex Galen \cite{pantanowitz2020artificial}, have obtained regulatory approval for cancer detection. These tools signal that clinical adoption is possible, but they also illustrate a central challenge: unlike long-established medical technologies, current AI models are new, with limited real-world deployment and longitudinal validation across diverse patient populations. Established medical devices build trust through years of validation, well-characterized limitations, and integration into medical training. Most current AI models in CompPath lack that evidence \cite{matthews2024public,liu2024regulatory,schmidt2024mapping,van2021deep}, and their failure modes, robustness, and generalizability beyond curated datasets remain under-studied \cite{Health_2024,jahanifar2025domain}. 

    A major reason is that CompPath pipelines, from data acquisition to model training and outcome generation, rely on complex deep learning (DL) models whose decision processes are not directly interpretable or explainable \cite{evans2022explainability}. The opacity of these systems means that model failures and lack of robustness cannot be directly understood, and pathologists cannot always trace why a particular decision was made. Whether interpretability is a strict prerequisite for clinical deployment is a matter of active debate. A recent Delphi consensus of 37 experts convened by ESMO classified ground truth, performance, and generalizability as essential requirements for AI-based biomarkers while treating explainability as recommended rather than mandatory, reflecting a view that reproducible tools validated in clinical trials can justify adoption without a mechanistic account of every prediction \cite{aldea2026ebai}. Ghassemi and colleagues have argued similarly that current XAI methods often provide false hope at the patient-decision level \cite{ghassemi2021false}. We take a middle position: explainability is not always strictly required, but is a strong contributor to trust, especially in early adoption, high-stakes clinical decisions, and regulatory review. To address this challenge, explainable AI (XAI) techniques have been proposed to provide insights into the decision-making process of these tools. Figure~\ref{fig:block_diagram} demonstrates the final recommendations and guidelines given in this review that highlights how XAI can illuminate the ``black-box'' by revealing the underlying model reasoning. 

    % What XAI contributes
    % XAI methods for CompPath serve multiple purposes beyond trust alone. They enable developers, regulators, and early users to interrogate model behaviour (is the model focusing on relevant features, or artifacts?), diagnose failure modes (why did the model fail?), and confirm that inference aligns with established pathology knowledge. Beyond validation, XAI offers a route to discovery: interpretability methods can decipher the predictive signatures learned by models trained on large cohorts, potentially revealing new morphology-biology relationships, and surfacing candidate biomarkers that pathologists can subsequently characterise. XAI is therefore relevant not only to clinical trust and regulatory approval, but also to expanding the biological and clinical value that AI models bring to pathology. The need for XAI is application-dependent: in high-risk use cases such as treatment response prediction, diagnostic decision support, or companion diagnostics, explainability may be necessary. The demand for XAI is context-specific, varying with clinical risk, regulatory scrutiny, and workflow integration. Some of the key questions are when, why, and for whom explainability matters in CompPath.
    XAI methods for CompPath can support model interrogation, failure analysis, and assessment of whether predictions align with established pathology knowledge. Their value is application-dependent: high-risk tasks such as treatment-response prediction, diagnostic decision support, or companion diagnostics may require stronger explainability evidence, whereas lower-risk or intrinsically inspectable outputs, such as segmentation maps, may require different forms of validation. The key question is therefore not whether XAI is universally required, but when, why, and for whom explainability matters in CompPath.

    % Describe the current gap in the literature in terms of xAI for computational path
    A persistent challenge in XAI literature is inconsistent terminology \cite{dwivedi2023explainable,doshi2017towards,holzinger2017we,hassija2024interpreting,arrieta2020explainable,lipton2018mythos} and lack of categorization and taxonomical definitions. Terms such as \textit{interpretability}, \textit{transparency}, and \textit{explainability} are often used interchangeably across technical and clinical audiences, creating confusion about what different XAI methods actually provide, and which method families are prevalent. This ambiguity interferes with efforts to define what constitutes sufficient explainability evidence for regulatory approval. It necessitates to establish what the underlying terms mean, and inconsistent usage across research, regulatory, and clinical communities creates delays. To address this, we build upon prior foundational work on XAI and its applications in biomedical imaging~\cite{adadi2018peeking,mueller2019explanation,murdoch2019definitions,ibrahim2023explainable,das2020opportunities,bhati2024survey,van2022explainable,evans2022explainability,pocevivciute2020survey} and adapt them specifically for CompPath. This review makes the following contributions:

    \begin{enumerate}
        \item \textbf{Terminology} Provide definitions for seven XAI concepts (interpretability, explainability, transparency, causal validity, causability, justifiability, and contestability) for CompPath 
        (Section~\ref{sec:terminology}, Table~\ref{tab:terminology}).
        
        \item \textbf{Three-axis taxonomy:} Position XAI methods along three orthogonal dimensions: stage (intrinsic vs\ post hoc), type (model-specific vs\ model-agnostic), and scope (local vs\ regional/slide vs\ global) (Section~\ref{section:XAI}, Figure~\ref{fig:xai_stage_type_scope}).
    
        \item \textbf{Method Families:} Organise the field into five method families (backpropagation-, perturbation-, feature-, concept-, and example-based) (Figure~\ref{fig:xai_taxonomy_families}), with strengths, limitations, and validation practices summarised in Supplementary Table~\ref{tab:xai_backprop_summary} --~\ref{tab:xai_example_summary}. Over 100 CompPath studies are surveyed (Section~\ref{sec:xai_dp}).
        
        \item \textbf{Recommendations:} Propose recommendations for task-driven application of XAI methods based on five clinical questions reflecting explainability versus a universal ``best'' method, with conditional guidance on model architecture, deployment context, and validation requirements (Section~\ref{sec:guide}).
    
    \end{enumerate}

%% =======================================================================
%% SECTION 2: XAI TERMINOLOGY (trimmed, with Table 1)
%% =======================================================================
\section{Definitions}
\label{sec:terminology}
    
    This article assesses seven interconnected terms, defined here and consolidated in Table~\ref{tab:terminology}. Detailed pathology-grounded definitions with examples are in Suppl. Section~\ref{supp:terminology}. Interpretability, explainability, and transparency are used inconsistently and these three terms form a closely related family with the common goal of making model behavior comprehensible to humans. The distinctions matter when: i) validation evidence must be documented, ii) methods are compared across studies, and iii) a clinical or regulatory reviewer asks what specific claims about a model actually means. A recent multi-disciplinary consensus convened by the Pathology Innovation Collaborative Community (PICC) illustrates the same pattern for the related term validation, which carries substantially different meanings across communication science, AI/ML, clinical and laboratory practice, regulatory science, and business contexts, with documented consequences for clinical translation \cite{dy2026clarifying}; the same fragmentation applies XAI.

    % Table 1: Terminology
    % [tab:terminology block goes here]
    \begin{table}[!h]
        \centering\small
        \caption{CompPath XAI terminology, treated as distinct, complementary concepts with different scopes and evaluation criteria.}
        \label{tab:terminology}
        \resizebox{\textwidth}{!}{%
        \begin{tabular}{@{}p{2.1cm} p{3.6cm} p{4.8cm} p{6.7cm}@{}}
        \toprule
        \textbf{Term} & \textbf{Core Question} & \textbf{Pathology Example} &
        \textbf{Evaluation Focus} \\
        \midrule
        
        \textbf{Interpretability} & Can a pathologist understand the model's reasoning directly from its structure or native outputs, without an added explanation? & Predicting tumor grade from nuclear area, mitotic count, \& gland density (structure-level); nuclear segmentation (output-level: inspectable output map). & Evaluated without generating external explanation. Two modes: (a) \emph{structure-level}, for simple parametric models (linear, generalised additive, sparse concept bottleneck) (b) \emph{output-level}, for complex models with a pathologist-recognisable quantity (segmentation map, mitotic count, Gleason grade) \\
        \midrule
        
        \textbf{Explainability} & Is faithful, context-appropriate evidence provided alongside the model's output? & Grad-CAM heatmap, concept-activation score for cribriform architecture, or calibrated probability accompanying a (for example) Gleason 4 prediction. & Explanation's faithfulness to the model, stability across scanners, stains, \& clinical plausibility; Categorizations of explanation-evaluation criteria identify 6 categories (faithfulness, robustness, localization, complexity, axiomatic, randomisation) \cite{hedstrom2022quantus}. Explanations are evaluated as artifacts and not assumed to reveal the model's true internal reasoning.\\
        \midrule
        
        \textbf{Transparency} & Is enough information disclosed to scrutinise, reproduce, monitor, and govern the system? & Public documentation of model architecture \& parameters; preprocessing details; subgroup performance across variables; known failure modes. & Completeness of the system-level disclosure; checklist-style against the relevant regulatory framework (FDA PCCP, EU AI Act). \\
        \midrule
        
        \textbf{Causal validity} & Does the highlighted feature influence the model's prediction? & Occluding a Grad-CAM-highlighted tumor region degrades the model's prediction, confirming model-causal influence. & Interventional or counterfactual evidence the highlighted feature influences the prediction (AOPC, insertion/deletion AUC, ROAR); distinguishes correlation from model-level causation.\\
        \midrule
        
        \textbf{Causability} & Does the explanation help a pathologist form a useful, clinically meaningful causal understanding? & Highlighting slide edges or artifacts (low causability) vs cribriform glandular architecture for a high-grade prostate prediction (high causability). & Expert agreement with the explanation; alignment with established clinical criteria (Gleason, Nottingham, WHO); evaluates the explanation from the user's clinical perspective.\\
        \midrule
        
        \textbf{Justifiability} & Can the deployment of the model and the use of its outputs be reasonably defended? & Defending a Gleason-grading tool with evidence of analytical \& clinical validation, calibration, subgroup fairness, \& standard of care alignment. & Completeness \& quality of the validation, fairness, \& regulatory dossier for the deployment decision. \\
        \midrule
        
        \textbf{Contestability} & Can pathologists and patients inspect, challenge, override, or appeal the model's outputs? & Pathologist overriding an AI-suggested Gleason grade, recording the disagreement in the report, \& triggering review. & Availability \& usability of inspection, override, \& audit mechanisms within the clinical workflow.\\
        
        \bottomrule
        \end{tabular}%
        }
    \end{table}

    Our proposed framework builds on prior XAI literature and adapts it for CompPath. Distinction between interpretability and explainability follows Lipton \cite{lipton2018mythos} and subsequent surveys \cite{arrieta2020explainable,doshi2017towards}. Our treatment of transparency as a system- and governance-level property follows Lipton \cite{lipton2018mythos} and emerging regulatory frameworks (EU AI Act, FDA guidance on AI/ML-based Software as a Medical Device) \cite{EU2024AIAct,FDAAIML}. The notion of \emph{causability}, the quality of an explanation as judged by a domain expert, is taken from Holzinger et al.\ \cite{holzinger2019causability}. \emph{Causal validity} as a term originates with Campbell, who proposed it to denote demonstrated causal relationships in a specific study context \cite{campbell1986relabeling,west2010campbell}; in XAI evaluation it has been adopted to describe if perturbing a model's input actually changes the prediction, distinguishing genuine input-output causation from mere correlation \cite{adebayo2018sanity,hooker2019benchmark}. The terms \emph{justifiability} and \emph{contestability} are drawn from clinical AI ethics and governance literature \cite{mittelstadt2019principles,graziani2020concept,ibrahim2023explainable,rueda2024just}, where they capture the ability to defend, audit, and challenge model outputs. We define the seven terms used throughout this review as summarized in Table~\ref{tab:terminology}, which consolidates each term's core question, a pathology-specific example, and the criterion against which it is evaluated (detailed examples are in Supplementary Section~\ref{supp:terminology}).

% \begin{itemize} 
% \item \textbf{Interpretability}: degree to which a human can understand a model's input-output behaviour directly from its structure or native outputs, without an additional explanation being generated. 
% \item \textbf{Explainability}: capacity to provide faithful, context-appropriate evidence (a heatmap, attribution score, segmentation, concept score, or natural-language rationale) alongside a model's prediction. 
% \item \textbf{Transparency}:  accessibility of information needed to scrutinise, reproduce, monitor, and govern an AI system across its lifecycle, delivered through documentation rather than through any computation on a specific case. 
% \item \textbf{Causal validity}: whether the features or regions highlighted by an explanation actually influence the model's prediction, established through interventional or counterfactual testing rather than correlation alone. 
% \item \textbf{Causability}: extent to which an explanation enables a domain expert to form a clinically meaningful causal understanding of the model's output, judged against established histopathological knowledge. 
% \item \textbf{Justifiability}: whether the deployment of an AI system and the use of its outputs can be reasonably defended within established clinical and regulatory standards. 
% \item \textbf{Contestability}: ability of pathologists, patients, and other stakeholders to inspect, challenge, override, or appeal the outputs of a deployed AI system. 
% \end{itemize}

    Three features distinguish our framework from generic XAI taxonomies. First, we keep \emph{causal validity} and \emph{causability} as technically separate concepts: causal validity is internal to the model (does perturbing a candidate region change the prediction?), while causability is external (does the resulting explanation align with established histopathological knowledge?). A model can satisfy one without the other, and clinical trust requires both. Second, every term is defined for CompPath such as nuclear pleomorphism, gland density, mitotic count, and tumor-infiltrating lymphocyte density, and we describe each term in relation to WSIs, TMAs, region-of-interest (ROI) analyses, and multiple instance learning (MIL) pipelines. Third, we treat the seven terms as a coherent system for clinical translation, rather than as competing synonyms: interpretability and explainability describe what is readable from the model and the evidence accompanying its output; transparency describes what the deployment system documents; causal validity and causability describe what an explanation must satisfy technically and clinically; justifiability and contestability describe the governance scaffolding that allows pathologists and regulators to act on the explanation.

%% =======================================================================
%% SECTION 3: TAXONOMY (single-sentence bullets, with Figure 2)
%% =======================================================================
\section{Taxonomy}
\begin{figure}[!b]
        \centering
    
        \begin{subfigure}[t]{0.95\textwidth}
            \centering
            \includegraphics[width=\textwidth]{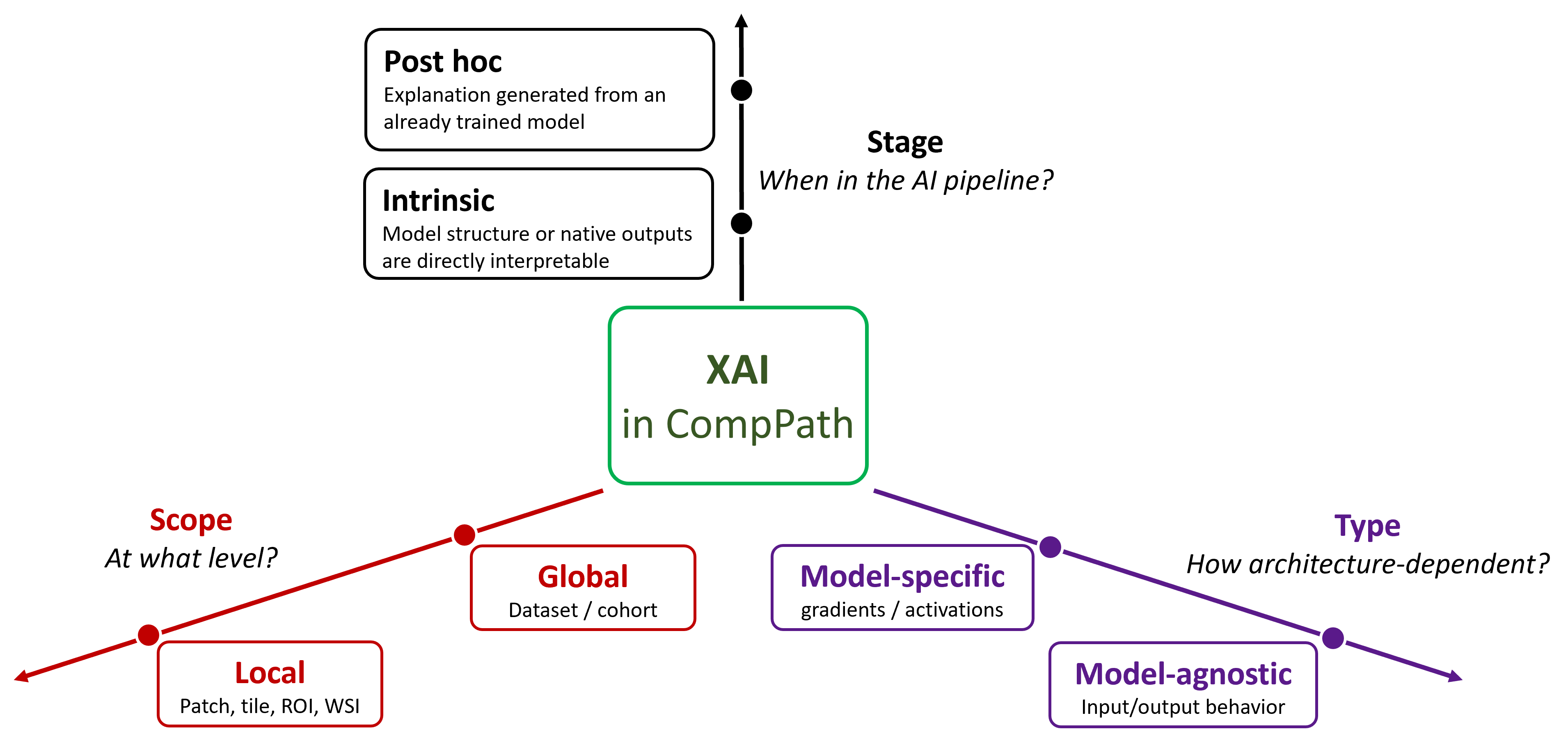}
            \caption{Stage-Type-Scope Taxonomy of XAI Methods in CompPath.}
            \label{fig:xai_stage_type_scope}
        \end{subfigure}
        \begin{subfigure}[t]{0.7\textwidth}
            \centering
            \includegraphics[width=\textwidth]{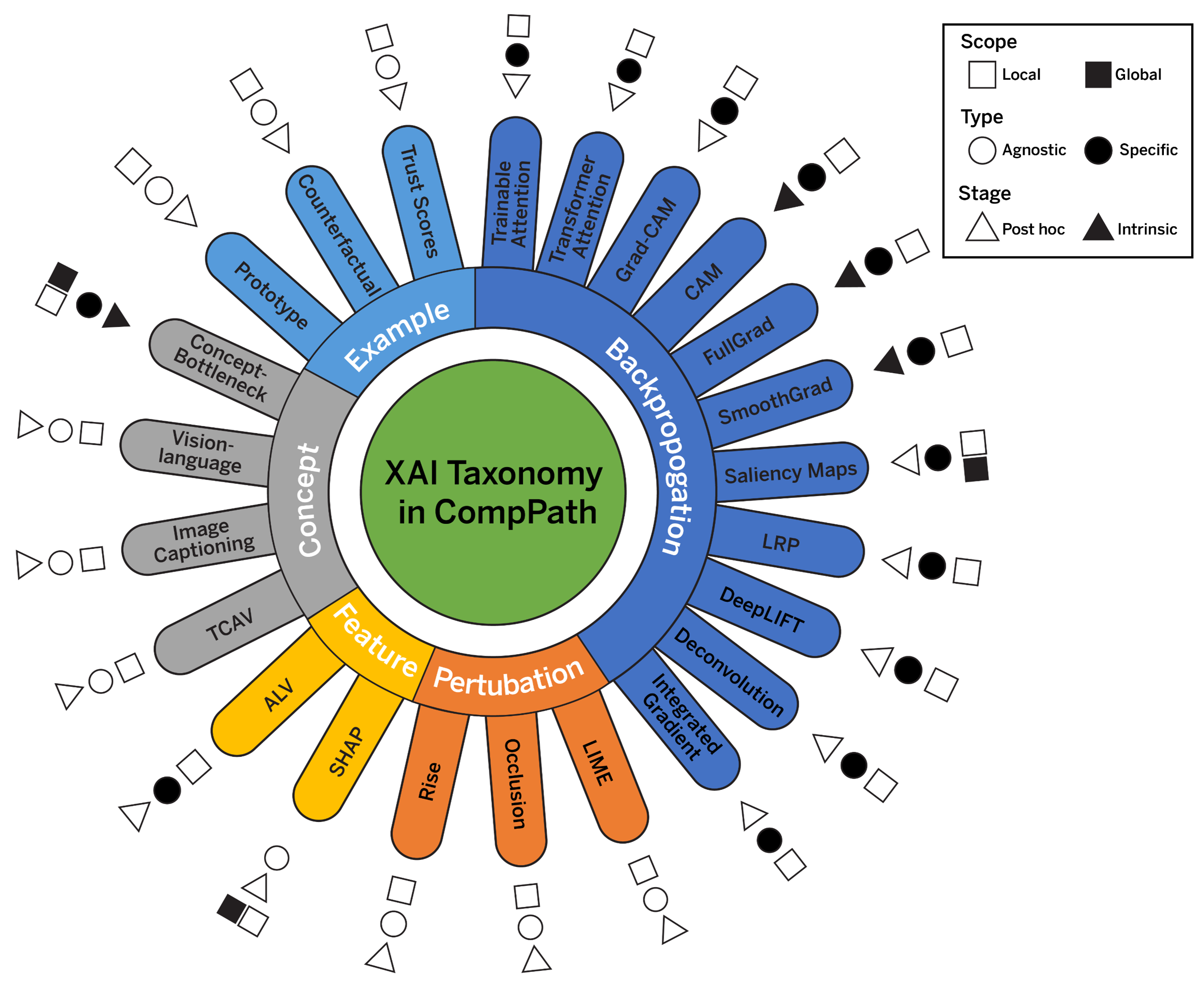}
            \caption{XAI Methodological Families in CompPath.}
            \label{fig:xai_taxonomy_families}
        \end{subfigure}
    
        \caption{Taxonomies of XAI methods in CompPath. \textbf{(a)} the Stage-Type-Scope taxonomy describes 3 orthogonal dimensions for each XAI method. Stage (intrinsic vs.\ post hoc) addresses when the explanation is produced. Type (model-specific vs.\ model-agnostic) addresses how dependent the explanation is on the model's architecture. Scope (local vs.\ regional/slide vs.\ global) addresses the level at which the explanation operates. \textbf{(b)} XAI Methodological Families and their positioning within the Stage-Type-Scope taxonomy.}
        \label{fig:families}
    \end{figure}
    
\subsection{Three Orthogonal Dimensions}
% \begin{figure}[!t]
%           \centering
%           \includegraphics[width=0.9\textwidth]{figs/XAI_stage_type_scope.png}
%           \caption{Stage-Type-Scope taxonomy for XAI in CompPath. Three orthogonal dimensions characterize each XAI method: Stage (intrinsic vs. post hoc) addresses when the explanation is produced; Type (model-specific vs. model-agnostic) addresses how dependent the explanation is on the model's architecture; and Scope (local vs. regional/slide vs. global) addresses at what level the explanation operates.}
%           \label{fig:xai_stage_type_scope}
%     \end{figure}
% \label{section:XAI}

    We position XAI methods along three orthogonal dimensions (Stage, Type, and Scope - Figure \ref{fig:xai_stage_type_scope}), providing a structured basis for comparing approaches and clarifying common terminology in the literature \cite{kamath2021explainable,adadi2018peeking,murdoch2019definitions}.
    \begin{itemize}
        \item \textbf{Stage} distinguishes \emph{intrinsic} methods, in which the explanation is part of the model's own computation (e.g., concept-bottleneck models, attention weights), from \emph{post hoc} methods, in which a separate procedure is applied to an already-trained model (e.g., Grad-CAM, SHAP).
        \item \textbf{Type} distinguishes \emph{model-specific} methods, which exploit gradients, activations, or architectural internals (and therefore depend on the model class), from \emph{model-agnostic} methods, which use only input--output behavior through perturbation, sampling, or surrogate fitting.
        \item \textbf{Scope} distinguishes \emph{local} explanations of an individual case (a single WSI, patch, or ROI) from \emph{global} explanations that characterize the model's behavior across a cohort or population.  
    \end{itemize}

    These axes are independent: a method's position on one does not fix its position on the others, although in CompPath most methods are clustered as post hoc, model-specific, and local. Detailed definitions and examples for each axis are provided in Supplementary Section~\ref{supp:taxonomy}, and Supplementary Table~\ref{supp:interpretability_techniques} categorizes each method.

%% =======================================================================
%% SECTION 4: METHOD FAMILIES (5 family overviews + Figure 3 + combined Table 2)
%% =======================================================================
\subsection{Methodological Families}
\label{sec:xai_dp}

% Opening prose
    XAI methods in CompPath can be categorized into five methodological families: (i) backpropagation-based, (ii) perturbation-based, (iii) feature-based, (iv) concept-based, and (v) example-based. Each family is characterized along five attributes relevant to clinical translation: a) operating principle, b) interpretability granularity (patch-, region-, or slide-level), c) computational cost, d) validation requirements, and e) readiness for clinical or regulatory integration. Figure~\ref{fig:xai_taxonomy_families} situates these families within the Stage--Type--Scope taxonomy, and Table~\ref{tab:xai_backprop_summary} -- ~\ref{tab:xai_example_summary} summarizes the strengths, limitations, and recommended validation practices for each method across the five families. A survey of over 100 CompPath studies across anatomies are provided in Supplementary Sections~\ref{supp:methods} and~\ref{supp:literature}.

% \begin{figure}[!t]
%           \centering
%           \includegraphics[width=0.9\textwidth]{figs/XAI_stage_type_scope.png}
%           \caption{Stage-Type-Scope taxonomy for XAI in CompPath. Three orthogonal dimensions characterize each XAI method: Stage (intrinsic vs. post hoc) addresses when the explanation is produced; Type (model-specific vs. model-agnostic) addresses how dependent the explanation is on the model's architecture; and Scope (local vs. regional/slide vs. global) addresses at what level the explanation operates.}
%           \label{fig:xai_stage_type_scope}
%     \end{figure}
% \label{section:XAI}

% % Figure 3: Taxonomy of families
% \begin{figure}[!h]
%   \centering
%   \includegraphics[width=0.7\textwidth]{figs/XAI_taxonomy_figure.jpg}
%   \caption{Overview of explainability families in CompPath, illustrating where backpropagation-, perturbation-, feature-, concept-, and example-based methods fit within the Stage-Type-Scope taxonomy.}
%   \label{fig:diagram}
% \end{figure}

    \label{section:XAI}

% Five family-overview paragraphs (one each for backprop, perturbation,
% feature, concept, example)
    \subsubsection{Backpropagation-based methods}
        Backpropagation-based methods are typically post hoc, model-specific, and local in scope. They compute gradients of a model's output with respect to its inputs (or to intermediate activations) to identify the pixels, patches, or features that most influence a prediction. In CompPath, this family produces the familiar heatmap-style explanations of WSI classifiers and MIL models, ranging from coarse class-level maps (CAM, Grad-CAM) to fine-grained, axiomatic attributions (integrated gradients, LRP, DeepLIFT). Attention-based methods (attention weights in MIL, self-attention in transformers) are often discussed alongside backpropagation-based methods because both operate on internal model components, but they differ in an important way: attention is computed during the forward pass as part of the model's native output, whereas gradient-based methods are applied post hoc. We treat attention within this family for organisational convenience but discuss its distinctive properties separately (Supplementary Section ~\ref{supp:methods_backprop}). Methods in this family share a common strength, namely, tight coupling to the trained model, and a common weakness, namely, that the gradient signal is correlational rather than causal and is sensitive to scanner, stain, and institutional variation \cite{jain2019attention,wiegreffe2019attention,jahanifar2025domain}. Examples are provided in Supplementary Section~\ref{supp:methods_backprop} and Supplementary Table~\ref{tab:xai_backprop_summary}.

    \subsubsection{Perturbation-based methods}
        Perturbation-based methods are typically post hoc and model-agnostic. They explain predictions by systematically modifying the input, occluding regions (occlusion sensitivity), sampling random masks (RISE), or fitting local surrogate models (LIME), and observing the resulting change in the model's output. Because perturbations are applied at inference time and do not require access to gradients or weights, this family generalizes across architectures, including black-box vendor models. The main practical constraints in CompPath are computational, namely thousands of forward passes per WSI for RISE-style estimators, and the biological realism of the perturbation itself, since zero-masking or solid-color occlusion of a WSI patch produces non-tissue artifacts that distort the resulting attribution. Supplementary Section~\ref{supp:methods_perturbation} and Supplementary Table~\ref{tab:xai_perturbation_summary} describes CompPath-specific examples for perturbation-based approaches.

    \subsubsection{Feature-based methods}
        Feature-based methods attribute predictions to interpretable input features rather than to raw pixels. They span post hoc and intrinsic designs, model-specific and model-agnostic implementations, and local-to-global scope. Two representatives are emphasized in this review: SHAP, which uses Shapley values from cooperative game theory to produce additive attributions over cells, glands, or tissue regions \cite{lundberg2017unified}, and activation-layer visualization (ALV), which inspects intermediate feature maps to characterize what a CNN has learned at successive depths \cite{yosinski2015understanding}. Feature-based methods are particularly useful when models operate on biologically meaningful features (e.g., nuclei, glands, regions), but their faithfulness depends on how those features are defined and on how the background distribution is sampled. See Supplementary Section~\ref{supp:methods_feature}  and Supplementary Table~\ref{tab:xai_feature_summary} for feature-based methods for CompPath.

    \subsubsection{Concept-based methods}
        Concept-based methods explain model predictions in terms of human-interpretable concepts rather than low-level features. In CompPath, the relevant concepts are pathologist-recognizable structures and patterns: necrosis, cribriform glands, lymphocytic infiltrate, tumor border, and mitotic figures. This family includes post hoc methods that probe an existing model for sensitivity to predefined concepts (TCAV \cite{kim2018interpretability}), intrinsic architectures that predict concepts as an intermediate output before the final clinical label (concept-bottleneck models \cite{koh2020concept}), and language-grounded methods that align image regions with diagnostic terminology (image captioning, vision-language models such as PLIP and BioViL \cite{zuo2024plip,bannur2023learning}). Concept-based methods trade annotation cost for interpretability gain, namely, concept labels need to be curated, but the resulting explanations are directly auditable by clinicians. Concept-based approaches for CompPath are provided in Supplementary Section~\ref{supp:methods_concept} and Supplementary Table~\ref{tab:xai_concept_summary}.

    \subsubsection{Example-based methods}
        Example-based methods explain a prediction by surfacing real cases: representative examples that resemble the input (prototypes), hypothetical edits that would change the prediction (counterfactuals), or neighborhood-based reliability estimates (trust scores). They mirror routine clinical reasoning, in which a pathologist contextualizes an ambiguous case by comparison with familiar reference cases. Strengths include intuitive case-based reasoning and natural support for uncertainty triage; limitations include sensitivity to the quality of the underlying embedding space and the difficulty of generating realistic counterfactual histology without violating biological constraints. Supplementary Section~\ref{supp:methods_example}  and Supplementary Table~\ref{tab:xai_example_summary} describes examples for CompPath.

%% =======================================================================
%% SECTION 5: TASK-DRIVEN RECOMMENDATIONS (as is)
%% =======================================================================
\section{Recommendations for Task-Driven Application of XAI in CompPath}
    \begin{figure}[!b]
          \centering
          \includegraphics[width=\textwidth]{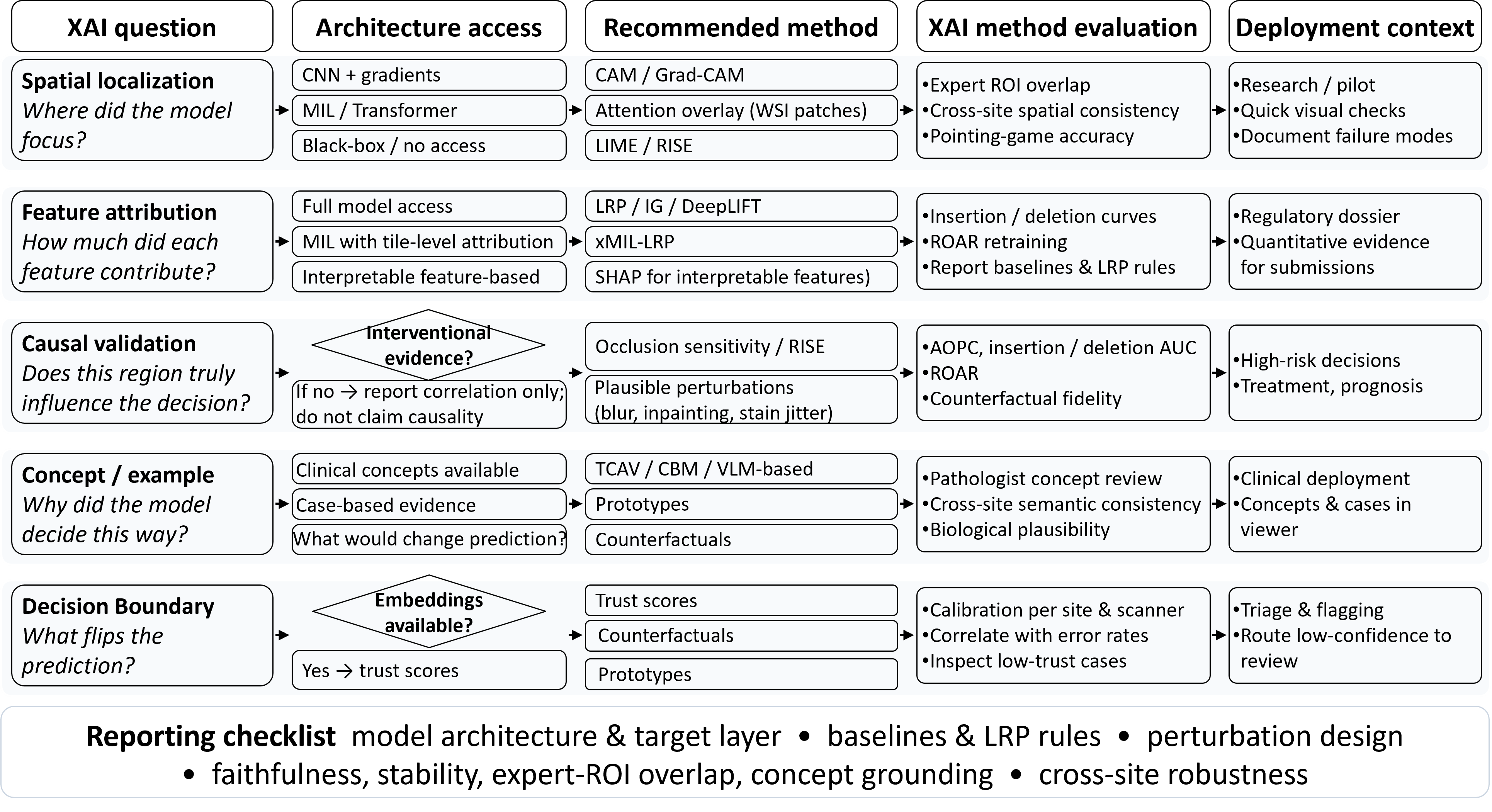}
          \caption{Task-driven recommendation framework for selecting XAI methods in CompPath according to ``Explanation questions". Five clinical questions (spatial localization, feature attribution, causal validation, concept-/example-based reasoning, and decision-boundary assessment) map to recommended methods, validation evidence, and deployment use. All the rows are complementary rather than sequential: spatial localization and feature attribution provide associational evidence, causal validation tests whether highlighted regions influence the model output, concept- and example-based methods translate model behaviour into pathologist-aligned reasoning, and decision-boundary methods identify unstable, ambiguous, or out-of-distribution cases that warrant review. Validation evidence and deployment requirements for each method family are summarised in Table~\ref{tab:xai_backprop_summary}--~\ref{tab:xai_example_summary} and discussed in Supplementary Section~\ref{supp:other}.} 
          \label{fig:rec}
    \end{figure}
\label{sec:guide}
    
    This section focuses providing practical guidance to researchers, clinicians, and regulators, based on the premise that XAI in CompPath is not one-size-fits-all. The choice of XAI method depends on three factors: the XAI question being asked, the model architecture being explained, and the deployment context (research exploration, regulatory submission, or clinical integration). Rather than prescribing universal ``best'' methods, we provide task-driven recommendations organized around five questions essential for XAI. Practical, regulatory, and deployment considerations are discussed in Supplementary Section~\ref{supp:other}. Figure \ref{fig:rec} provides an overview for selecting XAI methods in CompPath.

% Figure 4: Task-driven flowchart

% Five subsections, one per clinical question
% "Other Considerations" subsection moved to Supplementary
\subsection{Spatial localization: Where did the model focus?}

    Spatial localization verifies the model attends to histologically relevant regions (tumor epithelium, mitotic figures, immune infiltrates) rather than artifacts (pen marks, tissue folds, scanner signatures). Recommended methods include CAM, Grad-CAM for gradient-based localization, attention overlays for MIL and transformer models, and LIME or RISE for cases where the model's internal architecture, weights, or gradients are not accessible (for example, closed-source vendor models or third-party APIs). These approaches provide rapid, intuitive visualization  of model focus at the patch level, making them well-suited for preliminary confirmation that model activations correspond to biologically plausible regions.

    Attention-weighted aggregation in MIL or transformer models can generate slide-level maps, while preserving contextual relationships between patches. Patch-level heatmaps from Grad-CAM may be stitched into continuous WSI maps for high-resolution inspection. However, these methods primarily reveal correlation rather than causation and may overlook fine-grained histologic features. Spatial maps should be validated against expert-annotated ROIs and assessed for spatial consistency across scanners and institutions.

    % Slide-level foundation models (e.g., GigaPath) provide slide-level representations that are less amenable to attention-based visualization than patch-based MIL. Their aggregation happens over a single-vector slide embedding rather than a set of per-patch attention weights, so the "which patch drove the prediction?" question does not have the same direct answer as it does for ABMIL or transformer-based MIL. Attribution or perturbation at the tile level, or feature-space concept probing, remain applicable.

    %Before trusting a prediction, verify that the model attends to histologically relevant regions (tumor epithelium, mitotic figures, immune infiltrates) rather than artifacts (pen marks, tissue folds, scanner signatures). Recommended methods include CAM and Grad-CAM for gradient-based localization, attention overlays for MIL and transformer models, and LIME or RISE for black-box models or cases without architectural access. These approaches provide rapid, intuitive visualization of model focus at the patch level, well-suited for preliminary validation and sanity checks.

    % For WSI analysis, attention-weighted aggregation in MIL or transformer models can generate slide-level maps while preserving contextual relationships between patches, and patch-level Grad-CAM heatmaps may be stitched into continuous WSI maps. These methods reveal correlation rather than causation and may overlook fine-grained histologic features, so spatial maps should be validated against expert-annotated ROIs and assessed for spatial consistency across scanners and institutions.

    \subsection{Feature Attribution: How much did each feature contribute to a model prediction? }

        Quantifying feature importance enables systematic validation and provides numerical evidence for regulatory submissions, unlike qualitative visualization alone. Recommended methods include Layer-wise Relevance Propagation (LRP), Integrated Gradients (IG), and DeepLIFT for high-fidelity attribution with full model access, xMIL-LRP for MIL-based models with tile-level attribution needs. SHAP is best reserved for pipelines where the input features are already biologically interpretable (cell densities, gland counts, morphometric summaries). For end-to-end MIL models operating on raw patch embeddings, attention weights, LRP, or perturbation-based methods are typically more appropriate than SHAP, which lack interpretable feature semantics. These approaches provide mathematically grounded attribution scores that quantify the relative importance of pixels, patches, or features driving the prediction, enabling systematic assessment of whether the model relies on diagnostically meaningful regions.
    
        For WSIs, relevance propagation methods such as xMIL-LRP extend attribution through MIL aggregation layers, while feature-based summarization approaches (e.g., SHAP) can aggregate cell-level or region-level metrics into global slide-level importance distributions. Attribution maps depend on rule selection (for LRP) and baseline choice (for IG and DeepLIFT), and they are computationally intensive at the WSI scale. Implementation details, including propagation rules, baseline definitions, and integration steps should be reported, and faithfulness evaluation using insertion/deletion curves or RemOve And Retrain (ROAR) \cite{hooker2019benchmark}.

        %Quantifying feature importance enables systematic validation and provides numerical evidence for regulatory submissions, unlike qualitative visualization alone. Recommended methods include Layer-wise Relevance Propagation (LRP), Integrated Gradients (IG), and DeepLIFT for high-fidelity attribution with full model access, xMIL-LRP for MIL-based models, and SHAP for model-agnostic applications. Attribution maps depend on rule selection (LRP) and baseline choice (IG, DeepLIFT) and are computationally intensive at WSI scale. Always report implementation details (propagation rules, baseline definitions, integration steps) and evaluate faithfulness using insertion/deletion curves or RemOve And Retrain (ROAR) \cite{hooker2019benchmark}. 

    \subsection{Causal Validation: Does this region truly influence the decision?}
        Most localization and attribution methods, including attention maps, Grad-CAM, saliency maps, LRP, Integrated Gradients, DeepLIFT, and SHAP, identify regions or features associated with a prediction, but do not by themselves establish causal influence. Interventional or perturbation-based validation is therefore required before claiming that a highlighted region directly drives the model output. A region may receive high attribution scores because it correlates with the true causal feature, and not because it directly drives the prediction. Perturbation methods provide interventional evidence that a region influences the model output under a specified perturbation design. Recommended methods include Occlusion Sensitivity for conceptually straightforward causal validation and RISE for more stable results through randomized masking aggregation. When applying perturbation strategies, use biologically plausible perturbations such as Gaussian blurring (to simulate resolution loss), inpainting with neighboring tissue (to maintain texture continuity), or color jittering (to simulate stain variation), while avoiding zero-masking or solid colors, which can create unrealistic artifacts.
        
        These gradient-free methods perturb input regions and directly measure changes in model predictions, providing architecture-agnostic validation of causal influence. For WSI, perturbation-based aggregation can validate tile-level relevance at slide scale, with efficient batching and stratified sampling improving scalability. However, the approach is computationally intensive, and naive masking can produce unrealistic artifacts that mislead the model. Always quantify faithfulness using metrics such as insertion curves, deletion curves, or ROAR to verify that removing high-importance regions degrades performance more than removing low-importance regions.
        %Gradient-based methods such as Grad-CAM, LRP, and IG reveal correlation, not causation: a region may receive a high attribution score because it correlates with the true causal feature rather than directly driving the prediction. Perturbation methods provide interventional evidence under a specified perturbation design. Recommended methods include Occlusion Sensitivity for conceptually straightforward causal validation and RISE for more stable results through randomized masking aggregation. Use biologically plausible perturbations (Gaussian blurring, inpainting with neighboring tissue, color jittering) rather than zero-masking or solid colors, which introduce unrealistic artifacts, and quantify faithfulness using insertion curves, deletion curves, or ROAR.

    \subsection{Concept-/Example-based Reasoning: Why did the model decide this way?}

        Pixel-level or patch-level attribution maps answer ``where'' and ``how much'', but not ``why'' in clinical terms. Pathologists reason about concepts (glandular differentiation, immune infiltration, nuclear pleomorphism) rather than pixel gradients. Recommended methods include SHAP for feature-level attribution, Testing with Concept Activation Vectors (TCAV) for quantifying concept influence, Vision-Language Models (VLMs) for zero-shot concept retrieval, Prototypes for case-based reasoning, and Counterfactuals for generating hypothetical scenarios. These methods connect model reasoning to human-understandable concepts, clinical terminology, or representative examples. Concept-based methods link internal representations to quantitative morphometrics, while example-based approaches provide case-based evidence aligned with clinical reasoning.
        
        For WSI, concept overlays from TCAV or VLMs can project semantic sensitivity (e.g., necrosis, gland formation) spatially across slides, helping pathologists understand which clinical features the model is responding to. Prototype selection and counterfactual visualization can prioritize diagnostically informative regions for review. However, these methods require well-defined feature sets or concept sets. Poorly defined examples or semantic drift, for example, when a model's learned concept (e.g., ``stroma'') diverges from the clinical definition, can produce biased or misleading explanations. Always validate concept grounding by comparing model-identified concept regions with expert annotations and by testing for consistency across institutions.

        %Pixel- or patch-level attribution maps answer ``where'' and ``how much'' but not ``why'' in clinical terms. Pathologists reason about concepts (glandular differentiation, immune infiltration, nuclear pleomorphism) rather than pixel gradients. Recommended methods include SHAP for feature-level attribution, Testing with Concept Activation Vectors (TCAV) for quantifying concept influence, Vision-Language Models (VLMs) for zero-shot concept retrieval, Prototypes for case-based reasoning, and Counterfactuals for hypothetical scenario generation. These methods connect model reasoning to human-understandable concepts, clinical terminology, or representative examples, but require well-defined concept sets; semantic drift (when a model's learned concept diverges from the clinical definition) can produce misleading explanations. Always validate concept grounding against expert annotations and across institutions.

    \subsection{Decision Boundary: What flips the prediction?}
        For safe clinical deployment, cases that warrant pathologist review or have ambiguous morphology, out-of-distribution artifacts, or borderline diagnostic features must be identified. Example-based XAI methods address this by locating a test slide relative to the model's decision boundary and its training manifold. Three failure modes are diagnostically distinct. A case is \emph{unstable} when it sits close to the decision boundary, so small input changes flip the prediction; \emph{counterfactuals} are the relevant probe, since they identify the minimal change that crosses the boundary. A case is \emph{ambiguous} when it lies in a region of poorly learned manifold with inconsistent local neighbors; \emph{prototypes} are the relevant probe, since the labels of the nearest training cases reveal whether the model has learned a coherent local concept. A case is effectively \emph{out-of-distribution} when no training example is similar; \emph{trust scores} \cite{jiang2018trust} are the relevant probe, since they quantify distance to neighboring labeled examples in embedding space.
        
        For WSI applications, approximate nearest-neighbor indexing enables scalable deployment of prototype retrieval and trust scores, though similarity thresholds must be calibrated separately for each institution and scanner domain to account for distributional shift \cite{jahanifar2025domain}. Example-based evidence does not by itself explain why a prediction may be unreliable; it should be interpreted alongside attribution or concept-based evidence (e.g., ``which region is the model focusing on?''). Where the deployment context also requires calibrated probabilities of error or formal uncertainty estimates, these are the province of model-calibration and uncertainty-quantification methods, which are distinct from XAI and beyond the scope of this review; we refer readers to dedicated CompPath surveys \cite{abdar2021review,he2026survey}.

\section{Discussion}
\label{sec:discussion}
    In this article, we introduced a taxonomy of XAI Methods and offer practical, task-driven recommendations for researchers, clinicians, and regulators, based on a comprehensive literature review of over 100 XAI CompPath studies (Table~\ref{supp:tab:literature}). The presented taxonomy offers a two-fold categorization, by introducing a) a methodological organization across scale, type, and scope as orthogonal dimensions, and b) familial grouping based on core methodological principles. To facilitate these, we also provided clear definitions of the terms explainability, interpretability, justifiability, causality, and transparency, which have been traditionally used interchangeably in the literature.  

    The primary use of XAI in CompPath research has been visualization, enabled by its integration with CNN and MIL architectures, producing heatmaps, attention overlays, or prototype examples that make model behavior intuitively accessible. This is valuable for exploratory analysis and model debugging, but insufficient for clinical deployment or regulatory approval. visualization  answers ``where did the model look at?'', but not ``is the model reasoning correctly?''. 
    % Consider a Grad-CAM heatmap highlighting tumor epithelium in a prostate biopsy classified as Gleason grade 4. This visualization confirms the model is attending to a plausible region, but it does not prove the model has learned Gleason-relevant features (i.e., cribriform architecture, poor glandular differentiation). The model may be responding to correlated but diagnostically irrelevant patterns, such as nuclear density and spatial clustering, that happen to co-occur with grade 4 tumors in the training data. 
    A heatmap cannot distinguish between spurious correlations and causal reasoning. Verification requires three components that visualization alone cannot provide: 
    \begin{enumerate}
        \item \textbf{Causal attribution}: Does occluding the highlighted region actually change the prediction? Perturbation-based methods (Occlusion Sensitivity, RISE) provide this, but are computationally expensive for WSI.
        \item \textbf{Concept-level validation}: Does the model respond to clinically defined features (e.g., cribriform glands, necrosis, immune infiltration) or to uninterpretable activation patterns? Concept-based methods (TCAV, CBMs) provide this, but require expert concept annotations that are expensive to curate and prone to inter-observer variability.
        \item \textbf{Cross-site robustness}: Do explanations remain consistent across institutions, scanners, and staining protocols? Current XAI studies rarely test this. The validation for most is performed on held-out test sets from the same institution as training data.
    \end{enumerate}

    Only a few current CompPath XAI studies jointly evaluate model-level causal attribution, concept-level validity, and cross-site robustness. Achieving verification requires combining multiple methods such as spatial localization (Grad-CAM, attention), quantitative attribution (LRP, IG), perturbation testing (Occlusion, RISE), and concept validation (TCAV, CBMs) into integrated pipelines. Section \ref{sec:guide} provides recommendations for such pipelines. The shift from visualization to verification is not just a technical challenge, but a cultural one. Researchers trained to evaluate XAI methods by visual plausibility (``does the heatmap look reasonable?'') must adopt quantitative validation (faithfulness metrics, cross-site testing, concept grounding). Regulators accustomed to evaluating device performance through sensitivity/specificity must develop frameworks for evaluating explanation quality. For Pathologists to trust AI outputs they must learn to interrogate not just the predictions, but also the reasoning behind them. None of these have occurred at scale for clinical verification yet since:

   % Despite progress in the field of XAI, several challenges, if addressed, would accelerate its translation from a research tool to becoming a clinical verification framework:

    \begin{enumerate}
        \item XAI methods are currently evaluated qualitatively through visual inspection of heatmaps, with little quantitative validation. There is no CompPath equivalent of ImageNet for XAI, no shared dataset, ground-truth explanations, or standardized evaluation protocol. What is needed is a curated benchmark dataset, governed as a public utility \cite{haendel2026governing}, spanning multiple anatomies, institutions, scanners, and staining protocols, with expert-annotated ROIs, multi-pathologist annotations capturing inter-observer variability, reference-standard concept labels, and standardized metrics for faithfulness, stability, and cross-site robustness.
        \item Most XAI methods produce pixel- or patch-level attributions, but pathologists reason about concepts such as gland architecture, mitotic activity, and immune infiltration. Validating whether a model has learned clinically relevant concepts requires reference-standard concept annotations, which are expensive to obtain and prone to inter-observer variability\cite{bakas2024brats}, and concept-based XAI methods (TCAV, CBMs) remain underutilized as a result. Addressing this gap requires hierarchical concept taxonomies for common diagnostic tasks, multi-pathologist annotations with agreement metrics, automated concept discovery, and protocols to detect concept drift across institutions and populations.
        \item High-fidelity XAI methods (LRP, IG, RISE) are computationally expensive, often requiring hours per slide at a gigapixel scale. This forces most research onto fast but low-fidelity alternatives (Grad-CAM, attention overlays) that provide directional insights without quantitative rigor. Recent advances in foundation models for pathology show that gigapixel-scale computation is tractable when architectures and algorithms are co-optimized; applying similar engineering to XAI, together with feature-space rather than pixel-space attribution, is achievable.
        \item Regulatory frameworks (FDA, CE Mark) do not currently specify what constitutes sufficient explainability evidence, so developers do not know what to provide, and reviewers do not know what to require. Closing this gap requires explicit guidance stratified by risk level: qualitative visualization may suffice for low-risk applications (e.g., slide pre-screening), moderate-risk applications should require quantitative attribution with faithfulness metrics, and high-risk applications should mandate multi-method validation with cross-site robustness testing. Standardized reporting templates analogous to TRIPOD\cite{moons2015transparent,collins2015transparent} and CLEAR\cite{kocak2023checklist}, and trained regulatory reviewers, are needed to establish consistent standards.
        \item XAI outputs are typically presented as standalone visualizations disconnected from clinical workflows. Pathologists must context-switch between the diagnostic viewer and the XAI interface, creating friction that limits adoption, and most methods produce per-slide explanations, although diagnostic decisions integrate multiple slides, clinical history, and ancillary tests. Practical deployment requires XAI integrated directly into digital pathology viewers, multi-slide aggregation, unified dashboards combining attribution, concepts, uncertainty, and prototypes, and user studies of how pathologists actually use XAI outputs in diagnostic workflows.
    
    \end{enumerate}

    Two recent shifts deserve specific mention. First, the rise of pathology foundation models (FMs) raises new XAI challenges, rather than solving them. Tile-level FMs provide reusable embeddings but do not intrinsically add interpretability, whereas slide-level FMs (GigaPath, TITAN) aggregate over the WSI in ways that make even the modest interpretability of attention-based MIL less accessible. Vision-language models offer promising directions for concept-grounded retrieval and natural-language rationales, but they primarily describe what is in an image rather than which image features drive a specific downstream prediction (e.g., survival, treatment response).  Concept-bottleneck approaches point toward closing this gap but strong CompPath examples remain few. XAI in the FM era therefore requires new methods that operate on high-dimensional learned embeddings, rather than assuming that model scale itself confers interpretability. Second, different XAI methods routinely highlight different regions of the same slide. Disagreement is informative rather than a failure: convergence across methods (e.g., Grad-CAM, LRP, and perturbation testing all pointing to the same region) strengthens the evidence, while divergence flags either method-specific artifacts or genuinely ambiguous reasoning that warrants pathologist review. We recommend reporting at least one gradient-based and one perturbation-based method together, with disagreement explicitly characterized.

    % Several XAI directions active outside CompPath have yet to see meaningful uptake, and warrant attention as the field matures. Mechanistic interpretability techniques originating in language-model research \cite{templeton2024scaling,cunningham2023sparse}, notably sparse autoencoders that decompose activations into monosemantic features, could in principle recover pathologist-recognisable structures from FM activations without upfront concept annotations. Causal graph neural networks and causal representation learning \cite{scholkopf2021toward,zecevic2021relating} offer a principled route to distinguishing causal from confounded relationships in cell-graph and tissue-graph pipelines. Migrating these methods into gigapixel-scale CompPath, and validating them against pathologist-defined concepts, is a productive frontier for the field.
    
    The transition from visualization  to verification is the defining challenge for XAI in CompPath. Heatmaps confirm where a model considered, whereas verification confirms whether it reasoned correctly. The methods reviewed here provide the technical building blocks, but establishing clinical utility requires cultural shifts among researchers, regulators, and pathologists, cross-institutional benchmarks, and clear regulatory standards, outlined above. If these take place collectively and systematically, XAI can improve the adoption of AI in pathology, along with earning the trust required to scale it.

\subsubsection*{Contributors}
SI and SB conceived the review. SI led the literature analysis. SY provided substantial writing support across multiple sections of the original manuscript and contributed detailed constructive feedback during revision. SB supervised the work and oversaw the development of the final taxonomy. All other authors reviewed, provided constructive feedback, edited, and approved the content and structure final version of the manuscript.

\subsubsection*{Declaration of interests}
We declare no competing interests. % (If any author has a relevant interest, declare it here. TLDH % requires explicit disclosure for each author.)

\subsubsection*{Data sharing}
This review does not generate new primary data. All studies discussed are publicly available through their original publications.

%\subsubsection*{Acknowledgments}
%Research reported in this publication was partly supported by the National Cancer Institute (NCI) of the National Institutes of Health (NIH) under award number ITCR:U24CA279629. The content of this publication is solely the responsibility of the authors and does not represent the official views of the NIH.

%% =======================================================================
%% REFERENCES (Vancouver, superscript numbered, 75 max)
%% =======================================================================
\putbib[references]
\end{bibunit}

\newpage
\begin{bibunit}[vancouver]

\section*{Supplementary Material}
\setcounter{section}{0}
\setcounter{subsection}{0}
\setcounter{subsubsection}{0}

\renewcommand{\thesection}{S\arabic{section}}
\renewcommand{\thesubsection}{\thesection.\arabic{subsection}}
\renewcommand{\thesubsubsection}{\thesubsection.\arabic{subsubsection}}
% Unique hyperlink destinations for supplementary sections
\renewcommand{\theHsection}{supp.\arabic{section}}
\renewcommand{\theHsubsection}{supp.\arabic{section}.\arabic{subsection}}
\renewcommand{\theHsubsubsection}{supp.\arabic{section}.\arabic{subsection}.\arabic{subsubsection}}

This supplementary material provides expanded detail on the four components of the main paper that require depth beyond: (i) the seven pathology-centric XAI term definitions (Section~\ref{supp:terminology}), (ii) the three-axis taxonomy sub-dimensions and stage/type/scope categorisation table (Section~\ref{supp:taxonomy}), (iii) per-method narrative descriptions across the five method families (Section~\ref{supp:methods}), and (iv) a survey of over 100 CompPath studies organised by anatomy and XAI method (Section~\ref{supp:literature}). Section~\ref{supp:other} provides extended deployment guidance not included in the main recommendations section.

\section{Pathology-Centric Definitions of Seven XAI Terms} \label{supp:terminology}

% [Paste here: the seven detailed \item[] blocks from your current
% Section 2 — Interpretability, Explainability, Transparency,
% Causal validity, Causability, Justifiability, Contestability]
\begin{itemize}
    \item[] \textbf{Interpretability} in deep learning tools for CompPath refers to the degree to which a human user can understand and reason about a model's input-output behavior directly from the model's structure, learned components, or native outputs, without requiring any additional explanation to be generated \cite{lipton2018mythos,arrieta2020explainable,li2022interpretable}. Interpretability is a non-binary property of the model that depends on the model class, feature representation, task complexity, scale, and user expertise. Sparse linear models, decision trees, generalized additive models, concept bottleneck models, prototype-based models, and models using pathologist-defined semantic features may be interpretable when their components are inspectable, and their behavior can be traced. For example, a logistic regression predicting tumor grade from nuclear area, mitotic count, and gland density is interpretable because each feature's contribution can be examined directly. Models that output pathologist-recognizable quantities such as a mitotic count, a Gleason score, a nuclear grade, a TIL density, or a segmented gland map are interpretable at the level of their outputs, because the prediction itself is a quantity a pathologist already reasons about \cite{shaban2019novel,graham2019hover,shephard2024fully}. In CompPath applications, however, WSIs are gigapixel inputs and models routinely have hundreds of millions to billions of parameters \cite{lu2024visual}, so most deep architectures used for tasks such as whole-slide classification, biomarker prediction, or survival analysis are not directly interpretable. Attention-based MIL and transformer attention maps are partially interpretable: attention weights indicate which patches of a WSI were associated with the prediction and can be inspected without an additional method, but they do not by themselves provide faithful causal attribution \cite{jain2019attention,wiegreffe2019attention}. Interpretability is evaluated by inspecting the model itself: whether a pathologist can trace, from the model's structure, learned components, or native outputs, how the input maps to the prediction. Unlike the criteria that follow, no additional explanation artifact is generated or evaluated; the model itself is the object of evaluation. \\

    \item[] \textbf{Explainability} refers to the capacity to provide faithful, context-appropriate evidence for a model output in a form understandable to the intended user \cite{arrieta2020explainable,guidotti2018survey}. The key distinction from interpretability is what is being evaluated: interpretability is a property of the model itself, whereas explainability is a property of an additional artifact (a heatmap, attribution score, segmentation, concept score, or natural-language rationale) that is delivered alongside the prediction. The artifact may be produced by the model during inference or by a separate method applied after training; we defer that distinction to the Stage axis of the taxonomy \ref{supp:taxonomy} and use ``explainability'' here for the artifact itself, regardless of how it is generated.
    For a WSI-level Gleason 4 prediction, an explanation artifact might be a Grad-CAM heatmap that localizes the regions most associated with the prediction, a calibrated probability that quantifies confidence, a segmentation map of glandular structures, or a set of concept-level scores indicating which pathologist-recognizable patterns (e.g., cribriform architecture, glandular fusion) the model relied on. Several such artifacts can accompany the same prediction, each answering a different clinical question.
    Because an explanation is an artifact added on top of the prediction rather than the prediction itself, it can be evaluated independently of the model. Two methods applied to the same model can highlight different regions of the same WSI, and explanation artifacts have been shown to be sensitive to scanner, staining, and institutional variation in pathology data \cite{arun2021assessing,jahanifar2025domain}. Evaluation criteria specific to explanation artifacts include \textit{fidelity}, namely whether the artifact actually reflects the model's behavior (e.g., whether removing highlighted regions degrades the prediction); \textit{stability}, namely whether the same model yields consistent artifacts across stains, scanners, and institutions; and \textit{clinical plausibility}, namely whether the artifact aligns with pathologist judgment about what the model should be attending to. These criteria apply to the explanation artifact as an object of evaluation; interpretability, causal validity, and causability are evaluated against different objects (the model, the prediction-feature link, and the user's clinical understanding respectively).

        \item[] \textbf{Transparency} in CompPath refers to the accessibility of information needed to scrutinize, reproduce, monitor, and govern an AI system across its lifecycle \cite{lipton2018mythos,FDAAIML,EU2024AIAct}. Unlike interpretability and explainability, which are technical properties of the model or its accompanying evidence, transparency is a property of the system and is delivered through documentation rather than through any computation on a specific case. It encompasses model architecture, training data composition (the number and diversity of WSIs, scanners, staining protocols, and patient demographics), preprocessing pipelines (tissue segmentation, stain normalization, patch extraction), reference standards used for ground truth labeling, intended use, subgroup performance, known failure modes, and procedures for post-deployment monitoring. A pathology AI system can be highly explainable yet not transparent: a vendor product may produce attention-based heatmaps over each WSI while withholding the architecture, training cohort composition, and validation cohort demographics, making it impossible to assess whether the model's training population matches the deploying institution's case mix. Conversely, a system can be transparent yet not interpretable: a publicly released foundation model with billions of parameters \cite{lu2024visual} may have its architecture, training data, and weights fully disclosed and yet remain opaque to direct human inspection. Transparency requirements operate at two levels. At the technical level, they govern what a developer or independent researcher can reconstruct about a model's design, data, and behaviour from publicly available documentation. At the regulatory and legal level, they govern what a manufacturer must disclose to authorities and deploying institutions; the U.S. Food and Drug Administration's Predetermined Change Control Plan framework \cite{FDAAIML} and the EU AI Act's high-risk device requirements \cite{EU2024AIAct} both mandate disclosure of training data, intended use, subgroup performance characteristics, and change-management procedures. The two levels overlap but are not identical: a system can meet regulatory disclosure requirements while remaining technically opaque to external researchers, and a system can be technically transparent (e.g., open weights and training code) without yet meeting the structured disclosure expectations of a specific regulatory framework. Evaluation of transparency is based on the sufficiency, accessibility, and verifiability of the disclosure relative to the system's intended use and risk level \cite{mittelstadt2019principles,gebru2021datasheets,mitchell2019model}, rather than on any single property of the model's output on a specific case.\\

        \item[] \textbf{Causal validity.} Causal validity in computational pathology refers to whether the features or regions highlighted by an explanation actually influence the model's prediction, as opposed to merely correlating with it \cite{adebayo2018sanity,hooker2019benchmark}. Causal validity is a property of the \emph{explanation}, supported by interventional evidence, and is distinct from both technical fidelity and clinical plausibility: a method can produce a visually plausible heatmap over a tumor region that has no causal effect on the prediction \cite{adebayo2018sanity,jain2019attention}. Establishing causal validity requires perturbation- or counterfactual-based testing on the actual WSI or TMA input: occluding a region highlighted by Grad-CAM and verifying that the prediction degrades, performing insertion or deletion curves over patches ranked by attribution score, or generating counterfactual histology images in which a candidate feature (e.g., mitotic figures, cribriform glands) is removed or synthetically modified and observing whether the prediction changes \cite{petsiuk2018rise,wachter2017counterfactual}. In CompPath, causal validity is particularly important because models trained on WSIs can latch onto non-causal cues such as staining batch, scanner color profile, slide background, or pen marks \cite{baheti2024prognostic}. An explanation that highlights cribriform architecture but is unchanged when that architecture is occluded indicates a causal failure: the model is using something else, and the explanation is misleading. Evaluation of causal validity uses interventional metrics such as Area Over the Perturbation Curve (AOPC), insertion/deletion AUC, RemOve And Retrain (ROAR), and counterfactual fidelity \cite{samek2016evaluating,hooker2019benchmark}. Causal validity is internal to the model: it answers whether the explanation reflects the model's own decision process, not whether the highlighted feature is biologically causal for disease.\\ 
        
        \item[] \textbf{Causability.} Causability in computational pathology refers to the extent to which an explanation enables a pathologist to form a useful, clinically meaningful causal understanding of the model's output \cite{holzinger2019causability}. Whereas causal validity is internal to the model (does the explanation reflect the model's actual decision process?), causability is a property of the \emph{explanation} as evaluated by a human expert: does the explanation map to recognized histopathological features and biological mechanisms in a way that the pathologist can verify against domain knowledge? An explanation with high causal validity can still have low causability if it highlights regions or features with no established pathological meaning. For example, a survival model applied to glioblastoma WSIs may faithfully attribute its prediction to a region of dense pseudopalisading necrosis (high causal validity), and the pathologist recognizes pseudopalisading necrosis as an established adverse prognostic feature in high-grade gliomas (high causability). In contrast, the same model attributing its prediction to slide edges, tissue-folding artifacts, scanner-specific color casts, or out-of-tissue background regions has zero causability regardless of the technical fidelity of the attribution \cite{schmitt2021hidden}. Causability is therefore the bridge between technical explanation and clinical trust: it depends on the explanation aligning with pathologist-recognized structures such as nuclear atypia, gland formation, mitotic figures, lymphocytic infiltrate, or specific architectural patterns (cribriform, papillary, solid). Evaluation of causability requires pathologist input: reader studies in which pathologists rate the agreement between explanations and known histopathological features, comparison of model-highlighted regions against expert-annotated regions of interest, and structured assessment of whether explanations align with established grading criteria such as the Gleason, Nottingham, or WHO systems \cite{holzinger2020measuring}.\\ 
        
        \item[] \textbf{Justifiability.} Justifiability in computational pathology refers to whether the deployment of an AI system and the use of its outputs can be reasonably defended within established clinical and regulatory standards \cite{mittelstadt2019principles,graziani2020concept,ibrahim2023explainable,rueda2024just}. Justifiability is a property of the \emph{governance} surrounding deployment, not of the model or its explanations in isolation. It depends on the strength of validation evidence (analytical validation on WSI cohorts, clinical validation against pathologist diagnoses, and documented performance across scanners, stains, and patient populations \cite{jahanifar2025domain,dy2026clarifying}); fairness across demographic groups; compliance with regulatory frameworks (FDA clearance for the intended use, CE marking under the EU IVDR/MDR, or laboratory accreditation for laboratory-developed tests under CLIA); and alignment with the standard of care for the diagnostic task at hand. A heatmap may support a clinical justification, but it cannot by itself justify clinical deployment or patient-level action. Evaluation of justifiability relies on the completeness and quality of the validation, documentation, and regulatory dossier rather than on the model's output in any given case.\\ 
        
        \item[] \textbf{Contestability.} Contestability in computational pathology refers to the ability of pathologists, patients, and other stakeholders to inspect, challenge, override, or appeal the outputs of an AI system \cite{mittelstadt2019principles,EU2024AIAct}. Contestability is a property of the \emph{workflow} into which the system is deployed, encompassing user-facing mechanisms (e.g., a one-click override of an AI-suggested Gleason grade, mitotic count, or molecular subtype prediction, with the disagreement recorded in the diagnostic report), patient- and clinician-facing mechanisms (the ability to request human re-review of an AI-augmented diagnosis), and audit trails that permit regulators or quality teams to investigate specific cases retrospectively. Contestability requires that explanations be available, that overrides be possible without workflow penalty, and that disagreements feed back into model monitoring. The EU AI Act explicitly requires contestability for high-risk AI systems \cite{EU2024AIAct}. Evaluation of contestability assesses the availability and usability of these mechanisms within the clinical workflow.
    \end{itemize}

% [Paste here: the full Stage / Type / Scope subsubsections you have
% in current Section 3 — Intrinsic / Post hoc / Model-specific /
% Model-agnostic / Local / Global, plus tab:interpretability_techniques
% renamed to supp:tab:stage_type_scope]

\section{XAI Taxonomy for CompPath: Three Orthogonal XAI Dimensions} \label{supp:taxonomy}
    We organize XAI methods along three orthogonal dimensions: stage, type, and scope (Figure \ref{fig:xai_stage_type_scope}. This framework provides a structured basis for comparing XAI approaches and clarifying common terminology in the literature \cite{kamath2021explainable,adadi2018peeking,murdoch2019definitions}.

% Figure 2: Stage-Type-Scope taxonomy figure
% [Figure goes here]
% \begin{figure}[!t]
%           \centering
%           \includegraphics[width=0.9\textwidth]{figs/XAI_stage_type_scope.png}
%           \caption{Stage-Type-Scope taxonomy for XAI in CompPath. Three orthogonal dimensions characterize each XAI method: Stage (intrinsic vs. post hoc) addresses when the explanation is produced; Type (model-specific vs. model-agnostic) addresses how dependent the explanation is on the model's architecture; and Scope (local vs. regional/slide vs. global) addresses at what level the explanation operates.}
%           \label{fig:block_diagram_xai_stage_type_scope}
%     \end{figure}
    \subsection{Stage of XAI}
    The Stage axis distinguishes methods that are built into the model's own computation (intrinsic) from methods that are applied afterwards to an already-trained model (post hoc).
    
    \subsubsection{Intrinsic}
        Intrinsic XAI is achieved by models whose decision-making processes can be understood directly from their structure. These methods explicitly link inputs to outputs, making them transparent without additional explanation. Examples include linear models, decision trees, and generalized additive models \cite{murdoch2019definitions}. In CompPath, intrinsic XAI has been explored using feature-specific models that predict outcomes based on predefined histological attributes. For instance, methods that quantify tumor-infiltrating lymphocytes (TILs)\cite{shephard2024automated} or compute mitotic counts \cite{aubreville2023mitosis} both of which return quantities a pathologist already reasons about. Recent neural-network designs have introduced forms of intrinsic interpretability through self-explaining networks \cite{alvarez2018towards}, concept-bottleneck and concept-aware architectures \cite{koh2020concept}, and hybrid rule-based designs \cite{kierner2023taxonomy}.
        
        Attention-based multiple instance learning (MIL) sits in an intermediate position \cite{ilse2018attention}. Attention weights are computed during the model's forward pass and are available without an additional explanation procedure, which makes their generation intrinsic in the procedural sense. However, attention weights have been shown to not necessarily reflect feature importance in a faithful or causal way \cite{jain2019attention,wiegreffe2019attention}, so their use as an interpretability claim requires the same kind of faithfulness, stability, and clinical plausibility evaluation that post hoc explanations require. In this review, we therefore describe attention-based MIL as a widely used mechanism for visualizing model focus, but not as a sufficient substitute for explicit explainability evaluation.
            
    \subsubsection{Post hoc}
        Post hoc methods generate explanations by applying a separate procedure to an already-trained model. They are particularly relevant for deep architectures where the model's internal computation is not directly inspectable. Common families include feature attribution methods such as SHAP \cite{lundberg2017unified}, gradient-based visualization methods such as Grad-CAM \cite{Selvaraju_2017_ICCV} that highlight image regions contributing to a prediction, and example-based \cite{poche2023natural} and counterfactual \cite{verma2020counterfactual} explanations that illustrate how small input changes would affect the prediction. Post hoc methods can be applied to any trained model without altering its architecture, but the explanation is an output of the explanation method, not of the model, and may not faithfully reflect the model's internal reasoning. 
    
    \subsection{Type of XAI}
   XAI techniques are classified by their type based on their architectural dependency. Model-specific approaches are tailored to particular architectures and utilize internal model parameters to generate rationales, whereas model-agnostic approaches are universally applicable across heterogeneous pipelines because they infer importance from perturbations rather than internal mechanics.
   
    \subsubsection{Model-specific}
        Model-specific techniques are tailored to specific architectures and use internal components, such as gradients, activations, or attention weights, to generate explanations. In CNNs, for example, saliency maps or class activation mappings highlight spatial regions that influenced predictions \cite{murdoch2019definitions}. In pathology, such methods have been used to highlight nuclear features, stromal patterns, or glandular structures associated with cancer grade. Their advantage is that they leverage model weights and activations directly, but their limitation is narrow applicability; techniques designed for CNNs may not translate to transformers or MIL models.
    
    \subsubsection{Model-agnostic}
        Model-agnostic approaches operate independently of the model's architecture, relying only on inputs and outputs. They can therefore be applied universally, making them attractive for heterogeneous pipelines. Examples include SHAP \cite{lundberg2017unified}, which provides mathematically grounded feature attributions, and perturbation-based methods such as LIME \cite{ribeiro2016should} or occlusion sensitivity \cite{fong2017interpretable}, which systematically alter input regions to infer importance. In pathology, these methods have been applied to WSIs by perturbing patches or altering color channels (e.g., simulating stain variability). Their generality is valuable, but they can be computationally expensive and sensitive to noise.
    
    \subsection{Scope of XAI}
    The scope of explanation distinguishes whether an XAI method provides a local rationale for an individual case or a global overview of the model’s systemic behavior. Local methods identify the specific features responsible for a single prediction, whereas global methods capture the model's underlying logic across an entire dataset or patient population.
    \subsubsection{Global} 
        Global methods capture the model's overall behavior across a dataset. For example, feature importance rankings can reveal which histological features are consistently influential in tumor classification \cite{olden2004accurate}. Concept-based methods such as Testing with Concept Activation Vectors (TCAV) \cite{kim2018interpretability} extend this idea, quantifying the influence of higher-level concepts (e.g., presence of necrosis, gland morphology) on predictions. Global explanations are particularly useful in pathology when the goal is to validate whether a model captures clinically relevant features across populations.

    \subsubsection{Local}                           
        Local methods focus on individual predictions, providing case-specific rationales. These are particularly relevant for clinical settings where pathologists must trust or challenge individual model outputs. Techniques such as SHAP and LIME highlight the image regions or cellular features most responsible for a specific prediction. For example, a local explanation may show that a risk prediction for a single patient was driven by mitotic activity in a tumor hotspot. Local explanations thus align closely with clinical workflows, where decisions are made case by case.

    \begin{table}[!h]
        \centering
        \caption{XAI methods in CompPath categorized by families and the Stage-Type-Scope taxonomy. \footnotesize\textbf{Legend:} Type: Agnostic~$\circ$, Specific~$\bullet$; Scope: Local~$\square$, Global~$\blacksquare$; Stage: Post hoc~$\triangle$, Intrinsic~$\blacktriangle$}
        \label{supp:interpretability_techniques}
        \begin{tabular}{ccccc}
        \hline
        \textbf{Method} & \textbf{Section} & \textbf{Stage} & \textbf{Type} & \textbf{Scope} \\
        \hline
        \multicolumn{5}{c}{\textbf{Backpropagation-based}} \\
        Trainable Attention \cite{vaswani2017attention} & 3.1.1 & $\blacktriangle$ & $\bullet$ & $\square$ \\
        Transformer-based Attention \cite{chefer2021transformer} & 3.1.2 & $\blacktriangle$ & $\bullet$ & $\square$ \\
        CAM \cite{zhou2016learning} & 3.1.3 & $\triangle$ & $\bullet$ & $\square$ \\
        Grad-CAM \cite{Selvaraju_2017_ICCV} & 3.1.4 & $\triangle$ & $\bullet$ & $\square$ \\
        Saliency Maps \cite{mundhenk2019efficient} & 3.1.5 & $\triangle$ & $\bullet$ & $\square$ \\
        LRP \cite{montavon2019layer} & 3.1.6 & $\triangle$ & $\bullet$ & $\square$/$\blacksquare$ \\
        DeepLIFT \cite{shrikumar2017learning} & 3.1.7 & $\triangle$ & $\bullet$ & $\square$ \\
        Deconvolution \cite{zeiler2014visualizing} & 3.1.8 & $\triangle$ & $\bullet$ & $\square$ \\
        Integrated Gradients \cite{qi2019visualizing} & 3.1.9 & $\triangle$ & $\bullet$ & $\square$ \\
        \hline
        \multicolumn{5}{c}{\textbf{Perturbation-based}} \\
        LIME \cite{ribeiro2016should} & 3.2.1 & $\triangle$ & $\circ$ & $\square$ \\
        Occlusion Sensitivity \cite{fong2017interpretable} & 3.2.2 & $\triangle$ & $\circ$ & $\square$ \\
        RISE \cite{petsiuk2018rise} & 3.2.3 & $\triangle$ & $\circ$ & $\square$ \\
        \hline
        \multicolumn{5}{c}{\textbf{Feature-based}} \\
        Activation Layer Visualization (ALV) \cite{yosinski2015understanding} & 3.3.1 & $\triangle$ & $\bullet$ & $\square$ \\
        SHAP \cite{lundberg2017unified} & 3.3.2 & $\triangle$ & $\circ$ /$\bullet$ & $\square$/$\blacksquare$ \\
        \hline
        \multicolumn{5}{c}{\textbf{Concept-based}} \\
        TCAV \cite{kim2018interpretability} & 3.4.1 & $\triangle$ & $\bullet$ & $\square$ \\
        Image Captioning \cite{vinyals2015show} & 3.4.3 & $\triangle$ & $\circ$ & $\square$ \\
        Vision-Language Models \cite{zuo2024plip} & 3.4.4 & $\triangle$ & $\circ$ & $\square$ \\
        Concept-Bottleneck Models \cite{koh2020concept} & 3.4.2 & $\blacktriangle$ & $\bullet$ & $\square$/$\blacksquare$ \\
        \hline
        \multicolumn{5}{c}{\textbf{Example-based}} \\
        Prototypes \cite{chen2019looks} & 3.5.1 & $\triangle$ & $\circ$ & $\square$ \\
        Counterfactuals \cite{stepin2021survey} & 3.5.2 & $\triangle$ & $\circ$ & $\square$ \\
        Trust Scores \cite{jiang2018trust} & 3.5.3 & $\triangle$ & $\circ$ & $\square$ \\
        \hline
        \end{tabular}
    \end{table}

\section{Per-Method Descriptions of XAI Methods in CompPath} \label{supp:methods}
This section provides per-method narrative descriptions for each XAI technique summarised in Table 2 of the main paper.  We conducted a narrative literature review of XAI methods in CompPath: studies were included if they applied, developed, or evaluated XAI methods for histopathology or whole-slide image analysis, and excluded if they addressed general medical imaging without pathology-specific data or did not provide an explanation-related method or evaluation. A formal systematic review under PRISMA \cite{page2021prisma} was considered but deemed inappropriate for the scope of this work for three reasons. First, the field of XAI in CompPath is methodologically heterogeneous: studies vary widely in task (classification, segmentation, prognosis), tissue type, model architecture, and XAI evaluation criteria, which precludes the comparable outcome measures that systematic reviews and meta-analyses require. Second, our objective is conceptual synthesis, namely organizing methods into a unified taxonomy, terminology, and task-driven framework, rather than estimating a pooled effect or comparing reported performance. Third, the literature is evolving rapidly, with foundation models, vision-language models, and concept-bottleneck designs introduced after most systematic-search windows would have closed. We therefore adopted a narrative review approach following established guidance for non-systematic literature reviews in fast-moving fields \cite{greenhalgh2018time,sukhera2022narrative}. The tables consequently present representative rather than exhaustive coverage, prioritizing methodological diversity, anatomical breadth, and clinical relevance over completeness. Each description follows a uniform structure: operating principle, strengths, limitations, computational and practical considerations, current adoption in CompPath, and recommended validation practices.

{\footnotesize
\setlength{\tabcolsep}{3pt}
\renewcommand{\arraystretch}{1.08}
\setlength{\LTcapwidth}{\textwidth}

\begin{longtable}{@{}%
L{0.145\textwidth}
L{0.275\textwidth}
L{0.275\textwidth}
L{0.255\textwidth}@{}}

\caption{Summary of backpropagation-based XAI methods in CompPath: strengths, limitations, and recommended practices for validation and reporting.}
\label{tab:xai_backprop_summary}\\

\toprule
\textbf{Method} & \textbf{Strengths} & \textbf{Limitations} & \textbf{Cautions \& best practices} \\
\midrule
\endfirsthead

% No repeated caption or column headings on continuation pages
\endhead

% No continuation footer
\endfoot

\bottomrule
\endlastfoot

\textbf{Attention} \newline (MIL / Transformer) &
Highlights decision-driving patches; interpretable at the instance level; integrates naturally with weakly supervised slide-level learning. &
Attention weights capture correlation, not causation; sensitive to site-specific bias, stain, and scanner artefacts; diffuse across heads/layers in transformers. &
Validate attention maps against expert ROIs and morphometric features; test stability across scanners, stains, and institutions \cite{jahanifar2025domain}. \\ \hline

\textbf{CAM} &
Simple, fast, intuitive; identifies broad discriminative regions with minimal overhead. &
Requires a GAP layer; coarse spatial resolution misses subtle nuclear or sub-cellular features. &
Use for rapid qualitative visualisation; confirm findings with higher-resolution methods or expert annotations. \\ \hline

\textbf{Grad-CAM} &
Architecture-agnostic; widely supported; produces visually intuitive, class-specific localisations on standard CNNs. &
Coarse heatmaps; poor at localising multiple instances of the same class; gradients can be attenuated or distorted under MIL pooling; no signed relevance. &
Report the target convolutional layer; run sanity checks \cite{adebayo2018sanity}; verify faithfulness via insertion/deletion \cite{petsiuk2018rise} and cross-site stability \cite{jahanifar2025domain}. \\ \hline

\textbf{Grad-CAM++} &
Better multi-instance localisation than Grad-CAM through higher-order gradient weighting; handles multiple objects in a single image. &
Same correlational limitation as Grad-CAM; sensitive to gradient noise in very deep networks. &
Use when multiple instances of the target class are expected; benchmark against Grad-CAM on the same model; verify cross-site stability \cite{jahanifar2025domain}. \\ \hline

\textbf{Saliency maps} &
Pixel-level sensitivity; single-pass computation per tile; useful as a baseline against which newer methods are evaluated. &
Noisy and unstable; highly sensitive to preprocessing, stain normalisation, and scanner colour variation; often emphasises edges over morphology. &
Smooth or combine with IG; assess stability under stain/rotation perturbations; confirm overlap with expert annotations. \\ \hline

\textbf{SmoothGrad} &
Reduces high-frequency noise in saliency by averaging over noise-perturbed inputs; can wrap around any gradient-based method; produces spatially coherent maps. &
$N$ extra forward and backward passes per attribution; sensitive to noise level $\sigma$ and sample count $N$; can over-smooth and lose fine morphological detail. &
Report $\sigma$ and $N$; benchmark against vanilla saliency; verify cross-site stability \cite{jahanifar2025domain}. \\ \hline

\textbf{FullGrad} &
Aggregates per-layer gradient and bias contributions; satisfies a completeness axiom; higher spatial resolution than Grad-CAM with class-discriminative signal. &
Higher compute and memory cost than Grad-CAM; sensitive to post-processing choices; requires architectures with explicit bias terms. &
Report post-processing operations; benchmark against Grad-CAM and IG on the same model; assess cross-site stability \cite{jahanifar2025domain}. \\ \hline

\textbf{LRP} &
Conserves relevance through layers; produces stable, fine-grained attributions; adaptable to MIL and transformer architectures. &
Computationally heavy; results depend on the chosen propagation rule and architecture-specific implementation. &
Report propagation rules and parameter choices; benchmark against Grad-CAM and perturbation-based faithfulness metrics; evaluate consistency across cohorts and scanners \cite{jahanifar2025domain}. \\ \hline

\textbf{DeepLIFT} &
Captures signed contributions; handles feature dependencies often missed by simple gradients. &
Results vary substantially with the choice of baseline or reference input; implementation details affect outputs. &
Test multiple baseline choices; verify the sum-to-delta property; confirm consistency with IG and occlusion analyses. \\ \hline

\textbf{Deconvolution} &
Reveals what individual filters and layers have learned; useful for debugging and understanding learned representations. &
Produces generic, prototypical textures rather than case-specific attribution; layer-dependent reconstructions. &
Use for representation analysis and model debugging; do not present as the sole basis for case-level clinical justification. \\ \hline

\textbf{Integrated Gradients} &
Axiomatic guarantees; reduces gradient noise; widely used as a faithfulness reference. &
Sensitive to baseline definition and number of integration steps; computationally heavy when applied tile-by-tile across WSIs. &
Report baseline and integration step count; evaluate faithfulness via insertion/deletion and region perturbation; test robustness across staining and scanner conditions \cite{jahanifar2025domain}. \\

\end{longtable}
}-

\subsection{Backpropagation-based methods} \label{supp:methods_backprop}
    Backpropagation-based methods are typically of post hoc stage, model-specific type, and primarily local scope. They compute gradients of model outputs with respect to input features, identifying pixels or regions guiding predictions. In CompPath, backpropagation-based approaches that compute gradients are often used to identify histological ``hotspots'', which can be linked to tissue morphology. 

    \subsubsection{Trainable attention \cite{vaswani2017attention}} has been a widely applied method in CompPath, particularly with weakly supervised MIL models \cite{lu2021data,campanella2019clinical}. In such models, WSIs are partitioned in a set of image patches (instances), and attention weights are learned to indicate the relative contribution of each patch to slide-level prediction. The attention mechanism thus enables both classification and localization of patches that contributed most in the model's decision. A widely used example of such models is ABMIL \cite{ilse2018attention}, which generates heatmaps based on attention pooling to highlight morphological subregions that drive model decisions without the need for pixel-level supervision.
    
    Strengths of the `trainable attention' include the i) ability to identify influential patches using only slide-level labels, ii) suitability for MIL settings, and iii) generally greater stability and data efficiency compared to gradient-based explanation methods. However, several limitations and pitfalls remain, as i) their high attention weights do not necessarily imply causal importance, ii) the learned weights can be influenced by dataset specific biases (e.g., stain or scanner variations), and iii) attention maps often lack cellular-level precision where high attention scores in positive prediction are more interpretable rather than high attention in negative cases. Furthermore, prior studies have shown that attention visualizations may only be correlations rather than the true reasoning process of the model \cite{jain2019attention,wiegreffe2019attention}. From a computational and practical standpoint, trainable attention is efficient at inference, as attention weights are obtained during the forward pass and can be exported for visualization with minimal additional computational cost.
    
    In terms of adoption and readiness, trainable attention mechanisms are widely used in research-oriented MIL-based pathology models \cite{lu2021data,campanella2019clinical}. Demonstrated applications include predicting clinically relevant outcomes, such as the identification of conclusive diagnoses in glioma \cite{innani2025ai}. Despite this growing use, standardized evaluation protocols and regulatory-grade validation remain limited, hindering direct clinical deployment. Recommended validation practices include assessing overlap between attention-highlighted regions and expert-annotated areas or quantitative morphometric features (e.g., nuclear density or glandular organization). Robustness should be evaluated across cohorts, scanners, and staining variations, and quantitative sanity checks, such as attention perturbation or ablation studies, are encouraged to assess the stability, robustness, and reliability of the explanations.
    
    \subsubsection{Transformer-based Attention \cite{chefer2021transformer}} provides an intrinsic explanation through the self-attention weight computed during the forward pass. These weights encode contextual dependencies among image patches and captures relationships between spatially distinct regions using multi-head self-attention, enabling models to reason over local and global tissue context.

    Strengths of `transformer-based attention' include its inherent ability to i) provide hierarchical explanations, ii) model both cellular detail and higher-order tissue organization, and iii) naturally integrate contextual reasoning, e.g., tumor-immune spatial patterns. However, it is not without limitations and pitfalls as i) attention weights indicate correlation rather than causation, ii) multiple heads and layers produce complex diffuse attention maps that are difficult to interpret. In addition, transformers incur high computational and practical costs, particularly during training, due to the large token counts involved in processing WSIs. Hierarchical designs (e.g., HIPT \cite{chen2022scaling}) mitigate cost by structuring aggregation across scales, though visualization still requires resource-intensive layer aggregation.
    
    Regarding adoption and readiness, these mechanisms are increasingly used in pathology research (e.g., HIPT \cite{chen2022scaling}). Despite promising performance, attention-based visualization remains research-focused and lacks standardized clinical validation. Recommended validation strategies include i) inspecting individual attention heads and layers independently before aggregation, ii) correlating attention hotspots with expert annotations and quantitative features (e.g., cell density, immune infiltration), and iii) using gradient-based or prototype-based cross-validation (e.g., Chefer explainability \cite{chefer2021transformer}) to ensure faithfulness.

    \subsubsection{Class Activation Mapping (CAM)\cite{zhou2016learning}} is one of the earliest and widely used visualization techniques for explaining CNN predictions. In CompPath, CAM identifies image regions that contribute to a model's decision by linearly weighting feature maps from the final convolutional layer using class-specific activation weights. The method relies on Global Average Pooling (GAP) layer, which aggregates spatial information before classification. The resulting weighted activation maps are projected back onto the input space to create coarse heatmaps that highlight discriminative tissue regions such as tumor epithelium or necrotic foci.
    
    Strengths of CAM include its conceptual simplicity, minimal architectural requirements, and the production of intuitive/easy-to-interpret heatmaps, making it useful for quick verification that the model's focus aligns with pathological structures. However, notable limitations/pitfalls include i) the requirement for a GAP layer, ii) low spatial resolution maps, which may miss small or subtle histological structures, and iii) performance sensitivity to the design of the final convolutional layer and network depth. From a computational and practical perspective, CAM i) incurs very low cost, ii) does not require any architectural modifications, beyond GAP, and iii) can be applied directly from the trained model.
    
    In terms of adoption and readiness, CAM has been widely used in research (including for visualizing molecular subtype localization in bladder cancer \cite{woerl2020deep}), primarily as a qualitative interpretability tool rather than a quantitative or regulatory standard. Recommended validation practices include employing CAM for rapid, high-level localization in GAP-based CNNs and confirming highlighted areas with higher-resolution methods (e.g., Grad-CAM or LRP) or pathologist annotations, when available.
    
    \subsubsection{Gradient-weights Class Activation Mapping (Grad-CAM)\cite{Selvaraju_2017_ICCV}} is a widely used post hoc method that generates class-specific heatmaps by leveraging the gradients of a target output (e.g., class score) with respect to the feature map in CNNs. 
    
    In terms of strengths, compared with CAM, Grad-CAM removes the dependence on GAP layers (i.e., is architecture-agnostic). By weighting feature maps according to the gradient intensity associated with a target class, Grad-CAM highlights regions of an image that most strongly influence a model’s decision, offering a coarse but visually intuitive spatial explanation. Grad-CAM's primary advantage in the pathology workflow is practical, as it i) does not require any architectural modifications and is compatible with existing CNN-based pipelines, ii) adds low computational overhead, and iii) produces heatmaps that pathologists can inspect visually. Limitations and pitfalls are particularly relevant in CompPath and MIL settings, as gradient signals can be attenuated or distorted by instance-level pooling, resulting in misleading explanations. Grad-CAM may also highlight background regions or imaging artifacts, and its faithfulness can vary across datasets and tasks. From a computational and practical standpoint, Grad-CAM i) incurs moderate inference cost, ii) requires gradient computation and spatial upsampling, and iii) needs careful selection of the target convolutional layer to obtain localization of clinically relevant regions.
    
    Regarding adoption and readiness, Grad-CAM is commonly used in research, e.g., metastasis localization \cite{ji2019gradient}. Adaptations (e.g., MILPLE \cite{sadafi2023pixel}) extend it to instance-level MIL, despite reported inconsistencies in spatial visual explanations due to disrupted gradient flow. Recommended validation includes i) explicitly reporting the chosen target layer, ii) performing sanity checks to verify dependence on trained model parameters, by randomizing weights/labels and ensuring the maps change accordingly \cite{adebayo2018sanity}, iii) evaluating pixel-flipping tests and insertion/deletion metrics, to assess whether removing or adding Grad-CAM-highlighted regions meaningfully affects model confidence \cite{petsiuk2018rise}, and iv) cross-validate against expert annotations.

    \subsubsection{Grad-CAM++ \cite{chattopadhay2018grad}}
    Grad-CAM++ extends Grad-CAM to better localise multiple instances of the same object class in an image, which is a known weakness of the original Grad-CAM in CompPath applications where a WSI can contain many tumour foci, mitotic figures, or immune infiltrates simultaneously. Grad-CAM++ uses higher-order gradient information (weighted combinations of positive partial derivatives) to compute pixel-wise weights, producing sharper localisations and better multi-instance coverage than Grad-CAM. It has been applied in CompPath to lung histology classification \cite{han2022multi}. Limitations follow Grad-CAM's (coarse resolution, correlation not causation), with additional sensitivity to gradient noise in deep networks.

    \subsubsection{Saliency Maps \cite{simonyan2013deep}} are among the earliest gradient-based visualization methods for model explanation. This method estimates how sensitive a model’s output is to each input pixel by computing the gradient of the prediction score with respect to the input image. The resulting heatmap highlights pixels where small perturbations would induce the largest change in the output, thereby indicating regions most influential to the model's decision.
    
    Strengths of saliency maps include their extremely fast computation, while able to produce fine-grained, high-resolution maps, and their use as a reference baseline for evaluating newer XAI methods. However, significant limitations and pitfalls limit their utility in CompPath, as they are very sensitive to small image variations (e.g., such as slight color changes, scanner noise) and preprocessing steps (e.g., normalization or stain variation), which often result in noisy visualizations emphasizing edges or imaging artifacts rather than meaningful histological features. From a computational and practical perspective, saliency maps are fast to generate, and their results are often post-processed using smoothing or filtering to reduce visual noise to improve visualization.
    
    With respect to adoption and readiness, saliency maps are commonly used in research for patch-level interpretability in pathology (e.g., highlighting nuclei or tumor regions \cite{bejnordi2017diagnostic}), but are generally insufficient on their own for quality assurance or clinical reporting due to instability and a lack of quantitative validation. Recommended validation practices include i) assessing stability under small perturbations of the same patch (e.g., stain variation, rotation), ii) comparing results with more robust attribution methods, such as Integrated Gradients or occlusion-based methods, and iii) visually confirming overlap with expert annotations.

    Two extensions of vanilla saliency address its main weaknesses and are widely used in CompPath. SmoothGrad \cite{smilkov2017smoothgrad} mitigates the high-frequency noise in input gradients by averaging saliency maps over multiple noise-perturbed copies of the input ($x + \mathcal{N}(0, \sigma^2)$), producing visually cleaner and more spatially coherent attributions at the cost of a forward and backward pass per noise sample. It has been used in CompPath to generate denoised saliency maps for prostate cancer Gleason-grading \cite{dolezal2023deep} and is also commonly applied as a wrapper around Grad-CAM and Integrated Gradients rather than vanilla saliency. Full-Gradient representation (FullGrad) \cite{srinivas2019fullgrad} addresses the complementary problem of spatial coarseness by decomposing the model's output into per-layer contributions from the input gradient and the gradients of all intermediate bias terms; the resulting attribution map satisfies a completeness property (the sum of per-pixel attributions equals the model's scalar output) and produces visually sharper, more class-discriminative maps than vanilla saliency at higher compute cost than Grad-CAM but lower than Integrated Gradients. FullGrad has been applied in CompPath to breast cancer histology classification \cite{mondol2025graphite} and to interpretable prediction of tumor microenvironment phenotypes from H\&E whole-slide images \cite{chen2026interpretable}. The cautions that apply to vanilla saliency apply to both extensions: report post-processing operations (for SmoothGrad: noise level $\sigma$ and number of samples $N$; for FullGrad: absolute value, channel aggregation, and normalization), benchmark against Grad-CAM and Integrated Gradients on the same model, and verify cross-site stability of the resulting maps, since fine spatial detail can otherwise reflect scanner- or stain-specific texture rather than tissue biology \cite{jahanifar2025domain}.

    %This is summary of subsection 3.1.3-3.1.5
    % Although saliency maps, CAM, and Grad-CAM are all gradient-based visualization techniques, they differ in what they represent and how they are computed. Saliency maps compute first-order derivatives directly with respect to input pixels, producing high-resolution but often noisy sensitivity maps that show where small pixel-level changes affect the prediction. In contrast, CAM and Grad-CAM operate at a higher level of abstraction, using activations from the last convolutional layers (and in Grad-CAM, weighting them by class-specific gradients) to generate lower-resolution but more semantically meaningful region-level heatmaps. Thus, saliency maps provide fine-grained but unstable explanations, while Grad-CAM offers coarser yet more interpretable localization, emphasizing biologically coherent regions (e.g., tumor epithelium or necrotic foci). In practice, Grad-CAM has largely superseded saliency mapping in pathology due to its greater visual coherence and closer alignment with pathologist-defined regions of interest \cite{selvaraju2017gradcam,woerl2020deep,bejnordi2017diagnostic}.

    \subsubsection{Layer-wise Relevance Propagation (LRP)\cite{bach2015pixel}} is a backpropagation-based explainability method that redistributes a model’s output prediction score backward through the network to assign relevance values to individual input features. Relevance is propagated proportionally from each neuron to its inputs, based on predefined conservation rules, ensuring that total relevance is preserved across layers. This principle of ``relevance conservation'' distinguishes LRP from standard gradient-based methods and enables more stable attribution maps that explain why a model made a particular decision, rather than merely indicating regions with large gradients.
    
    Strengths of LRP include its explicit conservation property, which yields stable and fine-grained explanations, and its ability to attribute and propagate relevance across layers, even in complex MIL or transformer architectures. LRP has also been shown to provide faithful and visually interpretable explanations, and in some settings to even outperform Grad-CAM for MIL-based interpretability \cite{hense2024xmil}. Nevertheless, LRP is sensitive to the choice of propagation rule (e.g., $\epsilon$-rule, $\alpha\beta$-rule) and implementation requires network-specific adaptation, especially for MIL or transformer blocks. From a computational and practical perspective, LRP is computationally more demanding than simple gradient-based methods mainly due to relevance propagation through all layers, and requires comprehensive documentation of parameter settings (e.g., rule choice, normalization) towards facilitating reproducibility.
    
    In terms of adoption and readiness, LRP has successfully demonstrated slide-level interpretability in recent work, such as xMIL-LRP \cite{hense2024xmil} for breast and lung cancer WSIs, but remains limited to research use. Recommended validation practices include i) explicitly reporting the propagation rules and parameter choices, ii) comparing relevance maps with complementary methods, such as Grad-CAM, Integrated Gradients, and perturbation-based faithfulness metrics (insertion/deletion, ROAR), and iii) evaluating overlap with expert ROIs and consistency across architectures and datasets.

    \subsubsection{Deep Learning Important FeaTures (DeepLIFT)\cite{shrikumar2017learning}} explain model predictions by decomposing output differences relative to a predefined reference (baseline) input. Instead of relying on local gradients, DeepLIFT attributes importance by comparing each neuron's activation to its reference activation and propagating contribution scores backward through the network. This approach enables signed attributions that indicate which input features increase or decrease the model's confidence and supports both pixel-level and patch-level visualizations.
    
    Strengths of DeepLIFT include its i) ability to distinguish positive and negative contributions to a prediction, ii) capacity to capture feature dependencies that are often missed by simple gradient methods, and iii) improved stability compared with plain saliency maps in some CNN architectures. However, notable limitations and pitfalls arise from its strong dependence on the choice of baseline or reference image (for example, a blank patch), as well as on implementation details such as scaling and normalization, which affect the results. From a computational and practical perspective, DeepLIFT incurs a moderate computational cost, comparable to a single backward pass per patch once implemented, but careful verification and reporting of the selected baseline and the ``sum-to-delta'' property (ensuring all contributions add up to the prediction difference) are essential for reproducibility. 
    
    Regarding adoption and readiness, DeepLIFT has been applied in research settings for patch-level analysis of colon cancer WSIs \cite{hossain2022early}, but it has not yet been standardized for WSI or clinical workflows. Recommended validation practices include i) evaluating sensitivity across multiple baseline choices, ii) confirming consistency with related attribution methods such as integrated gradient or occlusion analyses, and iii) visually assessing whether highlighted structures correspond to known histologic features. 

    %Summary of subsection 3.1.6-3.1.7
    %While Grad-CAM provides coarse, feature-map-level visualizations and Integrated Gradients offer path-integrated attribution based on gradients, LRP differs in that it explicitly redistributes prediction scores across network layers according to layer-specific rules. This approach preserves both magnitude and direction of contribution, enabling a more faithful representation of the model’s reasoning. Consequently, LRP tends to produce smoother, semantically consistent explanations that align better with histological structures, making it especially effective for WSI-level interpretability in pathology.
    
    \subsubsection{Deconvolution \cite{zeiler2014visualizing}} visualizes internal representations of CNN by approximately reversing the sequence of convolution, pooling, and activation operations to reconstruct the input pattern that strongly activates specific filters or neurons. This approach produces patch-level visualizations of intermediate feature activations, highlighting textures, edges, or structural patterns learned by the network rather than directly explaining individual predictions.
    
    Strengths of deconvolution include its ability to provide insight into what different filters or layers of a CNN have learned, making it useful for debugging network behavior and understanding how histological features, such as nuclei or tissue textures, are internally represented. However, limitations and pitfalls include limited interpretability in decision-level settings. More fundamentally, deconvolution addresses a different question than attribution-based methods: it reveals what filter has learned to detect, not why a specific prediction was made, i.e., they often show generic or ``prototypical'' textures learned by the model rather than evidence specific to a particular case. Moreover, visualizations may vary depending on the chosen layer or reconstruction method. From a computational and practical perspective, deconvolution incurs moderate computational cost, requires access to intermediate activations, and can generate large visualization outputs for deep architectures. 
    
    In terms of adoption and readiness, deconvolution has been used in research for qualitative inspection in lung cancer \cite{coudray2018classification}. However, it is best suited for exploratory analysis and model debugging rather than clinical justification. Recommended validation involves using deconvolution to interpret learned features and complementing it with attribution-based or perturbation-based methods, when explaining individual predictions or supporting clinical conclusions.
    
    \subsubsection{Integrated Gradient(IG)\cite{qi2019visualizing}} attributes model predictions by integrating the gradients of the output with respect to the model input along with a continuous path from a predefined baseline to the actual input. This path-based integration accumulates attribution scores across interpolation steps, yielding smoother and more stable explanations than single-step gradient methods.
    
    Strengths of IG include its i) strong theoretical foundation, through the axioms of sensitivity and implementation invariance, and ii) ability to produce quantitatively interpretable attributions that reduce gradient noise. As a result, IG is widely recognized as a benchmark for evaluating faithfulness in neural explanations. Nevertheless, IG has several limitations and pitfalls that require careful selection of both the baseline and the number of integration steps: an inappropriate baseline choice can distort attributions, and too few steps can yield incomplete integration and leave results sensitive to staining variability and scanner differences. From a computational and practical standpoint, IG incurs substantially higher computational cost than single-step gradient approaches (e.g., saliency maps, Grad-CAM), as it is typically applied per tile or patch and then aggregated into a slide-level map. Moreover, its reproducibility depends on consistent baseline selection and integration step settings.
    
    In terms of adoption and readiness, IG has been used in research to highlight metastasis-relevant areas in lymph node WSIs \cite{rahnfeld2024comparative}. It is often treated as a reliable reference for benchmarking explainability methods, but it is not yet standardized for prospective or regulatory workflows due to its scalability and interpretive complexity. Recommended validation practices include i) explicit reporting of baseline choices (e.g., integration steps and implementation details), ii) evaluating faithfulness using insertion/deletion and region perturbation metrics, and iii) comparing attributions against expert-defined regions of interest across scanners and staining conditions.
    %     \item \underline{Limitations/pitfalls}: Requires careful selection of the baseline and the number of interpolation steps, poor baseline choice can distort attributions, while too few steps can yield incomplete integration. Computationally demanding for WSIs due to multiple forward and backward passes; still sensitive to staining variability and scanner differences.

    \subsection{Perturbation-based methods} \label{supp:methods_perturbation}
    {\footnotesize
\setlength{\tabcolsep}{3pt}
\renewcommand{\arraystretch}{1.08}

\begin{table}[!htbp]
\centering
\caption{Summary of perturbation-based XAI methods in CompPath: strengths, limitations, and recommended practices for validation and reporting.}
\label{tab:xai_perturbation_summary}
\begin{tabular}{@{}L{0.145\textwidth}L{0.275\textwidth}L{0.275\textwidth}L{0.255\textwidth}@{}}
\toprule
\textbf{Method} & \textbf{Strengths} & \textbf{Limitations} & \textbf{Cautions \& best practices} \\
\midrule

\textbf{LIME} &
Model-agnostic; provides an intuitive local surrogate of black-box behaviour. &
Unstable across runs; sensitive to kernel width, segmentation, and sampling; standard pixel perturbations are not biologically plausible in WSIs. &
Use meaningful superpixel or tissue-region segmentations; report kernel width and sampling parameters; assess stability via rank correlation across repeated runs. \\ \hline

\textbf{Occlusion sensitivity} &
Interventional test of region importance; gradient-free; conceptually simple. &
Slow; sensitive to mask size and shape; zero-masking introduces non-tissue artefacts that distort attribution. &
Use biologically plausible perturbations such as blur, inpainting, or stain jitter; test multiple mask sizes; report insertion/deletion curves and cross-site consistency \cite{jahanifar2025domain}. \\ \hline

\textbf{RISE} &
Model-agnostic Monte Carlo estimator; averages noise across many random masks; extends to transformers via token sampling. &
Requires thousands of mask samples for convergence; coarse spatial resolution; computationally intensive at WSI scale. &
Report number, size, and stride of masks; cross-validate against occlusion or IG; evaluate overlap with expert-defined regions. \\

\bottomrule
\end{tabular}
\end{table}
}

    Perturbation-based methods are typically post hoc and model-agnostic. They usually provide local explanations, but their results can be aggregated across patches, slides, or cohorts to support global analysis. It explains model behavior by systematically modifying input data and observing the impact on predictions. These approaches do not rely on gradients, making them model-agnostic. By perturbing input data, these methods help identify which features or components of the input most influence the model's predictions. Perturbation methods can also help uncover biases in the model by revealing how specific alterations to the input data affect the output. Below, we describe various perturbation-based techniques.
    
    \subsubsection{Local Interpretable Model-agnostic Explanations (LIME) \cite{ribeiro2016should}} explains individual predictions by locally approximating the decision surface of a black-box model with an interpretable surrogate, most commonly a sparse linear regression. For a given instance, LIME generates perturbed samples by modifying subsets of input features, queries the original model for corresponding predictions, and fits the surrogate model using weights based on similarity to the original instance. The resulting surrogate coefficients quantify the local contribution to the prediction. LIME is typically applied at the patch or region level, and explanations from multiple tiles can be aggregated to derive slide-level insights.
    
    Strengths of LIME include its model-agnostic nature and its ability to provide intuitive interpretability via simple local surrogate models without requiring access to gradients. However, its use in CompPath is limited because it is highly sensitive to segmentation choices, kernel width, and perturbation design, and its explanation fidelity strongly depends on the sampling strategy and surrogate regularization. Furthermore, standard implementations often produce unstable explanations across runs and non-realistic tissue perturbations that limit biological interpretability \cite{alvarez2018towards}. From a computational and practical perspective, LIME incurs moderate computational cost and requires hundreds of forward passes per instance, which limits scalability to gigapixel WSIs and necessitates careful tiling strategies. 
    
    With respect to adoption and readiness, LIME has been used in research for patch-level interpretability in tasks, such as breast cancer classification and lymph node metastasis detection \cite{peta2024explainable,palatnik2019local}. Although stabilized variants such as SLICE\cite{bora2024slice} and BayLIME\cite{zhao2021baylime} have demonstrated improved reproducibility and faithfulness at the expense of computational overhead, they are currently limited to experimental and research applications. Recommended validation practices include explicitly reporting kernel width, segmentation strategy, and sampling parameters. Other recommendations comprise i) assessing stability across repeated runs using rank correlation, ii) evaluating explanation fidelity with insertion/deletion or Area Over the Perturbation Curve (AOPC) curves, and iii) verifying overlap with expert-defined ROIs while ensuring that perturbations remain histologically plausible.

    \subsubsection{Occlusion Sensitivity (OS)\cite{fong2017interpretable}} evaluates the importance of image regions by systematically occluding different parts of the input and measuring the resulting change in model predictions. By directly observing how model confidence varies when specific regions are hidden, OS identifies areas of an image that most strongly influence the decision, offering an intuitive and interventional form of model-level sensitivity analysis. 
    
    Strengths of OS include its direct assessment of feature importance through controlled perturbations and its independence from model architectures or gradient information. However, OS is sensitive to the design and size of occlusion masks, since larger masks may remove multiple features, whereas smaller ones can amplify noise and produce unstable estimates. In addition, the use of non-biological occlusions (e.g., zeroing, solid colors) produces unrealistic tissue textures, leading to misleading results. From a computational and practical perspective, OS incurs a high computational cost that scales with the number and granularity of occluded regions, making it feasible only for patch subsets or representative tiles. Realistic perturbations (e.g., blur, inpainting) and efficient patch batching can improve OS's scalability.
    
    Regarding adoption and readiness, OS is commonly used in research to assess region importance in prostate cancer and other histopathology tasks \cite{gallo2023shedding}. It is currently a validation tool rather than a clinically validated method, although extended frameworks like HIPPO \cite{kaczmarzyk2024explainable} aim to provide clinically meaningful causal insights. Recommended validation practices include i) employing biologically plausible perturbations such as blurring or inpainting instead of zero masking, ii) testing multiple mask sizes to assess stability, iii) quantifying faithfulness using insertion/deletion metrics, and iv) confirming consistency of highlighted regions across institutions, scanners, and staining protocols.
    
    \subsubsection{Randomized Input Sampling for Explanation (RISE) \cite{petsiuk2018rise}} estimates pixel-level importance by generating random binary masks that stochastically occlude a subset of the input image regions and observing resulting changes in the model's predictions. For a given image, RISE creates thousands of random masks, each hiding a different random subset of pixels. The model is evaluated on each masked image, and the final importance map is computed as a weighted sum of the masks, with each mask’s contribution proportional to the model's predicted confidence for that masked input. Extensions such as Transformer Input Sampling \cite{englebert2023explaining} adapt RISE to vision transformers (ViTs) by sampling a subset of attention tokens rather than pixels, thereby improving patch-level localization.
    
    Strengths of RISE include its model-agnostic formulation, which does not rely on gradients or model internals; its unbiased Monte Carlo-based estimation of feature importance; and its inherent noise reduction through aggregation over randomly perturbed masks. However, RISE typically requires thousands of random masks to achieve convergence, resulting in substantial computational overhead. Moreover, random masking patterns may not correspond to realistic tissue regions, and the effective spatial resolution is limited by the mask size and stride. From a computational and practical standpoint, RISE is computationally intensive, as each masked input requires a forward pass. Efficiency can be improved through batching, GPU parallelization, or stratified sampling. Coarse-grained RISE variants can balance fidelity and runtime for WSI-scale models. 
    
    In terms of adoption and readiness, RISE has been used primarily to compare explainability methods, including for lymph node metastasis classification \cite{rahnfeld2024comparative}, and remains a research-level approach due to its high computational demand. Transformer variants (e.g., ViT-RISE) show promise for scalability in multi-head attention networks but still require systematic validation. Recommended validation practices include i) clearly reporting the number, size, and stride of masks, ii) comparing explanations with complementary occlusion-based and gradient-based methods (e.g., OS and IG), iii) evaluating faithfulness using insertion/deletion or ROAR metrics, and iv) verifying overlap between generated attribution maps and expert-defined regions.
    
    \subsection{Feature-based methods} \label{supp:methods_feature}
    {\footnotesize
\setlength{\tabcolsep}{3pt}
\renewcommand{\arraystretch}{1.08}

\begin{table}[!htbp]
\centering
\caption{Summary of feature-based XAI methods in CompPath: strengths, limitations, and recommended practices for validation and reporting.}
\label{tab:xai_feature_summary}
\begin{tabular}{@{}L{0.145\textwidth}L{0.275\textwidth}L{0.275\textwidth}L{0.255\textwidth}@{}}
\toprule
\textbf{Method} & \textbf{Strengths} & \textbf{Limitations} & \textbf{Cautions \& best practices} \\
\midrule

\textbf{SHAP} &
Additive, theoretically grounded attributions; bridges local and global explanations; interpretable at biologically meaningful feature scales. &
Computationally expensive, especially Kernel SHAP; results depend on feature definition and background dataset; noisy at the pixel level. &
Define features at biologically interpretable scales such as cells, glands, and tissue regions; report background set and sampling strategy; compare global SHAP rankings with established pathology markers. \\ \hline

\textbf{Activation layer visualisation (ALV)} &
Reveals hierarchical feature learning across CNN layers; useful as a representation-analysis and debugging tool. &
Qualitative only; interpretation is subjective and observer-dependent; storing feature maps is memory-intensive at the WSI scale. &
Correlate activated channels with morphometric features; assess consistency across staining protocols and scanner domains \cite{jahanifar2025domain}; do not present as case-level evidence. \\

\bottomrule
\end{tabular}
\end{table}
}
    Feature attribution and representation analysis methods are typically at a post hoc stage, can be model-specific or model-agnostic, and can provide a local-to-global scope.
    Feature attribution and representation analysis explain model predictions by analyzing the contribution or importance of individual features in the input data. These methods quantify the influence of a particular feature on a model's output, providing insights into which input elements drive specific predictions. Unlike perturbation or backpropagation-based techniques, feature-based methods often rely on statistical analyses, feature attribution scores, or model-inherent properties to determine feature relevance. 
    
    \subsubsection{Activation Layer Visualization (ALV)\cite{yosinski2015understanding}} is one of the earliest model-specific explainability methods designed to examine internal feature representations. ALV inspects the activation (feature maps) produced at different layers of the convolutional neural network (CNN) when an input image is propagated forward, revealing how information is progressively transformed across the network. 
    
    Strengths of ALV include its ability to expose the hierarchical construction of representations, from simple textures and edges in early layers to more complex morphological patterns in deeper layers, making it useful for understanding what types of features a model has learned. However, ALV provides qualitative rather than quantitative attributions, and interpretation can be subjective and dependent on the researcher's domain knowledge. Furthermore, individual feature maps can be selectively interpreted, potentially biasing conclusions. From a computational and practical standpoint, ALV is lightweight, requiring only a forward pass to extract activations, but storing large numbers of feature maps from WSI-scale data can be memory-intensive.
    
    In terms of adoption and readiness, ALV has primarily been used in research settings as an audit tool rather than as a clinical interpretability method. Example studies include visualizing learned feature hierarchies in breast cancer histology \cite{sitaula2020fusion}. Recommended validation practices include correlating activated channels with morphometric features (nuclear density, gland) and testing consistency across staining protocols and scanner domains to assess robustness.
    
    \subsubsection{SHapley Additive exPlanations (SHAP)}\cite{lundberg2017unified} is a unified method for attributing the contribution of individual features to a model's prediction using Shapley values from cooperative game theory. It quantifies how much each ``feature player'' contributes to shifting the model's output away from a baseline expectation by averaging its marginal contribution across all possible feature combinations. In practice, these values are approximated through either model-agnostic (Kernel SHAP) or model-specific implementations such as Tree SHAP and Deep SHAP.
    
    Strengths of SHAP include its strong theoretical guarantees, its ability to produce additive attributions that satisfy local accuracy and missingness axioms, and its ability to bridge local (instance-level) and global (cohort-level) explanations. On the other hand, kernel SHAP is computationally expensive, and accuracy depends on meaningful feature definitions and background datasets. When applied directly to raw pixels, SHAP attributions can appear noisy or biologically implausible, and naive aggregation of feature attributions without proper normalization may distort clinical interpretation. From a computational and practical perspective, SHAP incurs moderate-to-high computational cost. It is tractable for tabular or region-based pipelines, but becomes prohibitively expensive for pixel-level image explanations. Consistent baseline data selection and feature scaling are therefore essential for reproducibility.
    
    In terms of adoption and readiness, SHAP has been increasingly used in CompPath to assess feature importance in breast cancer and related pipelines \cite{bioengineering10040396,peta2024explainable,wang2024shap}. More recently, PATH-X framework \cite{hussein2025vision} integrated SHAP with ViT and autoencoders to explain patient-level risk stratification. While SHAP is relatively mature for feature- and region-based workflows, its application to end-to-end image attribution remains experimental. Recommended validation practices include i) defining features at biologically interpretable scales (cells, glands, tissue regions), ii) reporting background dataset and sampling strategy, iii) comparing global SHAP rankings with known pathology markers and ablation or occlusion analyses, and iv) visually inspecting local SHAP overlays for spatial and biological plausibility.

    \subsection{Concept-based methods} \label{supp:methods_concept}
    
\begin{table}[!h]
\centering
\caption{Summary of concept-based XAI methods in CompPath: strengths, limitations, and recommended practices for validation and reporting.}
\label{tab:xai_concept_summary}
\begin{tabular}{@{}L{0.145\textwidth}L{0.275\textwidth}L{0.275\textwidth}L{0.255\textwidth}@{}}
\toprule
\textbf{Method} & \textbf{Strengths} & \textbf{Limitations} & \textbf{Cautions \& best practices} \\
\midrule

\textbf{TCAV} &
Quantifies the influence of human-defined concepts on model predictions; enables hypothesis-driven auditing. &
Assumes linear concept separability in latent space; sensitive to quality and diversity of concept exemplars and to choice of network layer. &
Evaluate multiple concept sets across layers; use held-out exemplars from diverse sites and scanners; include negative-control concepts to detect spurious correlations. \\ \hline

\textbf{Concept-bottleneck models (CBM)} &
Intrinsically interpretable; predicts pathologist-defined concepts as intermediate outputs; supports quantitative auditing of concept-level reasoning. &
Requires dense, concept-specific expert annotations; fixed concept sets may miss composite morphologies; bottleneck can constrain learning. &
Validate concept predictions against expert annotation and inter-reader variability; assess cross-site and cross-scanner concept consistency \cite{jahanifar2025domain}; monitor concept leakage via ablation. \\ \hline

\textbf{Image captioning} &
Generates natural-language summaries of regions or slides; supports report-style outputs and multi-modal reasoning. &
Prone to hallucination; data-hungry; requires curated paired image-text corpora; quality depends heavily on the underlying language model. &
Develop standardised terminology with pathologists; ground captions to ROI-level evidence; verify factual correctness against ground-truth reports. \\ \hline

\textbf{Vision-language models (VLM)} &
Bidirectional alignment of image regions and pathology terminology; zero-shot classification and semantic retrieval without explicit labels. &
Susceptible to semantic drift; needs large, high-quality paired datasets; visual--semantic grounding requires additional verification. &
Evaluate both visual grounding and semantic grounding; conduct expert review of retrieved examples; test across institutions and scanners. \\

\bottomrule
\end{tabular}
\end{table}

    Concept-based explainability methods are typically post hoc, model-agnostic, and primarily local in scope. It focuses on identifying meaningful, human-interpretable representations within deep learning models. Rather than analyzing individual features or perturbing inputs in backpropagation-based or perturbation-based methods, these concept-based methods aim to capture high-level, abstract concepts that align with human understanding. This makes them particularly valuable in CompPath, where clinically relevant patterns such as tumor morphology, tissue characteristics, or histological features can provide more intuitive insights into model decision-making.
    
    \subsubsection{Testing with Concept Activation Vectors (TCAV)\cite{kim2018interpretability}} is a concept-based model-access dependent method that quantifies how strongly human-interpretable concepts influence model predictions. Rather than attributing importance to individual pixels or low-level features, TCAV evaluates the directional sensitivity of a model's internal representations to predefined, semantically meaningful concepts. These concepts are encoded as concept activation vectors (CAVs), linear directions in latent feature space that capture high-level properties such as ``necrosis'' or ``glandular pattern''. For each concept, a CAV is learned by training a simple linear classifier (e.g., SVM) to separate activations of concept examples from randomly sampled counterexamples at a selected intermediate layer. The directional derivative of the model output along the CAV measures how increasing the presence of the concept would affect the model's confidence for a target class. This formulation enables hypothesis-driven interpretation, allowing targeted questions such as ``Does the model rely on lymphocyte infiltration to predict tumor grade?''.
    
    Strengths of TCAV include its i) explicit linkage between human-understandable concepts and deep model representations, ii) support for hypothesis-driven model auditing and causal probing, and iii) flexibility in incorporating both expert-curated and automatically discovered concepts. However, several limitations and pitfalls must be considered as TCAV assumes linear separability of concepts in latent space, and its reliability is highly sensitive to the quality, quantity, and diversity of concept exemplars. Furthermore, the choice of network layer affects explainability, and spurious (as well as overlapping) concepts can distort attribution scores. From a computational and practical perspective, TCAV incurs a moderate computational cost, requiring one linear (SVM) classifier per concept and layer. The primary overhead lies in curating representative concept examples or in clustering patches and extracting corresponding activations.
    
    In terms of adoption and readiness, TCAV is emerging in CompPath research \cite{gamble2021determining}. Pathology-specific adaptations have been proposed, such as the work by Zitvnansky \cite{vzitvnanskyconcept}, which combines automatic concept discovery with TCAV-style to generate tissue-level concepts in prostate biopsy images \cite{vzitvnanskyconcept}. While these studies demonstrate promise, TCAV remains largely research-focused. Recommended validation practices include evaluating multiple concept sets across different network layers to assess robustness, using held-out concept examples from diverse sites or scanners, and incorporating negative-control concepts to detect spurious correlations or overfitting. 

    \subsubsection{Concept-Bottleneck Models (CBMs)} \cite{koh2020concept} provide intrinsic interpretability by embedding human-defined or pathologist-recognizable concepts directly into a model's architecture as intermediate representations. Rather than learning opaque latent features, CBMs constrain part of the network to predict interpretable concepts (e.g., cribriform glands, fused glands, poorly formed glands), which are subsequently combined to yield the final clinical prediction (e.g., Gleason grade group). This structure design closely mirrors expert diagnostic reasoning and enables direct inspection of how each histologic concept quantitatively contributes to the model's decision.
    
    Strengths of CBMs include their i) built-in transparency through explicitly modeling pathologist-defined tissue features, ii) support for quantitative validation of concept-level predictions, and iii) enhanced auditability for regulatory and clinical review. By exposing intermediate concept predictions, CBMs also facilitate uncertainty estimation when implemented with stochastic variants. However, limitations and pitfalls remain. CBMs require detailed, concept-specific expert annotations, which limit scalability across large, multi-institutional datasets. Fixed concept definitions may fail to capture subtle or composite morphologies, and the imposed bottlenecks can constrain learning if diagnostically relevant but unannotated cues are excluded. From a computational and practical standpoint, CBMs have computational requirements comparable to segmentation-style models, with training costs that scale with concept dimensionality and annotation density. Inference, however, is typically efficient, enabling practical integration into digital pathology pipelines.
    
    In terms of adoption and readiness, CBMs have been demonstrated in pathology-specific applications, such as Gleason grading for prostate cancer \cite{Mittmann2025}, representing one of the first intrinsic XAI designs tailored to CompPath. Their concept-level accountability makes them promising, but they are still limited by challenges in annotation and generalization. Recommended validation practices include i) evaluating concept predictions against expert annotations and inter-observer variability, ii) assessing cross-site and cross-scanner consistency of learned concepts, iii) monitoring potential concept leakage using ablation or counterfactual analyses, and iv) reporting per-concept uncertainty along with its impact on diagnostic downstream performance.

    \subsubsection{Image Captioning (IC)\cite{vinyals2015show}} is a concept-based explainability method that links visual patterns in WSI to human-readable textual descriptions, enabling AI models to communicate their reasoning in natural language. In CompPath, IC models often implemented using vision-language architectures are trained to generate descriptive text for image tiles or regions, which can then be aggregated into slide-level summaries or structured, report-like outputs.
    
    Strengths of image captioning include its ability to translate visual evidence into natural language, facilitate multi-modal reasoning by aligning image features with medical terminology, and support automated generation of descriptive pathology reports. In terms of limitations, IC requires large, well-curated paired image-text datasets, is prone to generating inaccurate or ``hallucinated'' descriptions, which may not be directly supported by the visual input, and hence its objective evaluation of factual accuracy remains challenging. Output quality is also highly dependent on the underlying language model, training data distribution, and terminology alignment. From a computational and practical perspective, IC models are computationally expensive to train and require substantial effort to curate pathology-specific vocabularies, although inference costs are typically moderate once trained.
    
    In terms of adoption and readiness, IC remains at an early research stage in CompPath \cite{lu2024visual}, but promising for generating interpretable clinical-style outputs. Recommended validation practices include i) developing standardized terminology in collaboration with expert pathologists, ii) verifying factual correctness of generated captions against ground-truth reports, and iii) testing model generalization across institutions and scanner variations.

    \subsubsection{Vision-Language Interpretability \cite{kazmierczak2025explainability}} extends concept-based methods beyond caption generation by aligning VLMs in a bidirectional and interactive manner. Rather than solely producing text from images, VLMs such as CLIP, BioViL, and PLIP learn shared embedding spaces that align visual regions with pathology-relevant language. This alignment enables zero-shot classification, matching images to textual diagnostic terms without explicit training labels, as well as semantic retrieval, such as identifying image regions corresponding to concepts like ``necrosis'', tumor border'', or querying which textual concepts best describe a given region. 
    
    Strengths of VLI include that it i) establishes a direct link between visual and text representations, ii) supports zero-shot concept matching and semantic search without explicit annotations, iii) allows clinically intuitive queries such as ``what visual evidence supports this term?'' or ``which pathology term describes this region?'', and iv) enhances transparency in multimodal computational workflows. However, VLI requires large-scale, high-quality paired datasets of WSIs and text reports for robustness, and interpretability depends on the precision of terminology and the quality of cross-modal alignment. Models may suffer from semantic drift, in which visual features are weakly or incorrectly associated with textual terms, leading to semantically plausible but visually ungrounded outputs. From a computational and practical standpoint, training VLMs is computationally intensive due to large paired datasets and high-dimensional embeddings, although inference is efficient and scalable for retrieval and similarity-based analyses. Notably, rigorous evaluation of grounding often requires additional tools for visual-text correspondence. 
    
    Regarding adoption and readiness, VLI has seen early adoption in CompPath, with models such as CLIP, BioViL, and PLIP used to align pathology images with diagnostic reports and structured concepts \cite{zuo2024plip,bannur2023learning}. Nevertheless, VLI is promising for multimodal learning and transparent clinical reporting, but remains in the early research phases. Recommended validation practices include evaluating both visual grounding (whether retrieved image regions truly match the described concept) and semantic grounding (whether generated or matched terms carry the correct pathological meaning), conducting expert review of retrieved examples, testing across institutions and scanners, and comparing with oncology-based term mappings to ensure factual and semantic consistency.
    
    \subsection{Example-based methods} \label{supp:methods_example}
    {\footnotesize
\setlength{\tabcolsep}{3pt}
\renewcommand{\arraystretch}{1.08}

\begin{table}[!htbp]
\centering
\caption{Summary of example-based XAI methods in CompPath: strengths, limitations, and recommended practices for validation and reporting.}
\label{tab:xai_example_summary}
\begin{tabular}{@{}L{0.145\textwidth}L{0.275\textwidth}L{0.275\textwidth}L{0.255\textwidth}@{}}
\toprule
\textbf{Method} & \textbf{Strengths} & \textbf{Limitations} & \textbf{Cautions \& best practices} \\
\midrule

\textbf{Prototypes} &
Intuitive case-based reasoning that aligns with routine clinical practice; surfaces typical supporting and opposing examples; identifies out-of-distribution cases. &
Representativeness depends on training-data diversity and embedding quality; clustering choices can over-represent frequent artefacts. &
Curate multi-institutional prototype banks; assess pathologist agreement on prototype relevance; analyse class boundaries via counter-prototypes. \\ \hline

\textbf{Counterfactuals} &
Test decision boundaries through targeted, interpretable edits; support causal-style reasoning and failure-mode discovery. &
Generating realistic counterfactual histology is hard; synthetic images may introduce artefacts or violate biological constraints; interpretation requires expert review. &
Conduct expert plausibility review; confirm that small, realistic edits produce consistent class changes; enforce morphological and staining constraints. \\ \hline

\textbf{Trust scores} &
Lightweight, data-driven proxy for prediction reliability; flags out-of-distribution predictions; complements visual attribution as a triage signal. &
Sensitive to embedding quality and domain shift; not an explanation of model reasoning; may assign high confidence to artefact-driven predictions. &
Calibrate thresholds separately by site and scanner \cite{jahanifar2025domain}; correlate with reader disagreement and known errors; jointly inspect low-trust cases with attribution maps. \\

\bottomrule
\end{tabular}
\end{table}
}
    Example-based methods are typically post hoc in stage, agnostic in type, and primarily local in scope. Example-based methods in explainable AI aim to enhance model explainability by leveraging real data examples to support predictions. These methods provide tangible references that help users understand the rationale behind a model's decisions. Evans et al.\cite{evans2022explainability} comprehensively reviewed these approaches in the context of CompPath. Below, we describe example-based techniques used in WSI analysis, highlighting their role in improving explainability and trust in deep learning models for clinical decision support.
        
    \subsubsection{Prototypes \cite{li2018deep}} are example-based explainability methods that provide insights into model behavior by identifying and visualizing instances representative of a particular class, feature, or prediction outcome. A prototype may be selected using clustering techniques (e.g., k-means), where cluster centroids or nearest examples serve as representative cases, or learned directly during model training via a prototype-based network that optimizes similarity between inputs and class-specific prototypes.
    
    Strengths of prototype-based explanations lie in their intuitive, case-based reasoning, which closely aligns with routine clinical practice by presenting concrete visual examples. Prototypes help identify class boundaries and detect out-of-distribution or anomalous patches, enhancing transparency in audit studies. However, several limitations and pitfalls should be noted, as the representativeness of prototypes depends strongly on the diversity and quality of the training data, as well as on the clustering choices and the embedding space. Poor embedding or a biased dataset may result in prototypes that over-represent common artifacts rather than biologically meaningful features. From a computational and practical standpoint, prototype extraction incurs a moderate cost, typically involving clustering or nearest-neighbor searches in the embedding space.
    
    In terms of adoption and readiness, prototype-based explanations have been reviewed and demonstrated in CompPath \cite{evans2022explainability}, including applications that identify representative Ki-67-positive and Ki-67-negative annotations within a region. Recommended validation practices include i) curating prototype banks across institutions, ii) assessing pathologist agreement on prototype relevance, and iii) analyzing class boundaries via counter-prototypes that illustrate contrasting cases.

    \subsubsection{Counterfactuals \cite{stepin2021survey}} provide hypothetical scenarios illustrating how a slight, targeted change to an input could lead to different model predictions. 
    
    Strengths of counterfactual explanations include their ability to support causal reasoning by testing how specific, interpretable changes affect predictions, making them useful for auditing model behavior and identifying key diagnostic cues. However, counterfactuals also come with their limitations as i) generating realistic counterfactual tissue images is challenging, ii) synthetic counterfactuals may introduce artifacts or unrealistic textures, iii) edits can violate biological constraints (such as nuclear shape or tissue organization) if not properly constrained, and iv) interpretation requires expert review. From a practical perspective, counterfactual generation can be computationally intensive when implemented using generative models (such as diffusion models or GANs), although latent space counterfactuals can be comparatively lightweight if suitable representations are available. In addition, counterfactuals require well-defined priors and morphological constraints to ensure biological realism.
    
    Regarding adoption and readiness, counterfactual explanations remain an emerging research area in CompPath, with recent work exploring the generation of plausible counterfactual patches for WSI classification tasks \cite{zigutyte}. Their use is currently research-focused and not yet suitable for clinical deployment. Recommended validation strategies include i) conducting expert ``Turing tests'' to assess visual plausibility, ii) quantitatively verifying faithfulness by confirming that small, realistic edits produce consistent induced class changes, and iii) enforcing morphological and staining constraints to maintain histologic credibility.
    
    \subsubsection{Trust Scores \cite{markus2021role}} provide a quantitative measure of how reliable a model's prediction is by comparing it to neighboring samples in the learned feature space. Trust scores are not explanations in the strict sense. They provide reliable evidence by estimating whether a test case lies near training examples with consistent labels. In XAI workflows, they are best used as uncertainty or triage signals alongside attribution, concept, or prototype-based explanations.
    
    Strengths of trust scores include their i) intuitive, data-driven proxy for prediction confidence by comparing each prediction to its nearest neighbors in feature space, ii) ability to identify unreliable predictions, outliers, or out-of-distribution predictions, and to complement visual attribution methods by quantifying uncertainty. However, trust scores depend heavily on the quality and calibration of the embedding space, domain shifts, scanner variability, or poorly learned representations can produce misleading confidence estimates. Moreover, trust scores do not explain underlying reasoning and may still assign high confidence to predictions driven/dominated by artifacts. From a computational and practical standpoint, trust scores are relatively lightweight, relying on k-nearest-neighbor searches in the embedding space that can be efficiently scaled using approximate nearest neighbor indexing for large datasets.
    
    In terms of adoption and readiness, trust scores have been applied in CompPath research as a quality assurance and triage tool for reviewing model outputs \cite{evans2022explainability}. Recommended validation practices include i) calibrating trust thresholds separately for each site or scanner domain, ii) correlating trust scores with pathologist disagreement rates and known error cases, iii) jointly inspecting low-trust predictions alongside attribution maps (e.g., Grad-CAM or LRP) to identify and investigate potential model failures.

\section{Extended Deployment Guidance}
\label{supp:other}
\subsection{Other Considerations}

        \subsubsection{Practical and Regulatory Readiness}
        
            XAI methods vary widely in computational cost, making deployment context essential to method selection. For computationally constrained environments (e.g., limited GPU memory, real-time inference requirements, or edge devices), lightweight visualization methods such as CAM, Grad-CAM, or attention export provide quick qualitative verification with minimal computational overhead (typically less than 10\% increase in inference time). These methods are suitable for exploratory analysis and preliminary model validation but may lack the quantitative rigor required for high-stakes applications.

            For prospective validation studies or regulatory submissions, a multi-method approach is recommended. Combine intrinsic interpretability (e.g., attention mechanisms, prototype-based models) with post hoc attribution methods (Integrated Gradients, LRP, DeepLIFT) and gradient-free validation (Occlusion Sensitivity, RISE) to generate measurable trustworthiness metrics. This layered validation provides both qualitative evidence (visualizations that pathologists can inspect) and quantitative evidence (faithfulness metrics that regulators can audit). The computational cost is substantial; high-fidelity methods such as LRP or RISE may increase inference time by 10-100×, but this investment is necessary to produce evidence suitable for regulatory documentation, internal validation, or risk management.

            To enable independent validation and regulatory review, XAI studies must document implementation details comprehensively. Report i) model architecture specifics, including which layer was analyzed for gradient-based methods, baseline definitions for Integrated Gradients and DeepLIFT, and propagation rules for LRP; ii) quantitative evaluation metrics such as faithfulness scores (insertion/deletion curves, ROAR), stability metrics (rank correlation across repeated runs for stochastic methods like LIME), and cross-site robustness (correlation of attribution maps across institutions and scanners); and iii) validation against expert knowledge, including overlap with pathologist-annotated regions of interest, inter-observer agreement metrics, and definitions of features or concepts used. Failure to document these details is a leading cause of irreproducibility in XAI literature.

            A practical development strategy balances validation rigor with deployment efficiency. During model development and validation, apply computationally expensive methods (perturbation-based validation, faithfulness metrics, concept grounding) to establish that the model reasons correctly and to generate evidence for regulatory submissions. Archive this validation evidence as supporting documentation. For production deployment, use only fast methods (Grad-CAM, attention overlays) that provide real-time feedback to clinicians without degrading workflow efficiency. This two-stage approach ensures both rigorous validation and practical usability.

        \subsubsection{Deployment Guidance by Stage of Adoption}

            Explainability requirements evolve as AI models progress from research prototypes to regulatory approval to clinical integration. At each stage, different XAI methods and validation standards are appropriate.

            During early-stage validation and clinical pilot studies, the primary goal is to verify that the model attends to histologically plausible regions and to identify systematic failure modes. Use fast visualization methods, such as Grad-CAM or attention overlays, to confirm that the model's focus aligns with expected histologic regions (e.g., tumor nuclei for grade prediction, immune infiltrates for immunotherapy response). Document cases where the model fails (e.g., misclassifications, ambiguous predictions, predictions driven by artifacts), and characterize failure patterns (e.g., model misclassified cases with extensive necrosis). This exploratory analysis informs model refinement and helps set realistic expectations for clinical performance. Quantitative faithfulness metrics are not required at this stage.

            For regulatory submissions or conformity assessment, including De Novo, 510(k), PMA, CE marking, or equivalent pathways, XAI evidence may support documentation of intended use, model limitations, human factors, failure modes, and risk controls. However, explanation outputs should be presented as supporting evidence, not as substitutes for analytical validation, clinical validation, usability assessment, calibration, robustness testing, and post-market monitoring. Combine spatial localization methods (Grad-CAM or attention) with quantitative attribution methods (LRP or Integrated Gradients with faithfulness metrics such as insertion/deletion curves) and perturbation-based causal validation (Occlusion Sensitivity or RISE). Document cross-site robustness by applying XAI methods to slides from multiple institutions, scanners, and staining batches, and report whether explanations remain consistent across these variations. Characterize failure modes systematically, identify which types of cases produce ambiguous or misleading explanations and explain why. 

            For clinical deployment and routine diagnostic use, prioritize methods that communicate reasoning in pathologist-aligned language. Concept-based methods (TCAV, Concept Bottleneck Models, Vision-Language Models) and example-based methods (Prototypes, Counterfactuals) are more useful in clinical workflows than pixel-level heatmaps. Pathologists need actionable insights for, e.g., ``this case is high-grade because of cribriform architecture and necrosis'', rather than abstract gradient visualizations. Combine explanations with uncertainty estimates (trust scores, prediction confidence) to flag ambiguous cases that warrant human review or second opinions. Integrate XAI outputs directly into digital pathology viewers to minimize workflow disruption; standalone explanation interfaces that require context-switching reduce adoption.
            
% [Paste here: Practical and Regulatory Readiness subsubsection,
% Deployment Guidance by Stage of Adoption subsubsection]
Several XAI directions active outside CompPath have yet to see meaningful uptake, and warrant attention as the field matures. Mechanistic interpretability techniques originating in language-model research \cite{templeton2024scaling,cunningham2023sparse}, notably sparse autoencoders that decompose activations into monosemantic features, could in principle recover pathologist-recognisable structures from FM activations without upfront concept annotations. Causal graph neural networks and causal representation learning \cite{scholkopf2021toward,zecevic2021relating} offer a principled route to distinguishing causal from confounded relationships in cell-graph and tissue-graph pipelines. Migrating these methods into gigapixel-scale CompPath, and validating them against pathologist-defined concepts, is a productive frontier for the field.

\section{Literature}
\label{supp:literature}
\newpage
{\footnotesize
\setlength{\tabcolsep}{3pt}
\renewcommand{\arraystretch}{1.08}

\begin{longtable}{c|c} 
    
    \caption{Studies reviewed in this article with corresponding XAI method in specific anatomies.}  \footnotesize
    \label{supp:tab:literature}  
    \\
    \hline
    \textbf{Anatomy} & \textbf{Method} \\ \hline
    \endhead 
    \hline
    \multicolumn{2}{r}{Continued on next page} \\
    \endfoot
    \hline
    \endlastfoot 
    
    Bladder,Brain       & CAM \cite{woerl2020deep} \\
        &  Grad-CAM \cite{tang2019interpretable,kubach2020same}           \\
         &  Attention \cite{baheti2025multimodal,innani2025artificial,innani2025ai,innani2025interpretable}   
         \\ \hline  
    Head \& Neck & Grad-CAM \cite{dorrich2023explainable} \\ \hline 
    
    Breast     &  Activation Layer Visualization \cite{gecer2018detection}   \\
        & Grad-CAM \cite{obikane2020weakly,pirovano2020improving}       \\
        &  CAM \cite{yang2019guided}\\
        & Saliency Maps \cite{zeiser2021deepbatch}\\
        & Attention \cite{liang2023interpretable,dooper2023gigapixel,sharma2021} \\
         &  SHAP \cite{bioengineering10040396,peta2024explainable,wang2024shap}\\
         &   LIME \cite{palatnik2019local,peta2024explainable}\\
         & LRP \cite{binder2021morphological,seegerer2020interpretable} \\
         & Feature Attribution \cite{graziani2020concept,graziani2019visualizing}  \\
         & Occlusion \cite{francis,afshar2024ibo} \\
         & Deconvolution \cite{gecer2018detection} \\
         &  TCAV \cite{gamble2021determining} \\ 
         & Integrated Gradient(IG)\cite{francis} \\ \hline 
    Chest   & Grad-CAM \cite{he2017deep,teramoto2019automated,zhang2022dtfd}\\
    &   Saliency Maps \cite{wang2019weakly,coudray2018classification} \\
     &   Attention \cite{zhao2021lung,lu2021data}\\ \hline 
     
    Cervical  &  Grad-CAM \cite{gupta2020region,kiran2019automatic}\\
         &   Attention \cite{fremond2023interpretable}               \\ \hline 
                    
    Gastrointestinal& CAM \cite{heinemann2019deep,kiani2020impact,yan2020prior,kather2019predicting} \\
        & Grad-CAM \cite{korbar2017looking,kowsari2020hmic,niehues2023generalizable,li2021multi,kanavati2021deep} \\
        & Saliency Maps \cite{laleh2021deep,kuntz2021gastrointestinal,neto2022imil4path}\\            &Attention \cite{zhu2020weakly,calderaro2023deep,zeng2023artificial,jiang2024end,zhang2019pathologist,li2021multi,chakraborty2022predicting}     \\
        & LIME \cite{hossain2022early} \\
        & DeepLIFT \cite{shrikumar2017learning,hossain2022early} \\
        & Counterfactual \cite{zigutyte} \\ \hline 
            
    Lymph nodes & Gradient \cite{ji2019gradient,rahnfeld2024comparative} \\
    &   Saliency Maps \cite{wang2021predicting,bejnordi2017diagnostic}    \\
    &Attention \cite{wu2023artificial,yu2023prototypical,lu2021data,rahnfeld2024comparative} \\
    &SHAP \cite{rahnfeld2024comparative} \\
    &RISE \cite{rahnfeld2024comparative} \\
    & Integrated Gradient \cite{rahnfeld2024comparative} \\
    Lung & Grad-CAM \cite{civit2022non} \\
    & Grad-CAM++\cite{han2022multi} \\ \hline 
     Ovarian & IG \cite{10643508} \\
      & Saliency mapS \cite{10643508} \\
      & Grad-CAM \cite{10643508} \\
      & DeepLift \cite{10643508} \\\hline
    Prostate & Grad-CAM \cite{pinckaers2021detection,ferrero2024histoem} \\
    & Saliency Maps \cite{silva2020going,singhal2022deep} \\
    & Attention \cite{tavolara2023one,xiang2023automatic,grisi2023hierarchical} \\
    & Occlusion based \cite{gallo2023shedding} \\
    & LIME, SHAP \cite{bhattacharjee2022explainable} \\
    & Concept bottleneck models \cite{Mittmann2025} \\
    & Activation Layer Visualization \cite{bhattacharjee2022explainable} \\ \hline  
    Skin &  Saliency Maps \cite{thomas2021interpretable,xie2019interpretable}  \\
    &  Attention \cite{del2021attention,lu2021data,geijs2023detection}  \\ \hline 
    
       Multiple   &     Feature Visualization \cite{xu2017large}\\
            &   Grad-CAM \cite{chan2019histosegnet,chan2019histosegnet}\\
            &   CAM \cite{huang2019evidence}\\   
            &   LRP \cite{hagele2020resolving,hense2024xmil,keyl2023decoding}\\
            &   Attention \cite{saldanha2023self,lee2022derivation,wang2023surformer,shao2021transmil,wang2022transformer,kapse2024si}\\
           & Feature Attribution \cite{diao2021human} \\ \hline 
    \end{longtable}
   }

\renewcommand{\refname}{Supplementary References}
\putbib[references]

\end{bibunit}

\begin{thebibliography}{10}

\bibitem{gurcan2009histopathological}
Gurcan MN, Boucheron LE, Can A, Madabhushi A, Rajpoot NM, Yener B.
\newblock Histopathological image analysis: A review.
\newblock IEEE reviews in biomedical engineering. 2009;2:147-71.

\bibitem{compayl}
Innani S, Nasrallah MP, Bell WR, Baheti B, Bakas S.
\newblock Multi-scale Whole Slide Image Assessment Improves Deep Learning based WHO 2021 Glioma Classification.
\newblock In: Proceedings of the MICCAI Workshop on Computational Pathology. vol. 254 of Proceedings of Machine Learning Research. PMLR; 2024. p. 142-53.

\bibitem{collins2025artificial}
Collins K, Innani S, Ebare K, Saad M, Siegmund SE, Williamson SR, et~al.
\newblock Artificial Intelligence--Based Classification of Renal Oncocytic Neoplasms: Advancing From a 2-Class Model of Renal Oncocytoma and Low-Grade Oncocytic Tumor to a 3-Class Model Including Chromophobe Renal Cell Carcinoma.
\newblock Archives of Pathology \& Laboratory Medicine. 2025.

\bibitem{innani2025ai}
Innani S, Bell WR, Nasrallah MP, Baheti B, Bakas S.
\newblock AI-driven WHO 2021 classification of gliomas based only on H\&E-stained slides.
\newblock Neuro-Oncology. 2025:noaf189.

\bibitem{innani2025artificial}
Innani S, Bell WR, Nasrallah M, Baheti B, Bakas S.
\newblock Artificial intelligence predicts 2021 WHO glioma subtypes from whole slide images.
\newblock Cancer Research. 2025;85(8\_Supplement\_1):6247-7.

\bibitem{baheti_eano}
Baheti B, Innani S, Mehdiratta G, Nasrallah MP, Bakas S.
\newblock {P13.13.B INTERPRETABLE WHOLE SLIDE IMAGE PROGNOSTIC STRATIFICATION OF GLIOBLASTOMA PATIENTS FURTHERING DISEASE UNDERSTANDING}.
\newblock Neuro-Oncology. 2023 09;25(Supplement\_2):ii103-4.

\bibitem{baheti_eano_clustering}
Baheti B, Innani S, Nasrallah MP, Bakas S.
\newblock {OS03.6.A UNSUPERVISED CLUSTERING OF MORPHOLOGY PATTERNS ON WHOLE SLIDE IMAGES GUIDE PROGNOSTIC STRATIFICATION OF GLIOBLASTOMA PATIENTS}.
\newblock Neuro-Oncology. 2023 09;25(Supplement\_2):ii15-5.

\bibitem{baheti2025multimodal}
Baheti B, Rai S, Innani S, Mehdiratta G, Bell WR, Guntuku SC, et~al.
\newblock Multimodal Explainable Artificial Intelligence for Prognostic Stratification of Glioblastoma Patients.
\newblock Modern Pathology. 2025:100797.

\bibitem{baheti2024prognostic}
Baheti B, Innani S, Nasrallah M, Bakas S.
\newblock Prognostic stratification of glioblastoma patients by unsupervised clustering of morphology patterns on whole slide images furthering our disease understanding.
\newblock Frontiers in Neuroscience. 2024;18:1304191.

\bibitem{isbi}
Innani S, Baheti B, Nasrallah MP, Bakas S.
\newblock Weakly Supervised IDH-Status Glioma Classification from H\&E-Stained Whole Slide Images.
\newblock In: 2024 IEEE International Symposium on Biomedical Imaging (ISBI). IEEE; 2024. p. 1-5.

\bibitem{sno_idh}
Innani S, Baheti B, Nasrallah MP, Bakas S.
\newblock {PATH-39. INTERPRETABLE IDH CLASSIFICATION FROM H\&E-STAINED HISTOLOGY SLIDES}.
\newblock Neuro-Oncology. 2023 11;25(Supplement\_5):v177-7.

\bibitem{sno24}
Innani S, Baheti B, Nasrallah MP, Bell WR, Bakas S.
\newblock PATH-39. AI-BASED IDENTIFICATION OF GLIOMA IDH MUTATIONAL STATUS FROM H\&;E-STAINED WHOLE SLIDE IMAGES.
\newblock Neuro-Oncology. 2024 11;26(Supplement\_8):viii187-7.

\bibitem{innani2025interpretable}
Innani S, Bell WR, Harmsen H, Nasrallah MP, Baheti B, Bakas S.
\newblock Interpretable artificial intelligence based determination of glioma IDH mutation status directly from histology slides.
\newblock Neuro-Oncology Advances. 2025;7(1):vdaf140.

\bibitem{innani2025path}
Innani S, Pitarch-Abaigar C, Marwan MM, Harmsen H, Bell WR, Makris D, et~al.
\newblock PATH-69. AI-based prognostic risk stratification of adult-type diffuse gliomas using H\&E-stained slides.
\newblock Neuro-Oncology. 2025;27(Supplement\_5):v257-7.

\bibitem{graham2023screening}
Graham S, Minhas F, Bilal M, Ali M, Tsang YW, Eastwood M, et~al.
\newblock Screening of normal endoscopic large bowel biopsies with interpretable graph learning: a retrospective study.
\newblock Gut. 2023;72(9):1709-21.

\bibitem{innani2025path67}
Innani S, Bell WR, Nasrallah MP, Baheti B, Bakas S.
\newblock PATH-67. AI-based WHO 2021 classification of adult-type diffuse glioma solely from H\&E-stained slides.
\newblock Neuro-Oncology. 2025;27(Supplement\_5):v256-6.

\bibitem{thakur2025img}
Thakur S, Malec S, Pitarc C, Linardos A, Innani S, Adap S, et~al.
\newblock IMG-121. BraTS-Pathology 2024: Insights and Future Directions Informed by the AI-RANO \& RANO-RGP Effort to Assess Glioblastoma Heterogeneity.
\newblock Neuro-Oncology. 2025;27(Supplement\_5):v303-4.

\bibitem{thakur2024tmic}
Thakur S, Faghani S, Moassefi M, Baid U, Chung V, Pati S, et~al.
\newblock TMIC-60. BRATS-PATH: Assessing Heterogeneous Histopathologic Regions in Glioblastoma.
\newblock Neuro-Oncology. 2024;26(Supplement\_8):viii312-2.

\bibitem{matthews2024public}
Matthews GA, McGenity C, Bansal D, Treanor D.
\newblock Public evidence on AI products for digital pathology.
\newblock NPJ Digital Medicine. 2024;7(1):300.

\bibitem{liu2024regulatory}
Liu Y, Yu W, Dillon T.
\newblock Regulatory responses and approval status of artificial intelligence medical devices with a focus on China.
\newblock NPJ Digital Medicine. 2024;7(1):255.

\bibitem{schmidt2024mapping}
Schmidt J, Schutte NM, Buttigieg S, Novillo-Ortiz D, Sutherland E, Anderson M, et~al.
\newblock Mapping the regulatory landscape for artificial intelligence in health within the European Union.
\newblock npj Digital Medicine. 2024;7(1):229.

\bibitem{poalelungi2024revolutionizing}
Poalelungi DG, Neagu AI, Fulga A, Neagu M, Tutunaru D, Nechita A, et~al.
\newblock Revolutionizing pathology with artificial intelligence: Innovations in immunohistochemistry.
\newblock Journal of Personalized Medicine. 2024;14(7):693.

\bibitem{mcgenity2024artificial}
McGenity C, Clarke EL, Jennings C, Matthews G, Cartlidge C, Stocken DD, et~al.
\newblock Artificial intelligence in digital pathology: a systematic review and meta-analysis of diagnostic test accuracy.
\newblock npj Digital Medicine. 2024;7(1):114.

\bibitem{scans2024paige}
Scans CH.
\newblock The Paige Prostate Suite: Assistive Artificial Intelligence for Prostate Cancer Diagnosis: Emerging Health Technologies.
\newblock Ottawa (ON): Canadian Agency for Drugs and Technologies in Health. 2024.

\bibitem{pantanowitz2020artificial}
Pantanowitz L, Quiroga-Garza GM, Bien L, Heled R, Laifenfeld D, Linhart C, et~al.
\newblock An artificial intelligence algorithm for prostate cancer diagnosis in whole slide images of core needle biopsies: a blinded clinical validation and deployment study.
\newblock The Lancet Digital Health. 2020;2(8):e407-16.

\bibitem{van2021deep}
Van~der Laak J, Litjens G, Ciompi F.
\newblock Deep learning in histopathology: the path to the clinic.
\newblock Nature medicine. 2021;27(5):775-84.

\bibitem{Health_2024}
Health I. Hallucinating AI Perfection in Healthcare: Navigating the Challenge of Hallucinations. Medium; 2024.
\newblock Available from: \url{https://inflecthealth.medium.com/hallucinating-ai-perfection-in-healthcare-navigating-the-challenge-of-hallucinations-4e052a4492e5}.

\bibitem{jahanifar2025domain}
Jahanifar M, Raza M, Xu K, Vuong TTL, Jewsbury R, Shephard A, et~al.
\newblock Domain generalization in computational pathology: survey and guidelines.
\newblock ACM Computing Surveys. 2025;57(11):1-37.

\bibitem{evans2022explainability}
Evans T, Retzlaff CO, Gei{\ss}ler C, Kargl M, Plass M, M{\"u}ller H, et~al.
\newblock The explainability paradox: Challenges for xAI in digital pathology.
\newblock Future Generation Computer Systems. 2022;133:281-96.

\bibitem{aldea2026ebai}
Aldea M, Salto-Tellez M, Marra A, Umeton R, Stenzinger A, Koopman M, et~al.
\newblock ESMO basic requirements for AI-based biomarkers in oncology (EBAI).
\newblock Annals of Oncology. 2025.

\bibitem{ghassemi2021false}
Ghassemi M, Oakden-Rayner L, Beam AL.
\newblock The false hope of current approaches to explainable artificial intelligence in health care.
\newblock The lancet digital health. 2021;3(11):e745-50.

\bibitem{dwivedi2023explainable}
Dwivedi R, Dave D, Naik H, Singhal S, Omer R, Patel P, et~al.
\newblock Explainable AI (XAI): Core ideas, techniques, and solutions.
\newblock ACM Computing Surveys. 2023;55(9):1-33.

\bibitem{doshi2017towards}
Doshi-Velez F, Kim B.
\newblock Towards a rigorous science of interpretable machine learning.
\newblock arXiv preprint arXiv:170208608. 2017.

\bibitem{holzinger2017we}
Holzinger A, Biemann C, Pattichis CS, Kell DB.
\newblock What do we need to build explainable AI systems for the medical domain?
\newblock arXiv preprint arXiv:171209923. 2017.

\bibitem{hassija2024interpreting}
Hassija V, Chamola V, Mahapatra A, Singal A, Goel D, Huang K, et~al.
\newblock Interpreting black-box models: a review on explainable artificial intelligence.
\newblock Cognitive Computation. 2024;16(1):45-74.

\bibitem{arrieta2020explainable}
Arrieta AB, D{\'\i}az-Rodr{\'\i}guez N, Del~Ser J, Bennetot A, Tabik S, Barbado A, et~al.
\newblock Explainable Artificial Intelligence (XAI): Concepts, taxonomies, opportunities and challenges toward responsible AI.
\newblock Information fusion. 2020;58:82-115.

\bibitem{lipton2018mythos}
Lipton ZC.
\newblock The mythos of model interpretability: In machine learning, the concept of interpretability is both important and slippery.
\newblock Queue. 2018;16(3):31-57.

\bibitem{adadi2018peeking}
Adadi A, Berrada M.
\newblock Peeking inside the black-box: a survey on explainable artificial intelligence (XAI).
\newblock IEEE access. 2018;6:52138-60.

\bibitem{mueller2019explanation}
Mueller ST, Hoffman RR, Clancey W, Emrey A, Klein G.
\newblock Explanation in human-AI systems: A literature meta-review, synopsis of key ideas and publications, and bibliography for explainable AI.
\newblock arXiv preprint arXiv:190201876. 2019.

\bibitem{murdoch2019definitions}
Murdoch WJ, Singh C, Kumbier K, Abbasi-Asl R, Yu B.
\newblock Definitions, methods, and applications in interpretable machine learning.
\newblock Proceedings of the National Academy of Sciences. 2019;116(44):22071-80.

\bibitem{ibrahim2023explainable}
Ibrahim R, Shafiq MO.
\newblock Explainable convolutional neural networks: a taxonomy, review, and future directions.
\newblock ACM Computing Surveys. 2023;55(10):1-37.

\bibitem{das2020opportunities}
Das A, Rad P.
\newblock Opportunities and challenges in explainable artificial intelligence (xai): A survey.
\newblock arXiv preprint arXiv:200611371. 2020.

\bibitem{bhati2024survey}
Bhati D, Neha F, Amiruzzaman M.
\newblock A survey on explainable artificial intelligence (xai) techniques for visualizing deep learning models in medical imaging.
\newblock Journal of Imaging. 2024;10(10):239.

\bibitem{van2022explainable}
Van~der Velden BH, Kuijf HJ, Gilhuijs KG, Viergever MA.
\newblock Explainable artificial intelligence (XAI) in deep learning-based medical image analysis.
\newblock Medical Image Analysis. 2022;79:102470.

\bibitem{pocevivciute2020survey}
Pocevi{\v{c}}i{\=u}t{\. e} M, Eilertsen G, Lundstr{\"o}m C.
\newblock Survey of XAI in digital pathology.
\newblock Artificial intelligence and machine learning for digital pathology: state-of-the-art and future challenges. 2020:56-88.

\bibitem{dy2026clarifying}
Dy A, Buetow SM, Bredemeyer AJ, Saini ML, Lucas F, Bennett S, et~al.
\newblock Clarifying validation terminologies in healthcare.
\newblock NPJ Digital Medicine. 2026;9(1):318.

\bibitem{hedstrom2022quantus}
Hedström A, Weber L, Krakowczyk D, Bareeva D, Motzkus F, Samek W, et~al.
\newblock Quantus: an explainable {AI} toolkit for responsible evaluation of neural network explanations and beyond.
\newblock Journal of Machine Learning Research. 2023;24(34):1-11.

\bibitem{EU2024AIAct}
{European Parliament and Council}. Regulation (EU) 2024/1689 of the European Parliament and of the Council of 13 June 2024 laying down harmonised rules on artificial intelligence; 2024.
\newblock Official Journal of the European Union.
\newblock Available from: \url{https://eur-lex.europa.eu/eli/reg/2024/1689/oj}.

\bibitem{FDAAIML}
Food U, gov) DA.
\newblock Artificial Intelligence and Machine Learning in Software as a Medical Device. 2021.
\newblock Accessed on 22 Sept 2021.
\newblock Available from: \url{https://www.fda.gov/files/medical%20devices/published/US-FDA-Artificial-Intelligence-and-Machine-Learning-Discussion-Paper.pdf}.

\bibitem{holzinger2019causability}
Holzinger A, Langs G, Denk H, Zatloukal K, M{\"u}ller H.
\newblock Causability and explainability of artificial intelligence in medicine.
\newblock Wiley Interdisciplinary Reviews: Data Mining and Knowledge Discovery. 2019;9(4):e1312.

\bibitem{campbell1986relabeling}
Campbell DT.
\newblock Relabeling internal and external validity for applied social scientists.
\newblock New Directions for Program Evaluation. 1986;1986(31):67-77.

\bibitem{west2010campbell}
West SG, Thoemmes F.
\newblock Campbell's and {Rubin's} perspectives on causal inference.
\newblock In: Causality in the Sciences; 2010. Spencer Foundation conference paper; available at https://jenni.uchicago.edu/Spencer\_Conference/.

\bibitem{adebayo2018sanity}
Adebayo J, Gilmer J, Muelly M, Goodfellow I, Hardt M, Kim B.
\newblock Sanity checks for saliency maps.
\newblock Advances in neural information processing systems. 2018;31.

\bibitem{hooker2019benchmark}
Hooker S, Erhan D, Kindermans PJ, Kim B.
\newblock A benchmark for interpretability methods in deep neural networks.
\newblock Advances in neural information processing systems. 2019;32.

\bibitem{mittelstadt2019principles}
Mittelstadt B.
\newblock Principles alone cannot guarantee ethical AI.
\newblock Nature machine intelligence. 2019;1(11):501-7.

\bibitem{graziani2020concept}
Graziani M, Andrearczyk V, Marchand-Maillet S, M{\"u}ller H.
\newblock Concept attribution: Explaining CNN decisions to physicians.
\newblock Computers in biology and medicine. 2020;123:103865.

\bibitem{rueda2024just}
Rueda J, Rodr{\'\i}guez JD, Jounou IP, Hortal-Carmona J, Aus{\'\i}n T, Rodr{\'\i}guez-Arias D.
\newblock “Just” accuracy? Procedural fairness demands explainability in AI-based medical resource allocations.
\newblock AI \& society. 2024;39(3):1411-22.

\bibitem{kamath2021explainable}
Kamath U, Liu J.
\newblock Explainable artificial intelligence: an introduction to interpretable machine learning. vol.~2.
\newblock Springer; 2021.

\bibitem{jain2019attention}
Jain S, Wallace BC.
\newblock Attention is not explanation.
\newblock arXiv preprint arXiv:190210186. 2019.

\bibitem{wiegreffe2019attention}
Wiegreffe S, Pinter Y.
\newblock Attention is not not explanation.
\newblock arXiv preprint arXiv:190804626. 2019.

\bibitem{lundberg2017unified}
Lundberg SM, Lee SI.
\newblock A unified approach to interpreting model predictions.
\newblock Advances in neural information processing systems. 2017;30.

\bibitem{yosinski2015understanding}
Yosinski J, Clune J, Nguyen A, Fuchs T, Lipson H.
\newblock Understanding neural networks through deep visualization.
\newblock arXiv preprint arXiv:150606579. 2015.

\bibitem{kim2018interpretability}
Kim B, Wattenberg M, Gilmer J, Cai C, Wexler J, Viegas F, et~al.
\newblock Interpretability beyond feature attribution: Quantitative testing with concept activation vectors (tcav).
\newblock In: International conference on machine learning. PMLR; 2018. p. 2668-77.

\bibitem{koh2020concept}
Koh PW, Nguyen T, Tang YS, Mussmann S, Pierson E, Kim B, et~al.
\newblock Concept bottleneck models.
\newblock In: International conference on machine learning. PMLR; 2020. p. 5338-48.

\bibitem{zuo2024plip}
Zuo J, Hong J, Zhang F, Yu C, Zhou H, Gao C, et~al.
\newblock Plip: Language-image pre-training for person representation learning.
\newblock Advances in Neural Information Processing Systems. 2024;37:45666-702.

\bibitem{bannur2023learning}
Bannur S, Hyland S, Liu Q, Perez-Garcia F, Ilse M, Castro DC, et~al.
\newblock Learning to exploit temporal structure for biomedical vision-language processing.
\newblock In: Proceedings of the IEEE/CVF Conference on Computer Vision and Pattern Recognition; 2023. p. 15016-27.

\bibitem{jiang2018trust}
Jiang H, Kim B, Guan M, Gupta M.
\newblock To trust or not to trust a classifier.
\newblock Advances in neural information processing systems. 2018;31.

\bibitem{abdar2021review}
Abdar M, Pourpanah F, Hussain S, Rezazadegan D, Liu L, Ghavamzadeh M, et~al.
\newblock A review of uncertainty quantification in deep learning: Techniques, applications and challenges.
\newblock Information fusion. 2021;76:243-97.

\bibitem{he2026survey}
He W, Jiang Z, Xiao T, Xu Z, Li Y.
\newblock A survey on uncertainty quantification methods for deep learning.
\newblock ACM Computing Surveys. 2026;58(7):1-35.

\bibitem{haendel2026governing}
Haendel MA, Ahern R, Bailey KB, Bakas S, Barth-Jones DC, Bohl A, et~al.
\newblock Governing real-world health data as a public utility.
\newblock Science. 2026;391(6789):993-6.

\bibitem{bakas2024brats}
Bakas S, Thakur SP, Faghani S, Moassefi M, Baid U, Chung V, et~al.
\newblock Brats-path challenge: Assessing heterogeneous histopathologic brain tumor sub-regions.
\newblock arXiv preprint arXiv:240510871. 2024.

\bibitem{moons2015transparent}
Moons KG, Altman DG, Reitsma JB, Ioannidis JP, Macaskill P, Steyerberg EW, et~al.
\newblock Transparent Reporting of a multivariable prediction model for Individual Prognosis or Diagnosis (TRIPOD): explanation and elaboration.
\newblock Annals of internal medicine. 2015;162(1):W1-W73.

\bibitem{collins2015transparent}
Collins GS, Reitsma JB, Altman DG, Moons KG.
\newblock Transparent reporting of a multivariable prediction model for individual prognosis or diagnosis (TRIPOD): the TRIPOD statement.
\newblock Journal of British Surgery. 2015;102(3):148-58.

\bibitem{kocak2023checklist}
Kocak B, Baessler B, Bakas S, Cuocolo R, Fedorov A, Maier-Hein L, et~al.
\newblock CheckList for EvaluAtion of Radiomics research (CLEAR): a step-by-step reporting guideline for authors and reviewers endorsed by ESR and EuSoMII.
\newblock Insights into imaging. 2023;14(1):75.

\end{thebibliography}


\begin{thebibliography}{100}

\bibitem{lipton2018mythos}
Lipton ZC.
\newblock The mythos of model interpretability: In machine learning, the concept of interpretability is both important and slippery.
\newblock Queue. 2018;16(3):31-57.

\bibitem{arrieta2020explainable}
Arrieta AB, D{\'\i}az-Rodr{\'\i}guez N, Del~Ser J, Bennetot A, Tabik S, Barbado A, et~al.
\newblock Explainable Artificial Intelligence (XAI): Concepts, taxonomies, opportunities and challenges toward responsible AI.
\newblock Information fusion. 2020;58:82-115.

\bibitem{li2022interpretable}
Li X, Xiong H, Li X, Wu X, Zhang X, Liu J, et~al.
\newblock Interpretable deep learning: Interpretation, interpretability, trustworthiness, and beyond.
\newblock Knowledge and Information Systems. 2022;64(12):3197-234.

\bibitem{shaban2019novel}
Shaban M, Khurram SA, Fraz MM, Alsubaie N, Masood I, Mushtaq S, et~al.
\newblock A novel digital score for abundance of tumour infiltrating lymphocytes predicts disease free survival in oral squamous cell carcinoma.
\newblock Scientific reports. 2019;9(1):13341.

\bibitem{graham2019hover}
Graham S, Vu QD, Raza SEA, Azam A, Tsang YW, Kwak JT, et~al.
\newblock Hover-net: Simultaneous segmentation and classification of nuclei in multi-tissue histology images.
\newblock Medical image analysis. 2019;58:101563.

\bibitem{shephard2024fully}
Shephard AJ, Bashir RMS, Mahmood H, Jahanifar M, Minhas F, Raza SEA, et~al.
\newblock A fully automated and explainable algorithm for predicting malignant transformation in oral epithelial dysplasia.
\newblock npj Precision Oncology. 2024;8(1):137.

\bibitem{lu2024visual}
Lu MY, Chen B, Williamson DF, Chen RJ, Liang I, Ding T, et~al.
\newblock A visual-language foundation model for computational pathology.
\newblock Nature Medicine. 2024;30(3):863-74.

\bibitem{jain2019attention}
Jain S, Wallace BC.
\newblock Attention is not explanation.
\newblock arXiv preprint arXiv:190210186. 2019.

\bibitem{wiegreffe2019attention}
Wiegreffe S, Pinter Y.
\newblock Attention is not not explanation.
\newblock arXiv preprint arXiv:190804626. 2019.

\bibitem{guidotti2018survey}
Guidotti R, Monreale A, Ruggieri S, Turini F, Giannotti F, Pedreschi D.
\newblock A survey of methods for explaining black box models.
\newblock ACM Computing Surveys. 2018;51(5):1-42.

\bibitem{arun2021assessing}
Arun N, Gaw N, Singh P, Chang K, Aggarwal M, Chen B, et~al.
\newblock Assessing the trustworthiness of saliency maps for localizing abnormalities in medical imaging.
\newblock Radiology: Artificial Intelligence. 2021;3(6):e200267.

\bibitem{jahanifar2025domain}
Jahanifar M, Raza M, Xu K, Vuong TTL, Jewsbury R, Shephard A, et~al.
\newblock Domain generalization in computational pathology: survey and guidelines.
\newblock ACM Computing Surveys. 2025;57(11):1-37.

\bibitem{FDAAIML}
Food U, gov) DA.
\newblock Artificial Intelligence and Machine Learning in Software as a Medical Device. 2021.
\newblock Accessed on 22 Sept 2021.
\newblock Available from: \url{https://www.fda.gov/files/medical%20devices/published/US-FDA-Artificial-Intelligence-and-Machine-Learning-Discussion-Paper.pdf}.

\bibitem{EU2024AIAct}
{European Parliament and Council}. Regulation (EU) 2024/1689 of the European Parliament and of the Council of 13 June 2024 laying down harmonised rules on artificial intelligence; 2024.
\newblock Official Journal of the European Union.
\newblock Available from: \url{https://eur-lex.europa.eu/eli/reg/2024/1689/oj}.

\bibitem{mittelstadt2019principles}
Mittelstadt B.
\newblock Principles alone cannot guarantee ethical AI.
\newblock Nature machine intelligence. 2019;1(11):501-7.

\bibitem{gebru2021datasheets}
Gebru T, Morgenstern J, Vecchione B, Vaughan JW, Wallach H, Iii HD, et~al.
\newblock Datasheets for datasets.
\newblock Communications of the ACM. 2021;64(12):86-92.

\bibitem{mitchell2019model}
Mitchell M, Wu S, Zaldivar A, Barnes P, Vasserman L, Hutchinson B, et~al.
\newblock Model cards for model reporting.
\newblock In: Proceedings of the Conference on Fairness, Accountability, and Transparency; 2019. p. 220-9.

\bibitem{adebayo2018sanity}
Adebayo J, Gilmer J, Muelly M, Goodfellow I, Hardt M, Kim B.
\newblock Sanity checks for saliency maps.
\newblock Advances in neural information processing systems. 2018;31.

\bibitem{hooker2019benchmark}
Hooker S, Erhan D, Kindermans PJ, Kim B.
\newblock A benchmark for interpretability methods in deep neural networks.
\newblock Advances in neural information processing systems. 2019;32.

\bibitem{petsiuk2018rise}
Petsiuk V.
\newblock Rise: Randomized Input Sampling for Explanation of black-box models.
\newblock arXiv preprint arXiv:180607421. 2018.

\bibitem{wachter2017counterfactual}
Wachter S, Mittelstadt B, Russell C.
\newblock Counterfactual explanations without opening the black box: Automated decisions and the GDPR.
\newblock Harv JL \& Tech. 2017;31:841.

\bibitem{baheti2024prognostic}
Baheti B, Innani S, Nasrallah M, Bakas S.
\newblock Prognostic stratification of glioblastoma patients by unsupervised clustering of morphology patterns on whole slide images furthering our disease understanding.
\newblock Frontiers in Neuroscience. 2024;18:1304191.

\bibitem{samek2016evaluating}
Samek W, Binder A, Montavon G, Lapuschkin S, M{\"u}ller KR.
\newblock Evaluating the visualization of what a deep neural network has learned.
\newblock IEEE transactions on neural networks and learning systems. 2016;28(11):2660-73.

\bibitem{holzinger2019causability}
Holzinger A, Langs G, Denk H, Zatloukal K, M{\"u}ller H.
\newblock Causability and explainability of artificial intelligence in medicine.
\newblock Wiley Interdisciplinary Reviews: Data Mining and Knowledge Discovery. 2019;9(4):e1312.

\bibitem{schmitt2021hidden}
Schmitt M, Maron RC, Hekler A, Stenzinger A, Hauschild A, Weichenthal M, et~al.
\newblock Hidden variables in deep learning digital pathology and their potential to cause batch effects: Prediction model study.
\newblock Journal of medical Internet research. 2021;23(2):e23436.

\bibitem{holzinger2020measuring}
Holzinger A, Carrington A, M{\"u}ller H.
\newblock Measuring the quality of explanations: the system causability scale (SCS) comparing human and machine explanations.
\newblock KI-K{\"u}nstliche Intelligenz. 2020;34(2):193-8.

\bibitem{graziani2020concept}
Graziani M, Andrearczyk V, Marchand-Maillet S, M{\"u}ller H.
\newblock Concept attribution: Explaining CNN decisions to physicians.
\newblock Computers in biology and medicine. 2020;123:103865.

\bibitem{ibrahim2023explainable}
Ibrahim R, Shafiq MO.
\newblock Explainable convolutional neural networks: a taxonomy, review, and future directions.
\newblock ACM Computing Surveys. 2023;55(10):1-37.

\bibitem{rueda2024just}
Rueda J, Rodr{\'\i}guez JD, Jounou IP, Hortal-Carmona J, Aus{\'\i}n T, Rodr{\'\i}guez-Arias D.
\newblock “Just” accuracy? Procedural fairness demands explainability in AI-based medical resource allocations.
\newblock AI \& society. 2024;39(3):1411-22.

\bibitem{dy2026clarifying}
Dy A, Buetow SM, Bredemeyer AJ, Saini ML, Lucas F, Bennett S, et~al.
\newblock Clarifying validation terminologies in healthcare.
\newblock NPJ Digital Medicine. 2026;9(1):318.

\bibitem{kamath2021explainable}
Kamath U, Liu J.
\newblock Explainable artificial intelligence: an introduction to interpretable machine learning. vol.~2.
\newblock Springer; 2021.

\bibitem{adadi2018peeking}
Adadi A, Berrada M.
\newblock Peeking inside the black-box: a survey on explainable artificial intelligence (XAI).
\newblock IEEE access. 2018;6:52138-60.

\bibitem{murdoch2019definitions}
Murdoch WJ, Singh C, Kumbier K, Abbasi-Asl R, Yu B.
\newblock Definitions, methods, and applications in interpretable machine learning.
\newblock Proceedings of the National Academy of Sciences. 2019;116(44):22071-80.

\bibitem{shephard2024automated}
Shephard AJ, Jahanifar M, Wang R, Dawood M, Graham S, Sidlauskas K, et~al.
\newblock An automated pipeline for tumour-infiltrating lymphocyte scoring in breast cancer.
\newblock In: 2024 IEEE International Symposium on Biomedical Imaging (ISBI). IEEE; 2024. p. 1-5.

\bibitem{aubreville2023mitosis}
Aubreville M, Stathonikos N, Bertram CA, Klopfleisch R, Ter~Hoeve N, Ciompi F, et~al.
\newblock Mitosis domain generalization in histopathology images—the MIDOG challenge.
\newblock Medical Image Analysis. 2023;84:102699.

\bibitem{alvarez2018towards}
Alvarez~Melis D, Jaakkola T.
\newblock Towards robust interpretability with self-explaining neural networks.
\newblock Advances in neural information processing systems. 2018;31.

\bibitem{koh2020concept}
Koh PW, Nguyen T, Tang YS, Mussmann S, Pierson E, Kim B, et~al.
\newblock Concept bottleneck models.
\newblock In: International conference on machine learning. PMLR; 2020. p. 5338-48.

\bibitem{kierner2023taxonomy}
Kierner S, Kucharski J, Kierner Z.
\newblock Taxonomy of hybrid architectures involving rule-based reasoning and machine learning in clinical decision systems: A scoping review.
\newblock Journal of biomedical informatics. 2023;144:104428.

\bibitem{ilse2018attention}
Ilse M, Tomczak J, Welling M.
\newblock Attention-based deep multiple instance learning.
\newblock In: International conference on machine learning. PMLR; 2018. p. 2127-36.

\bibitem{lundberg2017unified}
Lundberg SM, Lee SI.
\newblock A unified approach to interpreting model predictions.
\newblock Advances in neural information processing systems. 2017;30.

\bibitem{Selvaraju_2017_ICCV}
Selvaraju RR, Cogswell M, Das A, Vedantam R, Parikh D, Batra D.
\newblock Grad-CAM: Visual Explanations From Deep Networks via Gradient-Based Localization.
\newblock In: Proceedings of the IEEE International Conference on Computer Vision (ICCV); 2017. .

\bibitem{poche2023natural}
Poch{\'e} A, Hervier L, Bakkay MC.
\newblock Natural example-based explainability: a survey.
\newblock In: World Conference on eXplainable Artificial Intelligence. Springer; 2023. p. 24-47.

\bibitem{verma2020counterfactual}
Verma S, Dickerson J, Hines K.
\newblock Counterfactual explanations for machine learning: A review.
\newblock arXiv preprint arXiv:201010596. 2020;2(1):1.

\bibitem{ribeiro2016should}
Ribeiro MT, Singh S, Guestrin C.
\newblock " Why should i trust you?" Explaining the predictions of any classifier.
\newblock In: Proceedings of the 22nd ACM SIGKDD international conference on knowledge discovery and data mining; 2016. p. 1135-44.

\bibitem{fong2017interpretable}
Fong RC, Vedaldi A.
\newblock Interpretable explanations of black boxes by meaningful perturbation.
\newblock In: Proceedings of the IEEE international conference on computer vision; 2017. p. 3429-37.

\bibitem{olden2004accurate}
Olden JD, Joy MK, Death RG.
\newblock An accurate comparison of methods for quantifying variable importance in artificial neural networks using simulated data.
\newblock Ecological modelling. 2004;178(3-4):389-97.

\bibitem{kim2018interpretability}
Kim B, Wattenberg M, Gilmer J, Cai C, Wexler J, Viegas F, et~al.
\newblock Interpretability beyond feature attribution: Quantitative testing with concept activation vectors (tcav).
\newblock In: International conference on machine learning. PMLR; 2018. p. 2668-77.

\bibitem{vaswani2017attention}
Vaswani A, Shazeer N, Parmar N, Uszkoreit J, Jones L, Gomez AN, et~al.
\newblock Attention is all you need.
\newblock Advances in neural information processing systems. 2017;30.

\bibitem{chefer2021transformer}
Chefer H, Gur S, Wolf L.
\newblock Transformer interpretability beyond attention visualization.
\newblock In: Proceedings of the IEEE/CVF conference on computer vision and pattern recognition; 2021. p. 782-91.

\bibitem{zhou2016learning}
Zhou B, Khosla A, Lapedriza A, Oliva A, Torralba A.
\newblock Learning deep features for discriminative localization.
\newblock In: Proceedings of the IEEE conference on computer vision and pattern recognition; 2016. p. 2921-9.

\bibitem{mundhenk2019efficient}
Mundhenk TN, Chen BY, Friedland G.
\newblock Efficient saliency maps for explainable AI.
\newblock arXiv preprint arXiv:191111293. 2019.

\bibitem{montavon2019layer}
Montavon G, Binder A, Lapuschkin S, Samek W, M{\"u}ller KR.
\newblock Layer-wise relevance propagation: an overview.
\newblock Explainable AI: interpreting, explaining and visualizing deep learning. 2019:193-209.

\bibitem{shrikumar2017learning}
Shrikumar A, Greenside P, Kundaje A.
\newblock Learning important features through propagating activation differences.
\newblock In: International conference on machine learning. PMlR; 2017. p. 3145-53.

\bibitem{zeiler2014visualizing}
Zeiler M.
\newblock Visualizing and Understanding Convolutional Networks.
\newblock In: European conference on computer vision/arXiv. vol. 1311; 2014. .

\bibitem{qi2019visualizing}
Qi Z, Khorram S, Li F.
\newblock Visualizing Deep Networks by Optimizing with Integrated Gradients.
\newblock In: CVPR workshops. vol.~2; 2019. p. 1-4.

\bibitem{yosinski2015understanding}
Yosinski J, Clune J, Nguyen A, Fuchs T, Lipson H.
\newblock Understanding neural networks through deep visualization.
\newblock arXiv preprint arXiv:150606579. 2015.

\bibitem{vinyals2015show}
Vinyals O, Toshev A, Bengio S, Erhan D.
\newblock Show and tell: A neural image caption generator.
\newblock In: Proceedings of the IEEE conference on computer vision and pattern recognition; 2015. p. 3156-64.

\bibitem{zuo2024plip}
Zuo J, Hong J, Zhang F, Yu C, Zhou H, Gao C, et~al.
\newblock Plip: Language-image pre-training for person representation learning.
\newblock Advances in Neural Information Processing Systems. 2024;37:45666-702.

\bibitem{chen2019looks}
Chen C, Li O, Tao D, Barnett A, Rudin C, Su JK.
\newblock This looks like that: deep learning for interpretable image recognition.
\newblock Advances in neural information processing systems. 2019;32.

\bibitem{stepin2021survey}
Stepin I, Alonso JM, Catala A, Pereira-Fari{\~n}a M.
\newblock A survey of contrastive and counterfactual explanation generation methods for explainable artificial intelligence.
\newblock IEEE Access. 2021;9:11974-2001.

\bibitem{jiang2018trust}
Jiang H, Kim B, Guan M, Gupta M.
\newblock To trust or not to trust a classifier.
\newblock Advances in neural information processing systems. 2018;31.

\bibitem{page2021prisma}
Page MJ, McKenzie JE, Bossuyt PM, Boutron I, Hoffmann TC, Mulrow CD, et~al.
\newblock The PRISMA 2020 statement: an updated guideline for reporting systematic reviews.
\newblock bmj. 2021;372.

\bibitem{greenhalgh2018time}
Greenhalgh T, Thorne S, Malterud K.
\newblock Time to challenge the spurious hierarchy of systematic over narrative reviews?
\newblock European journal of clinical investigation. 2018;48(6):e12931.

\bibitem{sukhera2022narrative}
Sukhera J.
\newblock Narrative reviews: flexible, rigorous, and practical.
\newblock Journal of graduate medical education. 2022;14(4):414-7.

\bibitem{lu2021data}
Lu MY, Williamson DF, Chen TY, Chen RJ, Barbieri M, Mahmood F.
\newblock Data-efficient and weakly supervised computational pathology on whole-slide images.
\newblock Nature biomedical engineering. 2021;5(6):555-70.

\bibitem{campanella2019clinical}
Campanella G, Hanna MG, Geneslaw L, Miraflor A, Werneck Krauss~Silva V, Busam KJ, et~al.
\newblock Clinical-grade computational pathology using weakly supervised deep learning on whole slide images.
\newblock Nature medicine. 2019;25(8):1301-9.

\bibitem{innani2025ai}
Innani S, Bell WR, Nasrallah MP, Baheti B, Bakas S.
\newblock AI-driven WHO 2021 classification of gliomas based only on H\&E-stained slides.
\newblock Neuro-Oncology. 2025:noaf189.

\bibitem{chen2022scaling}
Chen RJ, Chen C, Li Y, Chen TY, Trister AD, Krishnan RG, et~al.
\newblock Scaling vision transformers to gigapixel images via hierarchical self-supervised learning.
\newblock In: Proceedings of the IEEE/CVF conference on computer vision and pattern recognition; 2022. p. 16144-55.

\bibitem{woerl2020deep}
Woerl AC, Eckstein M, Geiger J, Wagner DC, Daher T, Stenzel P, et~al.
\newblock Deep learning predicts molecular subtype of muscle-invasive bladder cancer from conventional histopathological slides.
\newblock European urology. 2020;78(2):256-64.

\bibitem{ji2019gradient}
Ji J.
\newblock Gradient-based interpretation on convolutional neural network for classification of pathological images.
\newblock In: 2019 International Conference on Information Technology and Computer Application (ITCA). IEEE; 2019. p. 83-6.

\bibitem{sadafi2023pixel}
Sadafi A, Adonkina O, Khakzar A, Lienemann P, Hehr RM, Rueckert D, et~al.
\newblock Pixel-level explanation of multiple instance learning models in biomedical single cell images.
\newblock In: International Conference on Information Processing in Medical Imaging. Springer; 2023. p. 170-82.

\bibitem{chattopadhay2018grad}
Chattopadhay A, Sarkar A, Howlader P, Balasubramanian VN.
\newblock Grad-cam++: Generalized gradient-based visual explanations for deep convolutional networks.
\newblock In: 2018 IEEE winter conference on applications of computer vision (WACV). IEEE; 2018. p. 839-47.

\bibitem{han2022multi}
Han C, Lin J, Mai J, Wang Y, Zhang Q, Zhao B, et~al.
\newblock Multi-layer pseudo-supervision for histopathology tissue semantic segmentation using patch-level classification labels.
\newblock Medical Image Analysis. 2022;80:102487.

\bibitem{simonyan2013deep}
Simonyan K, Vedaldi A, Zisserman A.
\newblock Deep inside convolutional networks: Visualising image classification models and saliency maps.
\newblock arXiv preprint arXiv:13126034. 2013.

\bibitem{bejnordi2017diagnostic}
Bejnordi BE, Veta M, Van~Diest PJ, Van~Ginneken B, Karssemeijer N, Litjens G, et~al.
\newblock Diagnostic assessment of deep learning algorithms for detection of lymph node metastases in women with breast cancer.
\newblock Jama. 2017;318(22):2199-210.

\bibitem{smilkov2017smoothgrad}
Smilkov D, Thorat N, Kim B, Vi{\'e}gas F, Wattenberg M.
\newblock Smoothgrad: removing noise by adding noise.
\newblock arXiv preprint arXiv:170603825. 2017.

\bibitem{dolezal2023deep}
Dolezal JM, Wolk R, Hieromnimon HM, Howard FM, Srisuwananukorn A, Karpeyev D, et~al.
\newblock Deep learning generates synthetic cancer histology for explainability and education.
\newblock NPJ precision oncology. 2023;7(1):49.

\bibitem{srinivas2019fullgrad}
Srinivas S, Fleuret F.
\newblock Full-gradient representation for neural network visualization.
\newblock Advances in neural information processing systems. 2019;32.

\bibitem{mondol2025graphite}
Mondol RK, Millar EK, Graham PH, Browne L, Sowmya A, Meijering E.
\newblock GRAPHITE: Graph-based interpretable tissue examination for enhanced explainability in breast cancer histopathology.
\newblock Computers in Biology and Medicine. 2025;197:111106.

\bibitem{chen2026interpretable}
Chen Q, Wang Z, Lin X, Shi Y, Xu B, Chai J, et~al.
\newblock Interpretable multitask model for clinical pathology image prediction and interpretation.
\newblock npj Systems Biology and Applications. 2026.

\bibitem{bach2015pixel}
Bach S, Binder A, Montavon G, Klauschen F, M{\"u}ller KR, Samek W.
\newblock On pixel-wise explanations for non-linear classifier decisions by layer-wise relevance propagation.
\newblock PloS one. 2015;10(7):e0130140.

\bibitem{hense2024xmil}
Hense J, Jamshidi~Idaji M, Eberle O, Schnake T, Dippel J, Ciernik L, et~al.
\newblock xMIL: Insightful Explanations for Multiple Instance Learning in Histopathology.
\newblock arXiv e-prints. 2024:arXiv-2406.

\bibitem{hossain2022early}
Hossain M, Haque SS, Ahmed H, Mahdi HA, Aich A. Early stage detection and classification of colon cancer using deep learning and explainable AI on histopathological images. Brac University; 2022.

\bibitem{coudray2018classification}
Coudray N, Ocampo PS, Sakellaropoulos T, Narula N, Snuderl M, Feny{\"o} D, et~al.
\newblock Classification and mutation prediction from non--small cell lung cancer histopathology images using deep learning.
\newblock Nature medicine. 2018;24(10):1559-67.

\bibitem{rahnfeld2024comparative}
Rahnfeld J, Naouar M, Kalweit G, Boedecker J, Dubruc E, Kalweit M.
\newblock A Comparative Study of Explainability Methods for Whole Slide Classification of Lymph Node Metastases using Vision Transformers.
\newblock medRxiv. 2024:2024-05.

\bibitem{peta2024explainable}
Peta J, Koppu S.
\newblock Explainable Soft Attentive EfficientNet for breast cancer classification in histopathological images.
\newblock Biomedical Signal Processing and Control. 2024;90:105828.

\bibitem{palatnik2019local}
Palatnik~de Sousa I, Maria Bernardes Rebuzzi~Vellasco M, Costa~da Silva E.
\newblock Local interpretable model-agnostic explanations for classification of lymph node metastases.
\newblock Sensors. 2019;19(13):2969.

\bibitem{bora2024slice}
Bora RP, Terh{\"o}rst P, Veldhuis R, Ramachandra R, Raja K.
\newblock Slice: Stabilized lime for consistent explanations for image classification.
\newblock In: Proceedings of the IEEE/CVF Conference on Computer Vision and Pattern Recognition; 2024. p. 10988-96.

\bibitem{zhao2021baylime}
Zhao X, Huang W, Huang X, Robu V, Flynn D.
\newblock Baylime: Bayesian local interpretable model-agnostic explanations.
\newblock In: Uncertainty in artificial intelligence. PMLR; 2021. p. 887-96.

\bibitem{gallo2023shedding}
Gallo M, Kraj{\v{n}}ansk{\`y} V, Nenutil R, Holub P, Br{\'a}zdil T.
\newblock Shedding light on the black box of a neural network used to detect prostate cancer in whole slide images by occlusion-based explainability.
\newblock New Biotechnology. 2023;78:52-67.

\bibitem{kaczmarzyk2024explainable}
Kaczmarzyk JR, Saltz JH, Koo PK.
\newblock Explainable AI for computational pathology identifies model limitations and tissue biomarkers.
\newblock ArXiv. 2024:arXiv-2409.

\bibitem{englebert2023explaining}
Englebert A, Stassin S, Nanfack G, Mahmoudi SA, Siebert X, Cornu O, et~al.
\newblock Explaining through transformer input sampling.
\newblock In: Proceedings of the IEEE/CVF International Conference on Computer Vision; 2023. p. 806-15.

\bibitem{sitaula2020fusion}
Sitaula C, Aryal S.
\newblock Fusion of whole and part features for the classification of histopathological image of breast tissue.
\newblock Health Information Science and Systems. 2020;8:1-12.

\bibitem{bioengineering10040396}
Altini N, Puro E, Taccogna MG, Marino F, De~Summa S, Saponaro C, et~al.
\newblock Tumor Cellularity Assessment of Breast Histopathological Slides via Instance Segmentation and Pathomic Features Explainability.
\newblock Bioengineering. 2023;10(4).
\newblock Available from: \url{https://www.mdpi.com/2306-5354/10/4/396}.

\bibitem{wang2024shap}
Wang J, Mao Y, Guan N, Xue CJ.
\newblock SHAP-CAT: A interpretable multi-modal framework enhancing WSI classification via virtual staining and shapley-value-based multimodal fusion.
\newblock arXiv preprint arXiv:241001408. 2024.

\bibitem{hussein2025vision}
Hussein A, Prasad M, Anaissi A, Braytee A.
\newblock Vision Transformers with Autoencoders and Explainable AI for Cancer Patient Risk Stratification Using Whole Slide Imaging.
\newblock arXiv preprint arXiv:250404749. 2025.

\bibitem{gamble2021determining}
Gamble P, Jaroensri R, Wang H, Tan F, Moran M, Brown T, et~al.
\newblock Determining breast cancer biomarker status and associated morphological features using deep learning.
\newblock Communications medicine. 2021;1(1):14.

\bibitem{vzitvnanskyconcept}
{\v{Z}}IT{\v{N}}ANSK{\'Y} BA.
\newblock Concept learning in digital pathology.
\newblock Academic Thesis. 2024.

\bibitem{Mittmann2025}
Mittmann G, Laiouar-Pedari S, Mehrtens HA, Haggenm{\"u}ller S, Bucher TC, Chanda T, et~al.
\newblock Pathologist-like explainable AI for interpretable Gleason grading in prostate cancer.
\newblock Nature Communications. 2025 Oct;16(1):8959.

\bibitem{kazmierczak2025explainability}
Kazmierczak R, Berthier E, Frehse G, Franchi G.
\newblock Explainability and vision foundation models: A survey.
\newblock Information Fusion. 2025;122:103184.

\bibitem{bannur2023learning}
Bannur S, Hyland S, Liu Q, Perez-Garcia F, Ilse M, Castro DC, et~al.
\newblock Learning to exploit temporal structure for biomedical vision-language processing.
\newblock In: Proceedings of the IEEE/CVF Conference on Computer Vision and Pattern Recognition; 2023. p. 15016-27.

\bibitem{evans2022explainability}
Evans T, Retzlaff CO, Gei{\ss}ler C, Kargl M, Plass M, M{\"u}ller H, et~al.
\newblock The explainability paradox: Challenges for xAI in digital pathology.
\newblock Future Generation Computer Systems. 2022;133:281-96.

\bibitem{li2018deep}
Li O, Liu H, Chen C, Rudin C.
\newblock Deep learning for case-based reasoning through prototypes: A neural network that explains its predictions.
\newblock In: Proceedings of the AAAI conference on artificial intelligence. vol.~32; 2018. .

\bibitem{zigutyte}
{\v Z}igutyte L, Lenz T, Han T, Hewitt KJ, Reitsam NG, Foersch S, et~al.
\newblock Counterfactual Diffusion Models for Mechanistic Explainability of Artificial Intelligence Models in Pathology.
\newblock bioRxiv. 2024.
\newblock Available from: \url{https://www.biorxiv.org/content/early/2024/11/03/2024.10.29.620913}.

\bibitem{markus2021role}
Markus AF, Kors JA, Rijnbeek PR.
\newblock The role of explainability in creating trustworthy artificial intelligence for health care: a comprehensive survey of the terminology, design choices, and evaluation strategies.
\newblock Journal of biomedical informatics. 2021;113:103655.

\bibitem{templeton2024scaling}
Templeton A, Conerly T, Marcus J, Lindsey J, Bricken T, Chen B, et~al.
\newblock Scaling monosemanticity: extracting interpretable features from {Claude} 3 {Sonnet}.
\newblock Transformer Circuits Thread. 2024.

\bibitem{cunningham2023sparse}
Cunningham H, Ewart A, Riggs L, Huben R, Sharkey L.
\newblock Sparse autoencoders find highly interpretable features in language models.
\newblock arXiv preprint arXiv:230908600. 2023.

\bibitem{scholkopf2021toward}
Sch{\"o}lkopf B, Locatello F, Bauer S, Ke NR, Kalchbrenner N, Goyal A, et~al.
\newblock Toward causal representation learning.
\newblock Proceedings of the IEEE. 2021;109(5):612-34.

\bibitem{zecevic2021relating}
Ze{\v{c}}evi{\'c} M, Dhami DS, Veli{\v{c}}kovi{\'c} P, Kersting K.
\newblock Relating graph neural networks to structural causal models.
\newblock In: arXiv preprint arXiv:2109.04173; 2021. .

\bibitem{tang2019interpretable}
Tang Z, Chuang KV, DeCarli C, Jin LW, Beckett L, Keiser MJ, et~al.
\newblock Interpretable classification of Alzheimer’s disease pathologies with a convolutional neural network pipeline.
\newblock Nature communications. 2019;10(1):2173.

\bibitem{kubach2020same}
Kubach J, Muhlebner-Fahrngruber A, Soylemezoglu F, Miyata H, Niehusmann P, Honavar M, et~al.
\newblock Same same but different: A Web-based deep learning application revealed classifying features for the histopathologic distinction of cortical malformations.
\newblock Epilepsia. 2020;61(3):421-32.

\bibitem{baheti2025multimodal}
Baheti B, Rai S, Innani S, Mehdiratta G, Bell WR, Guntuku SC, et~al.
\newblock Multimodal Explainable Artificial Intelligence for Prognostic Stratification of Glioblastoma Patients.
\newblock Modern Pathology. 2025:100797.

\bibitem{innani2025artificial}
Innani S, Bell WR, Nasrallah M, Baheti B, Bakas S.
\newblock Artificial intelligence predicts 2021 WHO glioma subtypes from whole slide images.
\newblock Cancer Research. 2025;85(8\_Supplement\_1):6247-7.

\bibitem{innani2025interpretable}
Innani S, Bell WR, Harmsen H, Nasrallah MP, Baheti B, Bakas S.
\newblock Interpretable artificial intelligence based determination of glioma IDH mutation status directly from histology slides.
\newblock Neuro-Oncology Advances. 2025;7(1):vdaf140.

\bibitem{dorrich2023explainable}
D{\"o}rrich M, Hecht M, Fietkau R, Hartmann A, Iro H, Gostian AO, et~al.
\newblock Explainable convolutional neural networks for assessing head and neck cancer histopathology.
\newblock Diagnostic Pathology. 2023;18(1):121.

\bibitem{gecer2018detection}
Gecer B, Aksoy S, Mercan E, Shapiro LG, Weaver DL, Elmore JG.
\newblock Detection and classification of cancer in whole slide breast histopathology images using deep convolutional networks.
\newblock Pattern recognition. 2018;84:345-56.

\bibitem{obikane2020weakly}
Obikane S, Aoki Y.
\newblock Weakly supervised domain adaptation with point supervision in histopathological image segmentation.
\newblock In: Pattern Recognition: ACPR 2019 Workshops, Auckland, New Zealand, November 26, 2019, Proceedings 5. Springer; 2020. p. 127-40.

\bibitem{pirovano2020improving}
Pirovano A, Heuberger H, Berlemont S, Ladjal S, Bloch I.
\newblock Improving interpretability for computer-aided diagnosis tools on whole slide imaging with multiple instance learning and gradient-based explanations.
\newblock In: Interpretable and Annotation-Efficient Learning for Medical Image Computing: Third International Workshop, iMIMIC 2020, Second International Workshop, MIL3ID 2020, and 5th International Workshop, LABELS 2020, Held in Conjunction with MICCAI 2020, Lima, Peru, October 4--8, 2020, Proceedings 3. Springer; 2020. p. 43-53.

\bibitem{yang2019guided}
Yang H, Kim JY, Kim H, Adhikari SP.
\newblock Guided soft attention network for classification of breast cancer histopathology images.
\newblock IEEE transactions on medical imaging. 2019;39(5):1306-15.

\bibitem{zeiser2021deepbatch}
Zeiser FA, da~Costa CA, de~Oliveira~Ramos G, Bohn HC, Santos I, Roehe AV.
\newblock DeepBatch: A hybrid deep learning model for interpretable diagnosis of breast cancer in whole-slide images.
\newblock Expert Systems with Applications. 2021;185:115586.

\bibitem{liang2023interpretable}
Liang M, Chen Q, Li B, Wang L, Wang Y, Zhang Y, et~al.
\newblock Interpretable classification of pathology whole-slide images using attention based context-aware graph convolutional neural network.
\newblock Computer Methods and Programs in Biomedicine. 2023;229:107268.

\bibitem{dooper2023gigapixel}
Dooper S, Pinckaers H, Aswolinskiy W, Hebeda K, Jarkman S, van~der Laak J, et~al.
\newblock Gigapixel end-to-end training using streaming and attention.
\newblock Medical Image Analysis. 2023:102881.

\bibitem{sharma2021}
Sharma Y, Shrivastava A, Ehsan L, Moskaluk CA, Syed S, Brown D.
\newblock Cluster-to-Conquer: A Framework for End-to-End Multi-Instance Learning for Whole Slide Image Classification.
\newblock In: Heinrich M, Dou Q, de~Bruijne M, Lellmann J, Schläfer A, Ernst F, editors. Proceedings of the Fourth Conference on Medical Imaging with Deep Learning. vol. 143 of Proceedings of Machine Learning Research; 2021. p. 682-98.

\bibitem{binder2021morphological}
Binder A, Bockmayr M, H{\"a}gele M, Wienert S, Heim D, Hellweg K, et~al.
\newblock Morphological and molecular breast cancer profiling through explainable machine learning.
\newblock Nature Machine Intelligence. 2021;3(4):355-66.

\bibitem{seegerer2020interpretable}
Seegerer P, Binder A, Saitenmacher R, Bockmayr M, Alber M, Jurmeister P, et~al.
\newblock Interpretable deep neural network to predict estrogen receptor status from haematoxylin-eosin images.
\newblock Artificial Intelligence and Machine Learning for Digital Pathology: State-of-the-Art and Future Challenges. 2020:16-37.

\bibitem{graziani2019visualizing}
Graziani M, Andrearczyk V, Muller H.
\newblock Visualizing and interpreting feature reuse of pretrained CNNs for histopathology.
\newblock Technological University Dublin. 2019.

\bibitem{francis}
Francis A U~Imouokhome OGE, Chete FO.
\newblock Diagnosis and Interpretation of Breast Cancer Using Explainable Artificial Intelligence.
\newblock NIPES - Journal of Science and Technology Research. 2023 Jun;5(2).
\newblock Available from: \url{https://journals.nipes.org/index.php/njstr/article/view/603}.

\bibitem{afshar2024ibo}
Afshar P, Hashembeiki S, Khani P, Fatemizadeh E, Rohban MH.
\newblock Ibo: Inpainting-based occlusion to enhance explainable artificial intelligence evaluation in histopathology.
\newblock arXiv preprint arXiv:240816395. 2024.

\bibitem{he2017deep}
He J, Shang L, Ji H, Zhang X.
\newblock Deep learning features for lung adenocarcinoma classification with tissue pathology images.
\newblock In: Neural Information Processing: 24th International Conference, ICONIP 2017, Guangzhou, China, November 14--18, 2017, Proceedings, Part IV 24. Springer; 2017. p. 742-51.

\bibitem{teramoto2019automated}
Teramoto A, Yamada A, Kiriyama Y, Tsukamoto T, Yan K, Zhang L, et~al.
\newblock Automated classification of benign and malignant cells from lung cytological images using deep convolutional neural network.
\newblock Informatics in Medicine Unlocked. 2019;16:100205.

\bibitem{zhang2022dtfd}
Zhang H, Meng Y, Zhao Y, Qiao Y, Yang X, Coupland SE, et~al.
\newblock DTFD-MIL: Double-tier feature distillation multiple instance learning for histopathology whole slide image classification.
\newblock In: Proceedings of the IEEE/CVF Conference on Computer Vision and Pattern Recognition; 2022. p. 18802-12.

\bibitem{wang2019weakly}
Wang X, Chen H, Gan C, Lin H, Dou Q, Tsougenis E, et~al.
\newblock Weakly supervised deep learning for whole slide lung cancer image analysis.
\newblock IEEE transactions on cybernetics. 2019;50(9):3950-62.

\bibitem{zhao2021lung}
Zhao L, Xu X, Hou R, Zhao W, Zhong H, Teng H, et~al.
\newblock Lung cancer subtype classification using histopathological images based on weakly supervised multi-instance learning.
\newblock Physics in Medicine \& Biology. 2021;66(23):235013.

\bibitem{gupta2020region}
Gupta M, Das C, Roy A, Gupta P, Pillai GR, Patole K.
\newblock Region of interest identification for cervical cancer images.
\newblock In: 2020 IEEE 17th International Symposium on Biomedical Imaging (ISBI). IEEE; 2020. p. 1293-6.

\bibitem{kiran2019automatic}
Kiran~GV K, Meghana~Reddy G.
\newblock Automatic classification of whole slide pap smear images using CNN with PCA based feature interpretation.
\newblock In: Proceedings of the IEEE/CVF Conference on Computer Vision and Pattern Recognition Workshops; 2019. p. 0-0.

\bibitem{fremond2023interpretable}
Fremond S, Andani S, Wolf JB, Dijkstra J, Melsbach S, Jobsen JJ, et~al.
\newblock Interpretable deep learning model to predict the molecular classification of endometrial cancer from haematoxylin and eosin-stained whole-slide images: a combined analysis of the PORTEC randomised trials and clinical cohorts.
\newblock The Lancet Digital Health. 2023;5(2):e71-82.

\bibitem{heinemann2019deep}
Heinemann F, Birk G, Stierstorfer B.
\newblock Deep learning enables pathologist-like scoring of NASH models.
\newblock Scientific reports. 2019;9(1):18454.

\bibitem{kiani2020impact}
Kiani A, Uyumazturk B, Rajpurkar P, Wang A, Gao R, Jones E, et~al.
\newblock Impact of a deep learning assistant on the histopathologic classification of liver cancer.
\newblock NPJ digital medicine. 2020;3(1):23.

\bibitem{yan2020prior}
Yan C, Xu J, Xie J, Cai C, Lu H.
\newblock Prior-aware CNN with multi-task learning for colon images analysis.
\newblock In: 2020 IEEE 17th International Symposium on Biomedical Imaging (ISBI). IEEE; 2020. p. 254-7.

\bibitem{kather2019predicting}
Kather JN, Krisam J, Charoentong P, Luedde T, Herpel E, Weis CA, et~al.
\newblock Predicting survival from colorectal cancer histology slides using deep learning: A retrospective multicenter study.
\newblock PLoS medicine. 2019;16(1):e1002730.

\bibitem{korbar2017looking}
Korbar B, Olofson AM, Miraflor AP, Nicka CM, Suriawinata MA, Torresani L, et~al.
\newblock Looking under the hood: Deep neural network visualization to interpret whole-slide image analysis outcomes for colorectal polyps.
\newblock In: Proceedings of the IEEE conference on computer vision and pattern recognition workshops; 2017. p. 69-75.

\bibitem{kowsari2020hmic}
Kowsari K, Sali R, Ehsan L, Adorno W, Ali A, Moore S, et~al.
\newblock Hmic: Hierarchical medical image classification, a deep learning approach.
\newblock Information. 2020;11(6):318.

\bibitem{niehues2023generalizable}
Niehues JM, Quirke P, West NP, Grabsch HI, van Treeck M, Schirris Y, et~al.
\newblock Generalizable biomarker prediction from cancer pathology slides with self-supervised deep learning: A retrospective multi-centric study.
\newblock Cell Reports Medicine. 2023;4(4).

\bibitem{li2021multi}
Li J, Li W, Sisk A, Ye H, Wallace WD, Speier W, et~al.
\newblock A multi-resolution model for histopathology image classification and localization with multiple instance learning.
\newblock Computers in biology and medicine. 2021;131:104253.

\bibitem{kanavati2021deep}
Kanavati F, Ichihara S, Rambeau M, Iizuka O, Arihiro K, Tsuneki M.
\newblock Deep learning models for gastric signet ring cell carcinoma classification in whole slide images.
\newblock Technology in Cancer Research \& Treatment. 2021;20:15330338211027901.

\bibitem{laleh2021deep}
Laleh NG, Echle A, Muti HS, Hewitt KJ, Schulz V, Kather JN.
\newblock Deep Learning for interpretable end-to-end survival prediction in gastrointestinal cancer histopathology.
\newblock In: COMPAY 2021: The third MICCAI workshop on Computational Pathology; 2021. .

\bibitem{kuntz2021gastrointestinal}
Kuntz S, Krieghoff-Henning E, Kather JN, Jutzi T, H{\"o}hn J, Kiehl L, et~al.
\newblock Gastrointestinal cancer classification and prognostication from histology using deep learning: Systematic review.
\newblock European Journal of Cancer. 2021;155:200-15.

\bibitem{neto2022imil4path}
Neto PC, Oliveira SP, Montezuma D, Fraga J, Monteiro A, Ribeiro L, et~al.
\newblock iMIL4PATH: A semi-supervised interpretable approach for colorectal whole-slide images.
\newblock Cancers. 2022;14(10):2489.

\bibitem{zhu2020weakly}
Zhu Z, Ding X, Zhang D, Wang L.
\newblock Weakly-supervised balanced attention network for gastric pathology image localization and classification.
\newblock In: 2020 IEEE 17th International Symposium on Biomedical Imaging (ISBI). IEEE; 2020. p. 1-4.

\bibitem{calderaro2023deep}
Calderaro J, Ghaffari~Laleh N, Zeng Q, Maille P, Favre L, Pujals A, et~al.
\newblock Deep learning-based phenotyping reclassifies combined hepatocellular-cholangiocarcinoma.
\newblock Nature Communications. 2023;14(1):8290.

\bibitem{zeng2023artificial}
Zeng Q, Klein C, Caruso S, Maille P, Allende DS, M{\'\i}nguez B, et~al.
\newblock Artificial intelligence-based pathology as a biomarker of sensitivity to atezolizumab--bevacizumab in patients with hepatocellular carcinoma: a multicentre retrospective study.
\newblock The Lancet Oncology. 2023;24(12):1411-22.

\bibitem{jiang2024end}
Jiang X, Hoffmeister M, Brenner H, Muti HS, Yuan T, Foersch S, et~al.
\newblock End-to-end prognostication in colorectal cancer by deep learning: a retrospective, multicentre study.
\newblock The Lancet Digital Health. 2024;6(1):e33-43.

\bibitem{zhang2019pathologist}
Zhang Z, Chen P, McGough M, Xing F, Wang C, Bui M, et~al.
\newblock Pathologist-level interpretable whole-slide cancer diagnosis with deep learning.
\newblock Nature Machine Intelligence. 2019;1(5):236-45.

\bibitem{chakraborty2022predicting}
Chakraborty S, Gupta R, Ma K, Govind D, Sarder P, Choi WT, et~al.
\newblock Predicting the Visual Attention of Pathologists Evaluating Whole Slide Images of Cancer.
\newblock In: International Workshop on Medical Optical Imaging and Virtual Microscopy Image Analysis. Springer; 2022. p. 11-21.

\bibitem{wang2021predicting}
Wang X, Chen Y, Gao Y, Zhang H, Guan Z, Dong Z, et~al.
\newblock Predicting gastric cancer outcome from resected lymph node histopathology images using deep learning.
\newblock Nature communications. 2021;12(1):1637.

\bibitem{wu2023artificial}
Wu S, Hong G, Xu A, Zeng H, Chen X, Wang Y, et~al.
\newblock Artificial intelligence-based model for lymph node metastases detection on whole slide images in bladder cancer: a retrospective, multicentre, diagnostic study.
\newblock The Lancet Oncology. 2023;24(4):360-70.

\bibitem{yu2023prototypical}
Yu JG, Wu Z, Ming Y, Deng S, Li Y, Ou C, et~al.
\newblock Prototypical multiple instance learning for predicting lymph node metastasis of breast cancer from whole-slide pathological images.
\newblock Medical Image Analysis. 2023;85:102748.

\bibitem{civit2022non}
Civit-Masot J, Ba{\~n}uls-Beaterio A, Dom{\'\i}nguez-Morales M, Rivas-P{\'e}rez M, Mu{\~n}oz-Saavedra L, Corral JMR.
\newblock Non-small cell lung cancer diagnosis aid with histopathological images using Explainable Deep Learning techniques.
\newblock Computer Methods and Programs in Biomedicine. 2022;226:107108.

\bibitem{10643508}
Radhakrishnan M, Sampathila N, Muralikrishna H, Swathi KS.
\newblock Advancing Ovarian Cancer Diagnosis Through Deep Learning and eXplainable AI: A Multiclassification Approach.
\newblock IEEE Access. 2024;12:116968-86.

\bibitem{pinckaers2021detection}
Pinckaers H, Bulten W, van~der Laak J, Litjens G.
\newblock Detection of prostate cancer in whole-slide images through end-to-end training with image-level labels.
\newblock IEEE Transactions on Medical Imaging. 2021;40(7):1817-26.

\bibitem{ferrero2024histoem}
Ferrero A, Ghelichkhan E, Manoochehri H, Ho MM, Albertson DJ, Brintz BJ, et~al.
\newblock HistoEM: A Pathologist-Guided and Explainable Workflow Using Histogram Embedding for Gland Classification.
\newblock Modern Pathology. 2024;37(4):100447.

\bibitem{silva2020going}
Silva-Rodr{\'\i}guez J, Colomer A, Sales MA, Molina R, Naranjo V.
\newblock Going deeper through the Gleason scoring scale: An automatic end-to-end system for histology prostate grading and cribriform pattern detection.
\newblock Computer methods and programs in biomedicine. 2020;195:105637.

\bibitem{singhal2022deep}
Singhal N, Soni S, Bonthu S, Chattopadhyay N, Samanta P, Joshi U, et~al.
\newblock A deep learning system for prostate cancer diagnosis and grading in whole slide images of core needle biopsies.
\newblock Scientific reports. 2022;12(1):3383.

\bibitem{tavolara2023one}
Tavolara TE, Su Z, Gurcan MN, Niazi MKK.
\newblock One label is all you need: Interpretable AI-enhanced histopathology for oncology.
\newblock In: Seminars in Cancer Biology. Elsevier; 2023. .

\bibitem{xiang2023automatic}
Xiang J, Wang X, Wang X, Zhang J, Yang S, Yang W, et~al.
\newblock Automatic diagnosis and grading of Prostate Cancer with weakly supervised learning on whole slide images.
\newblock Computers in Biology and Medicine. 2023;152:106340.

\bibitem{grisi2023hierarchical}
Grisi C, Litjens G, van~der Laak J.
\newblock Hierarchical Vision Transformers for Context-Aware Prostate Cancer Grading in Whole Slide Images.
\newblock arXiv preprint arXiv:231212619. 2023.

\bibitem{bhattacharjee2022explainable}
Bhattacharjee S, Hwang YB, Ikromjanov K, Sumon RI, Kim HC, Choi HK.
\newblock An explainable computer vision in histopathology: techniques for interpreting black box model.
\newblock In: 2022 International Conference on Artificial Intelligence in Information and Communication (ICAIIC). IEEE; 2022. p. 392-8.

\bibitem{thomas2021interpretable}
Thomas SM, Lefevre JG, Baxter G, Hamilton NA.
\newblock Interpretable deep learning systems for multi-class segmentation and classification of non-melanoma skin cancer.
\newblock Medical Image Analysis. 2021;68:101915.

\bibitem{xie2019interpretable}
Xie P, Zuo K, Zhang Y, Li F, Yin M, Lu K.
\newblock Interpretable classification from skin cancer histology slides using deep learning: A retrospective multicenter study.
\newblock arXiv preprint arXiv:190406156. 2019.

\bibitem{del2021attention}
del Amor R, Launet L, Colomer A, Moscard{\'o} A, Mosquera-Zamudio A, Monteagudo C, et~al.
\newblock An attention-based weakly supervised framework for spitzoid melanocytic lesion diagnosis in WSI.
\newblock arXiv preprint arXiv:210409878. 2021.

\bibitem{geijs2023detection}
Geijs D, Dooper S, Aswolinskiy W, Hillen LM, Amir AL, Litjens G.
\newblock Detection and subtyping of basal cell carcinoma in whole-slide histopathology using weakly-supervised learning.
\newblock Medical Image Analysis. 2023:103063.

\bibitem{xu2017large}
Xu Y, Jia Z, Wang LB, Ai Y, Zhang F, Lai M, et~al.
\newblock Large scale tissue histopathology image classification, segmentation, and visualization via deep convolutional activation features.
\newblock BMC bioinformatics. 2017;18:1-17.

\bibitem{chan2019histosegnet}
Chan L, Hosseini MS, Rowsell C, Plataniotis KN, Damaskinos S.
\newblock Histosegnet: Semantic segmentation of histological tissue type in whole slide images.
\newblock In: Proceedings of the IEEE/CVF International Conference on Computer Vision; 2019. p. 10662-71.

\bibitem{huang2019evidence}
Huang Y, Chung AC.
\newblock Evidence localization for pathology images using weakly supervised learning.
\newblock In: Medical Image Computing and Computer Assisted Intervention--MICCAI 2019: 22nd International Conference, Shenzhen, China, October 13--17, 2019, Proceedings, Part I 22. Springer; 2019. p. 613-21.

\bibitem{hagele2020resolving}
H{\"a}gele M, Seegerer P, Lapuschkin S, Bockmayr M, Samek W, Klauschen F, et~al.
\newblock Resolving challenges in deep learning-based analyses of histopathological images using explanation methods.
\newblock Scientific reports. 2020;10(1):6423.

\bibitem{keyl2023decoding}
Keyl J, Keyl P, Montavon G, Hosch R, Brehmer A, Mochmann L, et~al.
\newblock Decoding pan-cancer treatment outcomes using multimodal real-world data and explainable artificial intelligence.
\newblock medRxiv. 2023:2023-10.

\bibitem{saldanha2023self}
Saldanha OL, Loeffler CM, Niehues JM, van Treeck M, Seraphin TP, Hewitt KJ, et~al.
\newblock Self-supervised attention-based deep learning for pan-cancer mutation prediction from histopathology.
\newblock NPJ Precision Oncology. 2023;7(1):35.

\bibitem{lee2022derivation}
Lee Y, Park JH, Oh S, Shin K, Sun J, Jung M, et~al.
\newblock Derivation of prognostic contextual histopathological features from whole-slide images of tumours via graph deep learning.
\newblock Nature Biomedical Engineering. 2022:1-15.

\bibitem{wang2023surformer}
Wang Z, Gao Q, Yi X, Zhang X, Zhang Y, Zhang D, et~al.
\newblock Surformer: An interpretable pattern-perceptive survival transformer for cancer survival prediction from histopathology whole slide images.
\newblock Computer Methods and Programs in Biomedicine. 2023;241:107733.

\bibitem{shao2021transmil}
Shao Z, Bian H, Chen Y, Wang Y, Zhang J, Ji X, et~al.
\newblock Transmil: Transformer based correlated multiple instance learning for whole slide image classification.
\newblock Advances in neural information processing systems. 2021;34:2136-47.

\bibitem{wang2022transformer}
Wang X, Yang S, Zhang J, Wang M, Zhang J, Yang W, et~al.
\newblock Transformer-based unsupervised contrastive learning for histopathological image classification.
\newblock Medical image analysis. 2022;81:102559.

\bibitem{kapse2024si}
Kapse S, Pati P, Das S, Zhang J, Chen C, Vakalopoulou M, et~al.
\newblock SI-MIL: Taming Deep MIL for Self-Interpretability in Gigapixel Histopathology.
\newblock In: Proceedings of the IEEE/CVF Conference on Computer Vision and Pattern Recognition; 2024. p. 11226-37.

\bibitem{diao2021human}
Diao JA, Wang JK, Chui WF, Mountain V, Gullapally SC, Srinivasan R, et~al.
\newblock Human-interpretable image features derived from densely mapped cancer pathology slides predict diverse molecular phenotypes.
\newblock Nature communications. 2021;12(1):1613.

\end{thebibliography}
\end{document}